\documentclass{article}

\usepackage{microtype}
\usepackage{graphicx}
\usepackage{subfigure}
\usepackage{booktabs}
\usepackage{bbm}
\usepackage{comment}
\usepackage{hyperref}
\usepackage{amsmath}
\usepackage{amssymb}
\usepackage{amsthm}
\usepackage{authblk}
\usepackage{natbib}
\usepackage{algorithmic}
\usepackage{algorithm}
\usepackage{xcolor}
\usepackage{bbold}
\usepackage{xfrac}
\usepackage{mathtools}
\usepackage{physics}
\usepackage{wrapfig}
\usepackage{nicefrac}
\usepackage{tikz}
\usepackage[margin=1in]{geometry}
\usepackage[capitalize,noabbrev]{cleveref}
\setcitestyle{authoryear,round,citesep={;},aysep={,},yysep={;}}
\usepackage{math_commands}

\hypersetup{
    colorlinks = true,
    linkcolor = {blue},
    citecolor = {blue},
    urlcolor = {blue}
}

\tikzset{
  midarr/.style={decoration={markings,mark=at position #1 with {\arrow{stealth}}},postaction={decorate}},
  midarr/.default=0.5
}

\usetikzlibrary{calc,intersections,decorations.markings}

\definecolor{bulk}{RGB}{178,178,178}
\definecolor{phaseI}{RGB}{255,217,47}
\definecolor{phaseII}{RGB}{100,168,195}
\definecolor{phaseIII}{RGB}{166,216,84}
\definecolor{phaseIV}{RGB}{141,160,203}
\definecolor{phaseV}{RGB}{252,141,98}
\definecolor{phaseVI}{RGB}{80,214,165}

\author[1]{Leonardo Defilippis\thanks{These authors contributed equally to this work.}}
\author[2,3]{Yizhou Xu$^*$}
\author[2]{Julius Girardin}
\author[2]{Emanuele Troiani}
\author[2]{Vittorio Erba}
\author[2]{Lenka Zdeborová}
\author[1]{Bruno Loureiro}
\author[3]{Florent Krzakala}

\affil[1]{\small Departement d'Informatique, \'Ecole Normale Sup\'erieure, PSL \& CNRS}
\affil[2]{\small Statistical Physics of Computation Laboratory, \'Ecole Polytechnique F\'ed\'erale de Lausanne (EPFL)}
\affil[3]{\small Information, Learning and Physics Laboratory, \'Ecole Polytechnique F\'ed\'erale de Lausanne (EPFL)}

\title{Scaling Laws and Spectra of Shallow Neural Networks \\ in the Feature Learning Regime}

\date{}

\begin{document}

\maketitle
\begin{abstract}
Neural scaling laws underlie many of the recent advances in deep learning, yet their theoretical understanding remains largely confined to linear models. In this work, we present a systematic analysis of scaling laws for quadratic and diagonal neural networks in the feature learning regime. Leveraging connections with matrix compressed sensing and LASSO, we derive a detailed phase diagram for the scaling exponents of the excess risk as a function of sample complexity and weight decay. This analysis uncovers crossovers between distinct scaling regimes and plateau behaviors, mirroring phenomena widely reported in the empirical neural scaling literature. Furthermore, we establish a precise link between these regimes and the spectral properties of the trained network weights, which we characterize in detail. As a consequence, we provide a theoretical validation of recent empirical observations connecting the emergence of power-law tails in the weight spectrum with network generalization performance, yielding an interpretation from first principles.
\end{abstract}

\section{Introduction}
\label{sec:intro}

A central development in modern deep learning has been the recognition that neural network generalization does not improve unboundedly when training data, model size, or compute are scaled in isolation. Instead, extensive empirical evidence reveals the presence of performance bottlenecks unless these resources are increased together \citep{kaplan2020scaling,brown2020language,hoffmann2022empirical}. Characterizing these trade-offs, and in particular predicting the resulting \emph{neural scaling laws}, has emerged as a fundamental challenge for deep learning research, with significant implications for the design of efficient and resource-conscious models.

Our goal in this work is to investigate this question in the context of shallow neural networks. More precisely, consider the following supervised empirical risk minimization (ERM) problem for the class of two-layer neural networks $f(\vx;\mW,\va)=\va^{\top}\sigma(\mW\vx+\vb)$:
\begin{align}
\label{eq:def:erm:full}
    \underset{\mW,\va}{\min} \sum\limits_{\mu=1}^{n} \left(y_{\mu}-f(\vx_{\mu};\mW,\va)\right)^{2} + \lambda\left(||\mW||_{\rm F}^{2}+||\va||^{2}_{2}\right)
\end{align}
where $\mW\in\mathbb{R}^{p\times d}$ and $\va\in\mathbb{R}^{p}$ are the first- and second-layer weights, respectively. Although substantial progress has been achieved in recent years, our current understanding of scaling laws for the generalization performance of the ERM minimizer in \cref{eq:def:erm:full} remains largely confined to the random features regime \citep{bahri2024explaining,maloney2022solvable,paquette2024four,atanasov2024scaling,bordelon2024dynamical,kunstner2025scaling}. In this setting, the problem reduces to a kernel method, where scaling behavior has been classically studied, and is known as \emph{source and capacity conditions} \citep{caponnetto2007optimal,cui2021generalization,defilippis2024dimension}. 

In this work, we move beyond the random features regime and investigate neural scaling laws for the 
ERM minimizer in \cref{eq:def:erm:full} in the teacher-student setting. That is, we assume that the target task is generated by a teacher network of the same architecture
\begin{align}
    y_{\mu} = f(\vx_{\mu};\mW^{\star},\va^{\star})+\sqrt{\Delta}\xi_{\mu},
    \label{teacher}
\end{align}
where $\{\vx_\mu\}_{\mu=1}^n\overset{i.i.d.}{\sim}\mathcal{N}(0,I_d)$ denotes the dataset and $\{\xi_\mu\}_{\mu=1}^n\overset{i.i.d.}{\sim}\mathcal{N}(0,1)$ is an additive Gaussian label noise with variance $\Delta\geq0$. The statistics of the target weights $\mW^{\star},\va^{\star}$ will be specified later. Our main goal in this paper is to characterize the scaling-law and bottleneck behaviors of the excess risk
\begin{align}\label{eq:def:excess_risk}
    R(\mW,\va) = \mathbb{E}_{\vx\sim\mathcal{N}(0,\bI_{d})}\left[(f(\vx;\mW^{\star},\va^{\star})-f(\vx;\mW,\va))^{2}\right]
\end{align}
associated with the minimizers $\hat{\mW},\hat{\va}$ of \cref{eq:def:erm:full}.  Note again that, at variance with recent works (e.g. \citep{ren2026emergence,arous2025learning}), we shall {\it not} consider the learning dynamics here, and focus on the actual minimizer of the ERM. In the over-parametrized setting that we consider, this makes sense as the landscape of the minimization problem will be easy to optimize over \citep{venturi2019spurious}.

We will consider two specific classes of shallow neural networks. Thanks to exact mappings to classical problems in signal processing, these models admit a mathematical characterization, enabling an end-to-end analysis of the optimization problem in the feature learning regime.

\paragraph{Diagonal networks and LASSO.}  
    The first architecture is a diagonal neural network with $p=d$, diagonal first-layer weights $\mW = \operatorname{diag}(\vw)$, linear activation and no bias (${\boldsymbol b}=0$): 
    \begin{equation}
    \label{eq:def:diagonal_network}
        f(\vx;\mW,\va) = {\boldsymbol a}^{\top}\frac{({\boldsymbol w}\odot {\boldsymbol x})}{\sqrt{d}} \, .
    \end{equation}
    While the expressivity of this architecture is the same as that of a linear model, the reparameterization creates an effective implicit regularization that allows for feature selection and has made this setting popular among theoreticians. Indeed, adapting an argument by \cite{neyshabur2015norm,soudry2018implicit,pesme2023saddle} (see Appendix~\ref{equivalence}), the resulting empirical minimization problem is equivalent to the LASSO problem with parameters $\theta_i = a_i w_i / \sqrt{d}$ and objective
    \begin{align}\label{eq:def:lasso}
        \hat{\boldsymbol \theta} = \underset{{{\boldsymbol \theta}\in\R^d}}{\arg\min} 
        \frac{1}{2}\sum_{\mu=1}^{n} \big(y_\mu- {\boldsymbol \theta}^{\top} {\boldsymbol x}_{\mu}\big)^2
        + \lambda ||\boldsymbol{\theta}||_{1} \, .
    \end{align}
    In other words, the ERM problem for a diagonal two-layer linear network trained with $\ell_2$ weight-decay can be understood through the performance of LASSO.

\paragraph{Quadratic neural network and matrix compressed sensing.}  
    The second architecture is that of an over-parameterized two-layer network with a (centered) quadratic activation,
    \begin{align}\label{eq:erm:quad}
    f(\vx;\bW,\va) = \frac{1}{\sqrt p} \sum_{j=1}^{p}
    \left(\left(\frac{{\boldsymbol w}_{j}^{\top}{\boldsymbol x}}{\sqrt{d}} \right)^2 - \frac{||{\boldsymbol w}_{j}||^{2}_{2}}{d}\right) 
    = {\rm Tr}\!\left[\bS \, \frac{{\boldsymbol x}{\boldsymbol x}^{\top} - \bI_{d}}{\sqrt d}\right]\, , 
    \end{align}
    where $\bS:=\frac{\bW^{\top}\bW}{\sqrt{pd}}\in\mathbb{R}^{d\times d}$ and the normalization is taken for convenience. In this case, we fix the second-layer weight $\boldsymbol{a}$ of the model to be an all-one vector, but the target network may have arbitrary second layer weights. This class of quadratic neural networks have recently gained in popularity as simple models for non-convex tasks \citep{sarao2020optimization,arnaboldi2023escaping,martin2024impact,arous2025learning}. The ERM problem in \cref{eq:def:erm:full} for this architecture can be mapped to a sparse estimation setting \citep{gunasekar2017implicit,maillard2024fitting,bandeira2025exact,    xu2025fundamental,erba2025nuclear}, namely matrix compressed sensing (or low-rank matrix estimation): 
    \begin{align}\label{eq:def:quadratic_network}
    \hat \bS = \underset{\bS \succeq 0}{\arg\min} \sum_{\mu=1}^{n} \left(y_\mu-\Tr[\bS\bZ_{\mu}]\right)^2 + \lambda \|\bS\|_* \, ,
    \end{align}
where $\bZ_{\mu} := \tfrac{{\boldsymbol x_{\mu}}{\boldsymbol x_{\mu}}^{\top} - \boldsymbol{I}_{d}}{\sqrt d}$, $\|\cdot\|_*$ denotes the nuclear norm and we rescaled $\lambda \to \lambda / \sqrt{pd}$. We refer again to Appendix~\ref{equivalence} for the explicit mapping. Thus, the performance of a quadratic network trained with weight decay can be analyzed via low-rank matrix estimation with nuclear norm regularization.

These two equivalences underline the central theme of this work: by mapping neural network training problems to sparse vector and matrix estimation tasks, we can leverage the rich theoretical toolbox developed for LASSO and compressed sensing, and in particular approximate message passing and its high-dimensional state evolution \citep{donoho2009message,donoho2013phase,javanmard2013state,berthier2020state,erba2025nuclear}.
This bridge not only enables precise predictions for generalization error and scaling exponents, but also provides a principled understanding of the weight spectral distribution in neural networks.

\paragraph{Power-law/quasi-sparse targets.} 
To study scaling behavior, we adopt the classical assumption of a target with a power-law spectrum, as considered for instance in \citep{caponnetto2007optimal,steinwart2009optimal,spigler2020asymptotic,cui2021generalization,bordelon2024dynamical,arous2025learning}. In the language of compressed sensing, this corresponds to the notion of {\it quasi-sparsity} \citep{negahban2011estimation,raskutti2011minimax}, where the signal is not exactly sparse but its coefficients decay according to a heavy-tailed distribution. This makes the setting natural and relevant to both the machine learning and signal processing communities.  Concretely, in the case of diagonal linear networks we assume effective weights  
\begin{align}
    \theta^{\star }_{i}\overset{i.i.d.}{\sim}\mathcal N(0,\, d\, i^{-2\gamma}) \, ,
    \qquad \vtheta^\star := \mW^\star \va^\star \, ,
\end{align}
while for quadratic neural networks we assume that  
\begin{align}
    \bS^\star := \frac{1}{\sqrt{pd}} \sum_{j=1}^p a^{\star}_j\, \bw^\star_{j}(\bw^\star_{j})^T
\end{align}
is rotationally invariant with eigenvalues $\{\sqrt{d}\, i^{-\gamma}\}_{i=1}^d$. This setting was recently studied as well in \cite{arous2025learning} (who considered  noiseless target ($\Delta=0$) \update{and obtain one of our scaling exponents}). In both cases we fix $\gamma > 1/2$ to ensure square-summability of $\vtheta^\star$ and $\bS^\star$.

\subsection{Main Results} 
\begin{enumerate}

\item \textbf{Phase diagram and complete characterization of excess risk rates for power-law targets.}  
We provide a sharp characterization of the excess risk achieved by empirical risk minimization \eqref{eq:def:erm:full} for both diagonal linear networks and quadratic networks in the regime $n,d \gg 1$ with $p \geq d$, under a power-law design for the target function and varying regularization strength $\lambda$, summarized in \Cref{fig:phase}. Our results uncover a striking universality between the two settings, including a transition from {\it benign} to {\it harmful overfitting}. Exploring the extent of this universality beyond the setting here is an interesting avenue for future work. We also derive the risk rates under optimal regularization $\lambda$, and show that optimally-regularized ERM achieves the Bayes-optimal rates --- previously known only for the diagonal case \citep{raskutti2011minimax}. These findings are of independent interest for sparse vector and low-rank matrix estimation.

\item \textbf{Spectral behavior of the learned weights.}  
We characterize, across all phases, the spectral properties of the trained network weights. The learned spectrum reflects the implicit trade-off between signal, noise, and regularization, and exhibits phenomena directly connected to feature learning. Remarkably, the resulting spectral behavior mirrors observations in modern deep learning practice \citep{martin2021implicit,thamm2024random}.
\item \textbf{First-principles explanation of spectra–generalization connection.}  We provide a clear interpretation of the spectrum and its relation to generalization. Building on Result~\ref{res:decomposition}, which decomposes the error into {\it underfitting, overfitting, and approximation} terms, we show that each of these components is directly connected to the spectral statistics of the weights. In doing so, we provide a mathematical theory for the empirical observations of \cite{martin2021predicting} \update{and \cite{wang2023spectral}} for the spectral statistics of weights in large-scale trained networks.

\item \textbf{Non-asymptotic validity of state evolution.}  
Our derivations rely on approximate message passing (AMP) and its state evolution equations, which are rigorously valid only in the proportional asymptotic regime with fixed ratios $n/d$ (or $n/d^2$) and fixed $\lambda$. We extend these equations heuristically beyond their proven setting, covering arbitrary scalings of $n,d,\lambda$.  
Through extensive numerical experiments, we demonstrate that state evolution remains accurate down to constants across the whole parameter space, including far beyond proven guarantees. This surprising robustness, already established in ridge regression \citep{cui2021generalization,cheng2024dimension,misiakiewicz2024non,defilippis2024dimension}, suggests a broader conjecture: the AMP framework, and related tools from spin glass theory, may provide predictive power well outside their standard asymptotic assumptions. We hope this will further motivate work on non-asymptotic control of AMP \citep{rush2018finite,miolane2021distribution,li2022non,reeves2025dimension}.

\end{enumerate}

We also refer to the Github code repository \url{https://github.com/SPOC-group/QuadraticNetPowerlaw} for reproducibility. Together, these results provide a comprehensive theoretical and empirical understanding of scaling laws for feature learning in simple network models.

\subsection{Further Relevant work} 

\paragraph{Scaling laws  ---}  
   A large body of work has studied scaling laws in the lazy regime, where the features remain fixed. This includes kernel methods \citep{caponnetto2007optimal,spigler2020asymptotic,cui2021generalization}, random features \citep{defilippis2024dimension,atanasov2024scaling,bahri2024explaining,maloney2022solvable,paquette2024four,kunstner2025scaling}, and neural tangent kernels (NTK) \citep{bordelon2020spectrum}. \cite{bordelon2024dynamical, bordelon2025feature} analyzed how scaling laws change for linear networks when both weights are trained\update{, and \cite{worschech2025analyzing} explicitly solves the dynamics of a linear network to obtain the scalings}. Our goal in this work is to go beyond linear networks and the lazy regime and analyze scaling laws in the presence of genuine feature learning. 
   Two recent works \citep{ren2026emergence,arous2025learning} analyze settings related to ours, but with important differences. Both consider two-layer networks with sublinear width, orthogonal first-layer weights, and power-law decaying second-layer weights. \citet{ren2026emergence} study activation functions with large information exponents, which is orthogonal to our setting, while \citet{arous2025learning} focus on quadratic activations with  a specific SGD dynamics. Both works, additionally, consider noiseless targets and unregularized training ($\lambda = \Delta = 0$). Here, we study empirical risk minimization with weight decay $\lambda>0$ and noise $\Delta\geq0$.

\paragraph{Spectral properties of learned weights ---}  
A growing literature investigates the distribution of weight spectra in trained neural networks, with particular attention to the emergence of heavy-tailed eigenvalue distributions in both weights and activations \citep{mahoney2019traditional,martin2021predicting,martin2021post,martin2021implicit,thamm2024random,wang2023spectral,zhou2023temperature,hodgkinson2025models}. Despite these empirical observations, a precise theoretical characterization of the learned spectra and their relation to generalization remains elusive. Recent progress includes analyses of the spectrum after a single \update{or a few gradient steps \citep{dandi2024two,moniri2023theory,cui24Asymptotics,dandi2024random,kothapalli2025spikes}}, results showing convergence of SGD in mean-field models to spectral distributions reminiscent of those we obtain \citep{olsen2025sgd} and a semi-empirical theory on the relation between spectra and generalization propertiees \citep{martin2025setol}. Our description of the spectrum of the trained weight provides an analytic characterization of this phenomenon, and provides an interpretation of these properties from first principles.

\paragraph{AMP and State Evolution ---}  Our analysis relies on approximate message passing (AMP) and its state evolution (SE), which has become a central tool for studying high-dimensional inference problems with structure \citep{donoho2013phase,bayati2011dynamics,javanmard2013state,berthier2020state,zou2022concise,feng2022,gerbelot2023graph,dudeja2023universality,erba2025nuclear}. It has also been applied to learning problems beyond sparse recovery, such as kernel methods and learning rates \citep{cui2021generalization,loureiro2021learning}. In this work, we use the state evolution equations of AMP heuristically, to analyze quasi-sparse models {\it beyond} their rigorously proven asymptotic regimes (typically assuming a fixed ratio $n/d$). While recent advances in non-asymptotic control \citep{rush2018finite,miolane2021distribution,li2022non,reeves2025dimension} provide reassurance, 
a finer control of the limit is still required for a fully rigorous justification. Our experiments nevertheless show excellent agreement between SE predictions and numerical results across regimes, suggesting that AMP may be predictive well beyond its standard assumptions.  \looseness=-1

\paragraph{Compressed sensing ---}  
Quasi-sparse settings, where coefficients decay with a power law in Fourier or wavelet bases, have long been studied in statistics and signal processing. This is natural  since most real-world signals are not exactly sparse but have heavy-tailed coefficient distributions \citep{mallat1999wavelet}. Classical work on LASSO and matrix compressed sensing analyzed $\ell_p$-controlled targets, deriving minimax bounds on error and sample complexity \citep{raskutti2011minimax,negahban2011estimation}.  Our results extend this line of work by providing the full phase diagram across all regularization strengths and data scales. For instance, the optimal LASSO rate of \citet{raskutti2011minimax} arises from setting $\lambda=\Tilde{\Theta}(\sqrt{n/d})$ (here $\Tilde{\Theta}$ is up to logarithmic factors).

\subsection*{Notation}
We denote matrices $\bW\in\mathbb{R}^{p\times d}$ and vectors $\va\in\mathbb{R}^{p}$ by bold upper- and lower-case letters, respectively. $\mathcal{N}(\mu,\Sigma)$ denote the Gaussian distribution with mean $\mu$ and covariance $\Sigma$. We denote the Gaussian orthogonal ensemble as $\GOE(d)\subset\mathbb{R}^{d\times d}$, that is, if $\bZ\in\GOE(d)$ we have $Z_{ij}=Z_{ji}\sim\mathcal{N}(0,1/d)$ for $1\leq i<j\leq d$ and $Z_{ii}\sim\mathcal{N}(0,2/d)$ for $1\leq i\leq d$. $\boldsymbol{1}_{A}(x)$ denote the indicator function on the set $A$, i.e. $\boldsymbol{1}_{A}(x)=1$ if $x\in A$ and zero otherwise. We employ the standard big-O notation: We write $a_{d} = O(b_{d})$ if there exists a constant $C>0$ and an integer $d_{0}\geq 1$ such that $|a_{d}| \leq C\, b_{d}$ for all $d \geq d_{0}$ and $a_{d} = o(b_{d})$ if $\lim_{d\to\infty} \sfrac{a_{d}}{b_{d}} = 0$. We also denote $a_{d} = \Theta(b_{d})$ or $a_d\asymp b_d$, if both $a_{d} = O(b_{d})$ and $b_{d} = O(a_{d})$, i.e.\ there exist constants $C_{1},C_{2}>0$ and $d_{0}\geq 1$ such that $C_{1} b_{d} \leq |a_{d}| \leq C_{2} b_{d}$ for all $d \geq d_{0}$. Sometimes, we omit the $d$ subscript or write $b = O_{d}(a)$ when the dependence is implicit. We denote $\tilde{O},\tilde{\Theta}$ for $O, \Theta$ up to logarithmic factors, i.e. $\tilde{\Theta}(a_{d})=\Theta(a_{d}\log{d})$.

\section{Main results}

\subsection{Universal error rates}\label{sec:phase_diagrams}

In this section we discuss the excess risk rates associated to the two problems introduced above. Our analysis is based on a deterministic characterization of the risk $\hat{R}(\hat{\bW},\hat{\va})\simeq \mathsf{R}_{n,d}$ at large $n,d\gg 1$, which is discussed in \Cref{sec:nonasymp}. 
We only consider the noisy case ($\Delta>0$), and refer to Appendix \ref{app:noiseless} for the noiseless case.
In order to highlight the correspondence between the two neural network models, we express the results in terms of the effective sample size $\neff$ as follows:
\begin{align}
    \neff \equiv \begin{cases}
        n & \text{ for diagonal network} \\
        n/d & \text{ for quadratic network}
    \end{cases} \, .
\end{align}
Surprisingly, this definition will be enough to present both cases in a unified manner.
\begin{res}[Excess risk rates]
\label{res:noisy}
    Under the setting of Sec.~\ref{sec:intro} for $\Delta > 0$ and $n_{\rm eff},d \gg 1$, the excess risk satisfies
    \begin{align}\label{eq:rates_ERM_noisy}
\mathsf{R}_{n_{\rm eff},d}(\lambda)=
\begin{cases}
    \Theta\left(\neff^{-1+1/(2\gamma)}+\rho\left(\neff/d\right)\right)
    &\text{if}\quad 1\ll\neff\ll d \mathand 
    \lambda\ll\sqrt{\frac{\neff}{d}}
    \\
    \Theta\left(\lambda^{-2/3}\right)&\text{if}\quad \neff\sim d\mathand\lambda\ll1\\
    \Theta(d/\neff)
    &\text{if}\quad \neff\gg d \mathand 
    \lambda\ll\sqrt{\frac{\neff}{d}}
    \\
    \Theta\left(\left(\lambda d^{1/2}/\neff \right)^{2-1/\gamma}\right)
    &\text{if}\quad 
    \max\left(\sqrt{\frac{\neff}{d}},\frac{\neff}{d^{\gamma+1/2}}\right)\ll\lambda\ll\frac{\neff}{d^{1/2}}
    \\
    \Theta\left(\lambda^2 d^2 / \neff^2\right)
    &\text{if}\quad 
    \sqrt{\frac{\neff}{d}}\ll\lambda\ll\frac{\neff}{d^{\gamma+1/2}}
\end{cases}\, ,
\end{align}
and $\mathsf{R}_{n_{\rm eff},d} = \Theta(1)$ otherwise,
where $\rho(t) = -1/\log(t)$ in the diagonal network case and $\rho(t) = t^{2/5}$ in the quadratic network case. Notice that in both cases the $\rho$ term is monotone increasing with $\neff/d$, and dominates the error rate when $\neff \to d$. 
Additionally, in the diagonal network case, the first rate holds up to logarithmic factors that we specify in eq.~\eqref{app:eq:lasso_scaling_laws_noisy} in Appendix \ref{app:sec:lasso_rates}.
\end{res}
These rates  are summarized in \Cref{fig:phase}. For small (fixed) regularization $\lambda < 1/\sqrt{d}$, with $d$ fixed and $\neff$ increasing, the excess error moves from an initial plateau (\textcolor{phaseI}{Phase Ia}), driven by data scarcity, to a fast–decay (\textcolor{phaseIV}{Phase IV}), where  $\mathsf{R}_{n_{\rm eff},d} =\Theta(\neff^{-1 + 1/(2\gamma)})$, matching the minimax rate in \citep{raskutti2011minimax,donoho2011compressed}. 
As $\neff$ approaches $d$, the estimator begins to fit the noise, and we observe a harmful overfitting (\textcolor{phaseV}{Phase V}), in which the excess risk is dominated by the non-universal scale $\rho$ (arising from overfitting the noise as in Result~\ref{res:decomposition}). This transition, characteristic of the under-regularized and under-sampled regime ($1\ll\neff\ll d$, $\lambda\ll\sqrt{\frac{\neff}{d}}$), happens at 
\begin{align}
    \neff^{\rm cross}
     = \begin{cases}
        (\log d)^{\frac{4\gamma-1}{2\gamma-1}} & \text{ for diagonal network} \\
        d^{\frac{4\gamma}{14\gamma-5}} & \text{ for quadratic network}
    \end{cases} \, .
\end{align}
The excess risk reaches its maximum around $\neff \sim d$ with $\mathsf{R}_{n_{\rm eff},d} \sim \lambda^{-2/3}$. This non-monotonicity of the risk at interpolation is reminiscent of the double descent behavior \citep{belkin2019reconciling,mei2022generalizationB}\update{, and extends previous findings \citep{bartlett2020benign,wang2024near} to non-linear models}. For $\neff \gg d$, the excess risk then enters a second fast–decay \textcolor{phaseVI}{Phase VIa}, with rate proportional to $d/\neff$; this is the fastest decay we observe, provided $\neff \gg d^{2\gamma}$ (\textcolor{phaseVI}{Phase VIb}). 
For larger regularization strength $\lambda >1/\sqrt{d}$, the excess risk decay is described by the upper part of the phase diagram in Figure \ref{fig:phase}, eventually crossing to the lower part when $\neff \sim d\lambda^2$. In particular, if $\lambda\gg d^{\gamma-1/2}$, we observe that increasing $\neff$, the excess risk is initially in a plateau (\textcolor{phaseI}{Phase Ib}), induced by the strong regularization with respect to the sample size. For $n\sim \lambda d^{1/2}$, it crosses into a slow rate \textcolor{phaseII}{Phase II} with $\mathsf{R}_{n_{\rm eff},d}=\Theta(\lambda d^{1/2}/\neff)^{2-1/\gamma}$, which transitions to a faster rate (\textcolor{phaseIII}{Phase III}), still influenced by the large regularization, for $\neff\sim \lambda d^{\gamma + 1/2}$, with $\mathsf{R}_{n_{\rm eff},d}=\Theta((\lambda d/\neff)^2)$. \textcolor{phaseII}{Phase II} recovers the rate in \cite[Corollary 2]{negahban2011estimation}. The excess risk eventually transitions to the fast-decay \textcolor{phaseVI}{Phase VIb} for $\neff \sim \lambda^2 d$, where the effect of the regularization becomes negligible due to the large sample size.
These cross-overs are reminiscent of the ones observed for kernel and random feature ridge regression respectively in \cite{cui2021generalization,defilippis2024dimension}.

There exists a narrow boundary layer around these \textcolor{red}{red} lines with sharp error changes, see Appendix \ref{app:noisy_ERM} for more details. At the boundary $\neff=\Theta(d)$, we observe the aforementioned crossover between harmful overfitting and fast decay, with an interpolation peak emerging. Similarly, at the critical regularization $\lambda=\Theta(\sqrt{\neff/d})$ and for $\neff^{\rm cross}\ll \neff \ll d^{2\gamma}$ 
the excess risk jumps from $\neff^{-1+1/(2\gamma)}$ to much higher value when reducing the regularization. 
\begin{figure}[t]
\input{img/noisy_new}
\caption{Excess risk rates of Result \ref{res:noisy} as a function of $n$ and $\lambda(n,d)$, with a sketch of the corresponding spectral properties of the learned weights (Result \ref{res:structure}). \update{Red lines represent sharp phase boundaries with different rates on two sides.}}
\label{fig:phase}
\end{figure}

As a corollary of the above results,  we can immediately estimate the behavior of the optimal regularization $\lambda_{\rm opt}$ and the associated optimal ERM rates.
\begin{cor}[Optimal regularization and optimality of ERM]
\label{cor:optlambda}
    The optimal regularization satisfies
    \begin{align}
\lambda_{\rm opt}(n_{\rm eff},d) = \begin{cases}
O(\sqrt{\neff/d}),&\text{if }\, \Delta > 0\,\text{ and }\, \left(1\ll\neff\ll \neff^{\rm cross}\,\text{ or }\, \neff\gg d^{2\gamma}\right)\\
\Tilde{\Theta}(\sqrt{\neff/d}),&\text{if }\, \Delta> 0\,\text{ and }\, \neff^{\rm cross}\ll\neff\ll d^{2\gamma}
\end{cases} \, .
\end{align}
where $\tilde{\Theta}$ is up to logarithm factors in the argument. The excess risk rates in the optimally regularized case matches the large $n_{\rm eff},d\gg 1$ the Bayesian risk $R_{\rm BO}(\mathcal D) = \E[R(\mW,\va)|\mathcal D]\simeq \mathsf{R}_{n_{\rm eff,d}}^{\rm BO}$\footnote{See \cref{sec:nonasymp} for a formal statement.} (for the diagonal network, up to logarithmic factors), given by
\begin{align}
    \mathsf{R}_{n_{\rm eff},d}(\lambda_{\rm opt}) = \Theta(\mathsf{R}_{n_{\rm eff}}^{\rm BO})=
    \begin{cases}
         \Theta(\neff^{-1+1/(2\gamma)})
        &\mathif \Delta > 0 \mathand 1\ll \neff\ll d^{2\gamma}
        \\
        \Theta\left(d/\neff\right)
        &\mathif \Delta > 0 \mathand  \neff\gg d^{2\gamma}
    \end{cases} \, ,
\end{align}
and $ \mathsf{R}_{n_{\rm eff},d} = \Theta(1)$ otherwise.
Again, in the diagonal network case, in the first regime the rate holds up to logarithmic factors that we specify in eq.~\eqref{app:eq:lasso_scaling_laws_noisy} in Appendix \ref{app:sec:lasso_rates}.
\end{cor}
Corollary \ref{cor:optlambda} shows that by appropriately tuning the regularization allows to avoid the harmful overfitting phase in the noisy case and reach Bayesian optimality. 
Interestingly, the noisy rate $\Theta(\neff^{-1+1/(2\gamma)})$ in the regime $1 \ll n_{\rm eff} \ll d^{2\gamma}$ coincides with the classical minimax rate for high-dimensional linear regression over an $\ell_{q}$-ball with $q = \sfrac{1}{\gamma}$ \citep{raskutti2011minimax,donoho2011compressed}.   \Cref{cor:optlambda} not only recovers the well-known result that properly regularized LASSO achieves this minimax rate, but also extends it to additional regimes and to the matrix case, revealing a cross-over between the minimax rate and a faster $\Theta(d/n_{\rm eff})$ rate. 

\subsection{Spectra of the learned weights} 
\label{sec:spectral_properties}
\begin{figure}[t]
\centering
\includegraphics[width=\linewidth]{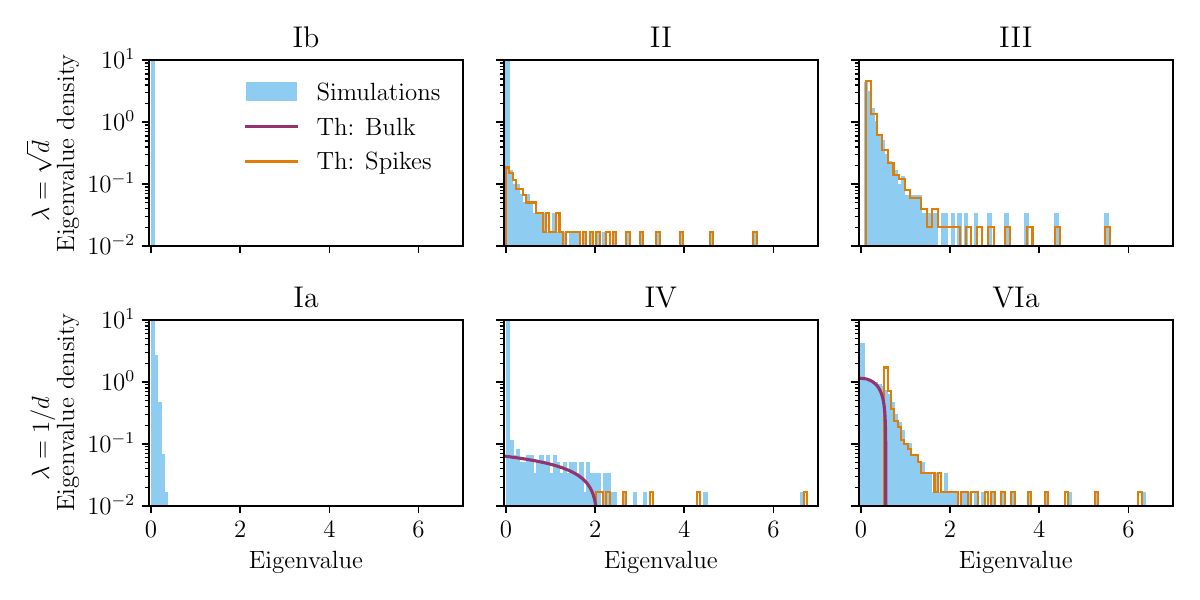}
\caption{Comparison between spectra from simulations and theory across different training phases. Blue: eigenvalue histograms after training. Purple/orange: theoretical predictions for bulk and spikes, respectively \eqref{eq:quadr_spect} (for clarity, spike histograms are shown separately). Notice that our theory also predicts a spike at zero, which we do not plot for visual clarity. All panels use $d\!=\!800$ except III, for which we have $d\!=\!400$. Bottom row: $\lambda\!=\!1/d$ with $n\!=\!100,\;1.94\times10^4,\;1.28\times10^6$. Top row: $\lambda\!=\!\sqrt{d}$ with $n=800,\;6.4\times10^6,\;2\times10^7$. We discuss the phenomenology in Section \ref{res:interpretability}.}
\label{fig:spectra}
\end{figure}
Our second set of results concerns the structural properties of the learned weights, that are given by a soft thresholding function applied to a noisy version of the target's weights. Notice that for diagonal neural networks, the weights $\vtheta$ can be seen as a diagonal matrix (modulo a sign), hence they coincide with the eigenvalues of $\bW$.
\begin{res}[Spectrum of the learned weights]
\label{res:structure}
    For the diagonal network case, there exists constants $\delta(n,d,\lambda)$ and $\epsilon(n,d,\lambda)$ (specified in Appendix \ref{app:sec:lasso_spectral}) such that the empirical risk estimator \eqref{eq:def:erm:full} satisfies (in distribution)
    \begin{align}
        \hat{\theta}_i \sim \sigma_{\rm d}( 
            \theta^\star_i + \delta z_i; \epsilon) \, ,
    \end{align}
    where $z_i \!\sim\! \mathcal{N}(0,1)$, and $\sigma_{\rm d}(x;a) \!=\! \max(x-a,0) \!-\! \max(-x-a,0)$ is the soft-thresholding function.
    For the quadratic network case, there exists constants $\delta(n,d,\lambda)$ and $\epsilon(n,d,\lambda)$ (that are obtained from \eqref{eq:SE_ERM}) such that the spectrum $\nu$ of the empirical risk estimator \eqref{eq:def:quadratic_network} satisfies
    \begin{align}\label{eq:quadr_spect}
    \nu(x)=F_{\mu_\delta}(\lambda\epsilon)\boldsymbol{\delta}_0(x)+\mu_\delta(x+\lambda\epsilon)\mathbf{1}_{x>0}.
    \end{align}
    $\boldsymbol{\delta}_0$ represents a Dirac mass at $0$, $\mathbf{1}_{A}$ is the indicator function of the set $A$ and $\mu_\delta$ represents the spectrum of $\bS^{\star}+\delta\bZ$ with its cumulative function $F_{\mu_\delta}$, where $\bZ\sim\GOE(d)$ (i.e. a symmetric matrix with $\mathcal{N}(0,1/d)$ elements up to symmetry).
\end{res}
Result~\ref{res:structure} characterizes the learned weights in both settings: they are {\it noisy, soft-thresholded versions of the target spectrum}.  The parameter $\delta$ quantifies the noise from the label noise and finite sample estimation of the target weights, while  $\lambda\epsilon$ sets the cutoff below which singular values vanish due to regularization.  
For any $n,d,\lambda$, the spectrum consists of a spike at zero, possibly a bulk near zero, and a few outliers aligned with the top eigenvectors of the target.\looseness=-1

\subsection{Interpretability, and a ``universal'' error decomposition}
\label{res:interpretability}

The spectra depends on $n,d,\lambda$ only through the functions $\delta,\epsilon$, leading to a qualitative structure shared by both models. Our theory predicts eight distinct spectral phases (Figure~\ref{fig:phase}) which are closely connected to the risk rates in Result~\ref{res:noisy}. Focusing now on the quadratic network, the result provides an interpretation of the risk in terms of the weights spectrum.
\begin{res}[``Universal'' error decomposition of feature learning]
\label{res:decomposition}
Let $\{s_i\}_{i=1}^d$ of $\bS^{\star}$ be the eigenvalues of $\bS^{\star}$ in non-increasing order. Consider the following two cases. 

(i): Under-regularization. Assume that the constants $\delta(n,d,\lambda)$ and $\epsilon(n,d,\lambda)$ in Result~\ref{res:structure} satisfy $\lambda\epsilon<2\delta$ and there exists a cutoff $K(\delta)\ll d$ satisfying $s_{K(\delta)}=\delta$. Then the excess risk reads
\begin{align}
\mathsf{R}_{n,d}\!=\underbrace{\delta^2\!\!\int_{\lambda\epsilon/\delta}^2\!\!\!\!\!\mu_{\rm sc}({\rm d} x)\left(x-\frac{\lambda\epsilon}{\delta}\right)^2+\frac{1}{d}\delta K'(\delta)(2\delta-\lambda\epsilon)^2}_{\substack{\text{overfitting} \\ \text{(learned noise)}}}
\!\!+\!\!\!\!\!\!\!\!\!\!\!\!\underbrace{\!\!\!\!\sum_{i=K(\delta)+1}^d\!\!\!\frac{s_i^2}{d}\!\!\!}_{\substack{\text{underfitting} \\ \text{(not learned features)}}}
\!\!\!\!\!\!\!+\!\underbrace{\frac{1}{d}\sum_{i=1}^{K(\delta)}\left[\left(\frac{\delta^2}{s_i}-\lambda\epsilon\right)^2\!\!\!+\frac{\delta^2}{s_i}\left(s_i+\frac{\delta^2}{s_i}-\lambda\epsilon\right)\right]}_
{\substack{\text{approximation error} \\ \text{for learned features}}}
\label{eq:error_decomposition}
\end{align}
where $\mu_{\rm sc}({\rm d} x) = (2\pi)^{-1}\sqrt{4-x^{2}}\boldsymbol{1}_{x\in[-2,2]}{\rm d} x$ denotes the Wigner semi-circle law.

(ii): Over-regularization. Assume that the constants $\delta(n,d,\lambda)$ and $\epsilon(n,d,\lambda)$ in Result~\ref{res:structure} satisfy $\lambda\epsilon\geq2\delta$ and there exists a cutoff $K(\delta,\lambda\epsilon)\ll d$ satisfying $s_{K(\delta,\lambda\epsilon)}+\frac{\delta^2}{s_{K(\delta,\lambda\epsilon)}}-\lambda\epsilon=0$. Then 
\begin{align}
\mathsf{R}_{n,d}\!=\underbrace{\frac{1}{d}\!\!\!\sum_{i=K(\delta,\lambda\epsilon)+1}^ds_i^2}_{\substack{\text{underfitting} \\ \text{(not learned features)}}}
\!+\underbrace{\frac{1}{d}\sum_{i=1}^{K(\delta,\lambda\epsilon)}\left[\left(\frac{\delta^2}{s_i}-\lambda\epsilon\right)^2\!\!\!+\frac{\delta^2}{s_i}\left(s_i+\frac{\delta^2}{s_i}-\lambda\epsilon\right)\right]}_
{\substack{\text{approximation error} \\ \text{for learned features}}}.
\label{eq:error_decomposition2}
\end{align}
\end{res}
Equations (\ref{eq:error_decomposition}) and (\ref{eq:error_decomposition2}) have clear interpretations. The first component of the overfitting term corresponds to the second moment of the bulk spectrum, representing the power of the learned noise and thus quantifying the degree of overfitting. The second part of the overfitting term is proportional to the square of the bulk size, but always remains subdominant compared to the first term. As shown in Equation (\ref{eq:error_decomposition2}), the overfitting term diminishes with increasing regularization strength.

The underfitting term measures the mean power of the unrecovered spikes, indicating how many features are lost due to the cutoff $K(\delta,\lambda\epsilon)$ (which depends on the noise and regularization). The approximation error term reflects the average error in the learned spikes, which depends on the effective signal-to-noise ratio $\frac{s_i}{\delta}$ and the effective regularization $\lambda\epsilon$. Notably, when the effective noise $\delta$ is zero, the approximation error increases with regularization; conversely, when the regularization $\lambda$ is zero, the approximation error increases with noise $\frac{\delta}{s_i}$. In general, however, the approximation error arises from a non-trivial interplay between the effective noise and regularization.

This decomposition is “universal” in that it does not depend on the target spectrum, dataset size, or regularization, holds for all $\Delta\geq0$ and applies across different spectral phases. While derived here for quadratic networks, similar expressions hold as well for diagonal networks. See Appendix \ref{app:sec:lasso_spectral}. Extending it to broader architectures is an interesting direction for future work. 

Based on the error decomposition in Result~\ref{res:decomposition}, we give an interpretation of the rates in Result~\ref{res:noisy} in terms of the weight spectral properties (see Figure \ref{fig:phase} for illustrations and Figure \ref{fig:spectra} for experiments): the bulk corresponds to learned noise, the spikes hidden by the bulk are the unrecovered features, and the outliers are the learned features. This provides a mathematical theory from first principles for the observations of
\citep{martin2021implicit,martin2021predicting}, whose terminology (e.g. ``bleed-out'', ``rank collapse'', \ldots) we borrow in the following. We begin by the case of large regularization, considering an increasing number of samples.
\begin{itemize}
    \item \textbf{\textcolor{phaseI}{Ib (Rank collapse)}}: All eigenvalues are zero. Data scarcity and strong regularization imply that the ERM estimator is zero.  Result~\ref{res:decomposition} then gives 
    a trivial risk $\mathsf{R}=\Tr[(\bS^\star)^2]/d$. 
    \item \textbf{\textcolor{phaseII}{II (Outliers)}}: The spectrum contains approximately $N_{\rm out}=\left(\frac{4n}{\lambda d^{3/2}}\right)^{1/\gamma}$ outliers, while the remaining eigenvalues are zero. Spikes are shifted by $\approx\frac{\lambda d^2}{4n}$. In this regime, some features are learned with noise, while others are lost due to over-regularization. Result~\ref{res:decomposition} implies that the risk is determined by the number and shift of the spikes, yielding \(\mathsf{R} = \Theta\left(\frac{1}{d}\sum_{i \geq N_{\rm out}} (\sqrt{d}i^{-\gamma})^2\right) = \Theta\left((\lambda d^{3/2}/n)^{2-1/\gamma}\right)\), since the error from the shift is of the same order.
    \item \textbf{\textcolor{phaseIII}{III (Heavy-tail)}}: The spectrum is a perturbed version of the target spectrum with a heavy-tail \(\rho(x) \sim x^{-1-1/\gamma}\). Regularization shifts the bulk leftward by \(\approx\frac{\lambda d^2}{4n}\), yielding \(\mathsf{R}=\Theta\left((\lambda d^2/n)^2\right)\).
\end{itemize}
As more eigenvalues emerge from the spike at zero, more features are learned and the risk decreases. Strong regularization suppresses any spurious bulk of small eigenvalues, as well as some of the smaller spikes.
Consider now the case of small regularization.
\begin{itemize}
    \item \textbf{\textcolor{phaseI}{Ia (Rank collapse)}}: The spectrum resembles a small portion of a semi-circle law along with many zero eigenvalues. Perhaps surprisingly, the contribution of the bulk is negligible even for vanishing regularization. Neither features nor noise are learned, so the risk remains $\mathsf{R}=\Tr[(\bS^\star)^2]/d$.
    \item \textbf{\textcolor{phaseIV}{IV (Bulk + Outliers)}}: The spectrum exhibits \(N_{\rm out}=\left( \Delta d / 4n \right)^{-1/2\gamma}\) outliers and a bulk with eigenvalues of order \(\Theta\left((\Delta^5 d^2 / n)^{1/10}\right)\). As in \textcolor{phaseII}{Phase II}, the risk decreases as more spikes emerge from the bulk. Since the bulk contribution is sub-leading, Result~\ref{res:decomposition} implies that the risk scales as the average power of the unrecovered spikes: \(\mathsf{R}=\frac{1}{d}\sum_{i \geq N_{\rm out}} (\sqrt{d}i^{-\gamma})^2 = \Theta\left((d/n)^{1 - \frac{1}{2\gamma}}\right)\).
    \item \textbf{\textcolor{phaseV}{V (Bulk + Outliers)}}: Similar to \textcolor{phaseIV}{Phase IV}, but the risk is now dominated by the bulk of eigenvalues of order \(\Theta\left((\Delta^5 d^2 / n)^{1/10}\right)\). The ERM estimator approaches interpolation and begins to fit noise. Although the number of outliers $\left( N_{\rm out} = \Delta d / 4n \right)^{-1/2\gamma}$ continues to increase and the bulk range shrinks, the bulk's second moment grows. Altogether, Result~\ref{res:decomposition} implies the risk increases.
    \item \textbf{\textcolor{red}{Interpolation peak (Bulk + Outliers)}}: The spectrum is dominated by a large semi-circle bulk of order \(\Theta(\Delta^{2/3} \lambda^{-1/3})\). There may still be \(\Theta(\Delta^{2/3} \lambda^{-1/3} d^{-1/2})\) spikes if \(\lambda \gg d^{-3/2}\), but their contribution is negligible. Even if the model learns features, the noise is overwhelming, and Result~\ref{res:decomposition} implies the risk is dominated by the bulk second moment \(\mathsf{R}=\Theta(\lambda^{-2/3})\).
    \item \textbf{\textcolor{phaseVI}{VIa (Bulk + Bleed-out + Outliers)}}: 
    The spectrum contains \(\left(\Delta d / 4n\right)^{-1/2\gamma}\) outliers and a bulk of small eigenvalues of order \(\sqrt{\Delta d^2 / 4n}\). The smallest outliers merge at the bulk boundary, creating a {\it bleed-out} effect. The risk decreases as more outliers emerge. The spikes are perturbed by \(\Theta(\sqrt{\Delta d^2 / n})\), and Result~\ref{res:decomposition} implies the risk scales as \(\mathsf{R}=\Theta(d^2 / n)\), since this dominates over the unrecovered features.
    \item \textbf{\textcolor{phaseVI}{VIb (Heavy-tail)}}: As in  \textcolor{phaseIII}{Phase III}, the spectrum is a perturbed version of the target  (a \emph{heavy-tailed} bulk) with perturbations \(\Theta(\sqrt{\Delta d^2 / n})\). The risk decays with the perturbation as \(\mathsf{R}=\Theta(d^2 / n)\).
\end{itemize}
Therefore, for the under-regularized case, as the number of samples is increased, the bulk keeps shrinking as the spikes pop out. However, as the shape of the bulk and the number of zero eigenvalues changes, the risk changes non-monotonically. In other words, the model learns an increasing number of features, but the influence of noise leads to a non-monotonic behavior in the risk. Furthermore, Result~\ref{res:decomposition} shows that regularization only affects the first term of eq.~\eqref{eq:error_decomposition} and may increase the last two terms. Thus the optimal regularization strategy is to truncate the bulk, setting the first term to zero, leaving the second unchanged and minimally increasing the third. If the bulk contribution is negligible, weaker regularization may be chosen (i.e. in phases IV and VIb). This reasoning explains Corollary~\ref{cor:optlambda}.

\update{Finally, we should note that although phases \textcolor{phaseVI}{VIa} and \textcolor{phaseVI}{VIb} exhibit similar risk decay rates, only VIb achieves the Bayes-optimal rate. In the regime $d \ll n_{\rm eff} \ll d^2$, the optimal performance is reached in phase II with $\lambda=\sqrt{n_{\rm eff}/d}$. The corresponding spectral density shows a transition from outlier-dominated (zero eigenvalues+spikes) to heavy-tailed behavior, which supports the argument of \cite{martin2021predicting} that heavy-tailed spectra are associated with superior generalization.}

\subsection{Non-asymptotic state evolution}
\label{sec:nonasymp}

The results of Sections \ref{sec:phase_diagrams}, \ref{sec:spectral_properties} and
\ref{res:interpretability} build on the theory of state evolution and approximate message passing algorithms \citep{donoho2009message,javanmard2013state,gerbelot2023graph}, whose formal guarantees hold in the high-dimensional limit $\neff,d\to\infty$ with fixed ratio $\neff/d$ and constant strength $\lambda$. In this regime, state evolution allows to characterize the asymptotic risk and the spectrum of the weights in both the neural networks models under consideration, both for the empirical risk minimizer and for the Bayes-optimal estimator (see Appendices \ref{app:sec:diagonal_derivation_rates} and \ref{app:quadratic}). \looseness=-1

\begin{figure}[t]
\centering
\includegraphics[width=\linewidth]{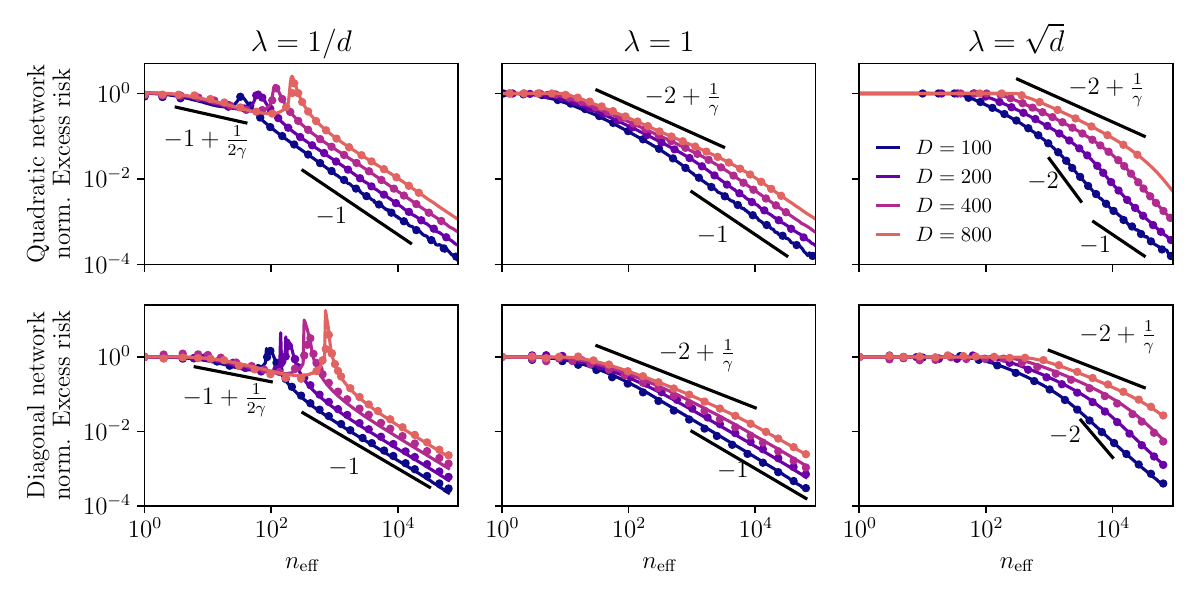}
\caption{Excess risk in simulations (dots, $d=100,200,400,800$) versus non-asymptotic state evolution (solid lines) as a function of $\neff$ ($\neff=n$ for the diagonal case, $\neff=n/d$ for the quadratic case) with $\lambda=1/d,1,\sqrt{d}$ and $\Delta=0.5$. The risk is rescaled such that it starts at $1$. We find excellent agreement, despite state evolution being rigorous only in the asymptotic limit $\neff/d=\Theta(1)$ with $d\gg1$. Black lines indicate the decay rates of the excess risk predicted by Result~\ref{res:noisy}, again showing good agreement. The networks are trained in PyTorch using LBFGS as detailed in Appendix \ref{app:numerical}.}
\label{fig:rates}
\end{figure}

Non-rigorous analyses in the ridge regression literature have employed asymptotic formulas to estimate excess risk rates under source \& capacity conditions, recovering classical results while also identifying new regimes in striking agreement with finite-size numerical experiments \citep{bordelon2020spectrum,cui2021generalization,simon2023eigenlearning}. The validity of these formulas beyond proportional asymptotics was subsequently established through non-asymptotic multiplicative bounds, thereby placing these rates on rigorous grounds \citep{cheng2024dimension,misiakiewicz2024non,defilippis2024dimension}. Motivated by this line of work, we derive our results under an analogous assumption: namely, that the state evolution equations for LASSO and matrix compressed sensing remain valid beyond proportional asymptotics. This assumption is supported both by extensive numerical evidence, depicted in \Cref{fig:rates} and Appendices~\ref{app:sec:diagonal_derivation_rates},~\ref{app:numerical}, and by rigorous results on the convergence rates for the LASSO state evolution \citep{miolane2021distribution}\footnote{\update{However, we need to extend their results to finite sample analysis \citep{rush2018finite}}}. 
\Cref{fig:rates} also confirms the theoretical decay rates of the excess risk across all phases, with state evolution and simulations in excellent agreement (see Appendix~\ref{app:numerical} for details on the implementation). Nonetheless, establishing non-asymptotic multiplicative bounds for LASSO and matrix compressed sensing remains a challenging open problem. Our results provide both motivation and supporting evidence for this direction, which we leave for future work.\looseness=-1
For conciseness, we only present the conjecture regarding quadratic networks, and refer to Appendix \ref{app:sec:diagonal_derivation_rates} for the conjecture concerning diagonal networks.
\begin{conjecture}
\label{conjecture:quadratic}
Let $\lambda>0$, $\Delta\geq0$ and consider $n,d\gg 1$ sufficiently large. Then with a probability at least $1-o_n(1)-o_d(1)$, both the excess risk associated to the empirical risk minimizer \cref{eq:def:quadratic_network} and the Bayes-optimal risk \update{(i.e. either $R_{\rm BO}$ or $R(\hat{\boldsymbol{W}},\hat{\boldsymbol{a}})$)} satisfy $
    |R - \mathsf R_{n,d}| = \mathsf R_{n,d}\cdot o_{n,d}(1)$. More precisely, for the Bayes-optimal case we have $\mathsf{R}^{\rm BO}_{n,d}=Q^\star-q$ with $Q^\star:=\frac{1}{d}\Tr[(\bS^\star)^2]$ and $q$ given by the fixed point of the following equation 
\begin{align} 
\hat{q}=\frac{4n/d^2}{\Delta+2(Q^\star -q)}, && 1-2\alpha+\frac{\Delta\hat{q}}{2}=\frac{4\pi^2}{3\hat{q}}\int\mu_{1/\sqrt{\hat{q}}}(x)^3\rd x,
\label{eq:SE_Bayes}
\end{align}
where $\mu_{1/\sqrt{\hat{q}}}$ denotes the spectrum of $\bS^{\star}+\frac{1}{\sqrt{\hat{q}}}\bZ$ with $\bZ\sim\GOE(d)$. For the ERM, $\mathsf{R}_{n,d}=\frac{2n}{d^2}\delta^2-\frac{\Delta}{2}$, where $\delta$ is given by the fixed point of the following equation
\begin{align}
\begin{cases}
4\frac{n}{d^2}\delta-\frac{\delta}{\epsilon}=\partial_1 J(\delta, \lambda\epsilon), \\
Q^\star +\frac{\Delta}{2}+2\frac{n}{d^2}\delta^2-\frac{\delta^2}{\epsilon}=(1 - \lambda\epsilon \del_2) J(\delta,\lambda\epsilon) \, ,
\end{cases}
\quad 
J(a,b):=\int_b^{+\infty}\mu_a(x)(x-b)^2\rd x .
\label{eq:SE_ERM}
\end{align}
with $\mu_a$ denoting the spectrum of $\bS^{\star}+a\bZ$ with $\bZ\sim\GOE(d)$.
\end{conjecture}

\section{Conclusion}
We studied a theoretical framework for scaling laws in shallow networks with feature learning by mapping them to sparse vector and low-rank matrix estimation. This allowed us to derive a comprehensive phase diagram for the excess risk scaling laws, uncovering a universality between diagonal and quadratic networks. Our analysis provides a first-principles explanation of the weight spectra–generalization connection: underfitting, overfitting, and approximation errors correspond directly to distinct spectral features, yielding a firm foundation for empirical observations of heavy-tailed weight spectra and their link to generalization.

There are many natural extensions of this work, such as exploring additional structures present in the data (e.g., non-trivial covariances \citep{wortsman2025kernel}), extending beyond two-layer networks and quadratic activations \citep{barbier2025statistical}, providing a rigorous proof of the state evolution conjecture following \cite{miolane2021distribution}. Moreover, our current work only analyzes the global minimum, so we should also look at compute scaling laws of GD/SGD \citep{arous2025learning} as well as the implicit biases of SGD towards heavy tails and its relation to generalization \citep{gurbuzbalaban2021heavy,simsekli2020hausdorff,hodgkinson2022generalization}. We hope these results will motivate further progress toward a systematic theory of neural scaling laws.

\section*{Acknowledgements}
We would like to thank Murat Erdogdu, Surya Ganguli, Antoine Maillard, Pierre Mergny and Denny Wu for insightful discussions. BL and LD were supported by the French government, managed by the National Research Agency (ANR), under the France 2030 program with the project references ``ANR-23-IACL-0008'' (PR[AI]RIE-PSAI) and ``ANR-25-CE23-5660'' (MAPLE), as well as the Choose France - CNRS AI Rising Talents program. We also acknowledge funding from the Swiss National Science Foundation grants SNSF SMArtNet (grant number 212049), OperaGOST (grant number 200021 200390) and DSGIANGO (grant number 225837). This work was supported by the Simons Collaboration on the Physics of Learning and Neural Computation via the Simons Foundation grant ($\#1257412$ (FK) and $\#1257413$ (LZ)).

\bibliography{main}
\bibliographystyle{abbrvnat}

\appendix
\section{The bridge from sparse estimation to neural networks}
\label{equivalence}

\subsection{\texorpdfstring{Equivalence between diagonal networks with $\ell_2$ weight decay and LASSO}{Equivalence between diagonal networks with l2 weight decay and LASSO}}

The first equivalence was discussed in a number of papers \citep{neyshabur2015norm,soudry2018implicit,pesme2023saddle}. We  consider the diagonal two–layer network with parameters $u,w \in \mathbb{R}^d$ and effective predictor
\[
\theta = u \odot w, \qquad f(x) = \sum_{i=1}^d \theta_i x_i,
\]
trained with squared loss and $\ell_2$ weight decay on both layers:
\begin{align}\label{app:eq:diagonal_ERM}
\min_{u,w \in \mathbb{R}^d}
\;\; \frac{1}{2}\|y - X(u \odot w)\|_2^2
+ \frac{\lambda}{2}\big(\|u\|_2^2 + \|w\|_2^2\big).
\end{align}
Alternatively, one may also consider a diagonal two-layer ReLU network with two branches per coordinate:
\[
f(x) \;=\; \sum_{i=1}^d \Big( u_i\,\sigma(w_i x_i) \;+\; u_i\,\sigma(-w_i x_i) \Big),
\qquad \sigma(z)=\max\{z,0\}.
\]
Using $\sigma(z)-\sigma(-z)=z$, each pair of branches along coordinate $i$ induces an
effective linear coefficient $\theta_i$ such that
\[
f(x) \;=\; \sum_{i=1}^d \theta_i x_i.
\]

We know show the reduction of this problem to the LASSO one:
\paragraph{Step 1. Lower bound via AM–GM.}
For each coordinate $i$ we have
\[
u_i^2 + w_i^2 \;\geq\; 2|u_i w_i| = 2|\theta_i|
\quad \text{(by AM--GM, with $a=u_i^2$, $b=w_i^2$)}.
\]
Summing over $i$ gives
\[
\|u\|_2^2 + \|w\|_2^2 \;\geq\; 2\|\theta\|_1.
\]
Therefore, for any factorization with $u \odot w = \theta$,
\begin{align}\label{app:eq:ERM_bound}
\frac{1}{2}\|y - X\theta\|_2^2
+ \frac{\lambda}{2}\big(\|u\|_2^2+\|w\|_2^2\big)
\;\;\geq\;\;
\frac{1}{2}\|y - X\theta\|_2^2 + \lambda \|\theta\|_1.
\end{align}

\paragraph{Step 2. Tightness.}
For any $\theta$, choose a factorization
\[
u_i = \operatorname{sign}(\theta_i)\,|\theta_i|^{1/2},
\qquad
w_i = |\theta_i|^{1/2}.
\]
Then $u_i^2 = w_i^2 = |\theta_i|$, so that
\[
u_i^2 + w_i^2 = 2|\theta_i|,
\quad
u_i w_i = \theta_i.
\]
Hence equality holds in (\ref{app:eq:ERM_bound}), and the regularizer becomes
\[
\frac{\lambda}{2}(\|u\|_2^2 + \|w\|_2^2) = \lambda \|\theta\|_1.
\]

Plugging back into (\ref{app:eq:diagonal_ERM}), we obtain the exact equivalence
\[
\min_{u,w}
\;\; \frac{1}{2}\|y - X(u \odot w)\|_2^2
+ \frac{\lambda}{2}\big(\|u\|_2^2 + \|w\|_2^2\big)
\;\equiv\;
\min_{\theta \in \mathbb{R}^d}
\;\; \frac{1}{2}\|y - X\theta\|_2^2 + \lambda \|\theta\|_1,
\]
which is precisely the \emph{LASSO} loss.

\subsection{\texorpdfstring{Equivalence between quadratic networks with $\ell_2$ weight decay and matrix compressed sensing}{Equivalence between quadratic networks with l2 weight decay and matrix compressed sensing}}

Again the equivalance has been discussed in a number of work \citep{gunasekar2017implicit,maillard2024fitting,erba2025nuclear,bandeira2025exact}

We consider the two-layer quadratic network with centered activations
\[
f(\bx;\bW) \;=\; \frac{1}{\sqrt{p}} \sum_{j=1}^p \Big[(\bw_j^\top \bx)^2 - \mathbb{E}[(\bw_j^\top \bx)^2]\Big].
\]
Centering is equivalent to learning (and absorbing) the constant offset via a bias term, and can also be naturally implemented in practice by batch/layer normalization applied after squaring.

This network can be written as
\[
f(\bx) \;=\; \Tr[\bS\bZ],
\]
where $\bS:=\tfrac{1}{\sqrt p}\bW\bW^\top \succeq 0$ and $\bZ:=\bx\bx^\top-\boldsymbol{\Sigma}_x$, $\boldsymbol{\Sigma}_x = \mathbb{E}[xx^\top]$. Thus the network corresponds exactly to a PSD matrix sensing model with centered measurements 
$\bZ$. 
Centering removes only a constant offset, which in practice would be absorbed by a bias term 
or handled automatically by batch/layer normalization. 
Moreover, weight decay on $\bW$ induces a trace penalty on $\bS$, since
\[
\|\bW\|_F^2 = \sqrt{p}\,\mathrm{tr}(\bS),
\]
so that training is equivalent to trace-regularized PSD matrix sensing.

Following the universality results for matrix sensing 
(see, e.g., \cite{maillard2024fitting,bandeira2025exact,maillard2024bayes,xu2025fundamental,erba2025nuclear}),
the analysis can be simplified by replacing the empirical sensing operators 
$\bZ$ by i.i.d.\ Gaussian symmetric matrices 
with matching covariance structure. 
In particular, for $x_i \sim \mathcal{N}(0,I_d)$, the centered measurements 
are distributed as rank-one Wishart fluctuations, which are asymptotically equivalent, 
in the sense of state evolution and AMP analysis, to Gaussian measurements with the same variance. 
Hence, without loss of generality, we may study the trace-regularized PSD matrix sensing problem
with Gaussian measurement operators
\[
y_\mu=\Tr[\bS\boldsymbol{G}_\mu] + \xi_\mu,
\qquad \boldsymbol{G}_\mu \sim \GOE(d).
\]

\section{Noiseless setting: error rates and spectral properties}
\label{app:noiseless}
In the noiseless case $\Delta=0$, we have the following rates.

\begin{res}[Noiseless rates]
\label{res:noiseless}
    Under the setting of Section \ref{sec:intro} for $\Delta = 0$ and $n_{\rm eff} \gg 1$, the excess risk satisfies
 \begin{equation} \label{eq:ERM_rates_noiseless}
R=\begin{cases}
\Theta\left(\neff^{1-2\gamma}\right)
    &\text{if}\quad 
    \neff\ll d\mathand\lambda\ll\frac{1}{d^{1/2}\neff^{\gamma-1}}
    \\
    \Theta\left(\lambda^2d^2/\neff^2\right)
    &\text{if}\quad \neff\gg d\mathand \lambda\ll\frac{\neff}{d^{\gamma+1/2}}
    \\
    \Theta\left((\lambda d^{1/2}/\neff)^{2-1/\gamma}\right)
        &\text{if}\quad 
        \max\left(\frac{1}{d^{1/2}\neff^{\gamma-1}},\frac{\neff}{d^{\gamma+1/2}}\right)\ll\lambda\ll\frac{\neff}{d^{1/2}}
    \end{cases} \, ,
\end{equation}
and $R = \Theta(1)$ otherwise.
In the diagonal network case, the first rate holds up to logarithmic factors that we specify in eq. (\ref{app:eq:noiseless_log_factors}).
\end{res}
Note that as in the least-squares case, the noiseless rates are faster \citep{aubin2020generalization,berthier2020tight, cui2021generalization}. Perhaps more surprising, they are also universal, with no difference between the quadratic and diagonal networks. Moreover, the error rate $\Theta\left(\neff^{1-2\gamma}\right)$
agrees with \cite{arous2025learning} as in their setting there is no regularization.

\begin{cor}[Optimal regularization and optimality of ERM, noiseless setting]
\label{cor:optlambda:noiseless}
    The optimal regularization satisfies
    \begin{equation}
\lambda_{\rm opt} = \begin{cases}
\Theta(\neff^{-\gamma+1}d^{-1/2}),&\mathif \Delta = 0\mathand 1\ll\neff\ll d\\
0^+&\mathif \Delta = 0\mathand \neff\gg d
\end{cases} \, .
\end{equation}
where $\tilde{\Theta}$ is up to logarithm factors in the argument. The excess risk rates in the optimally regularized case match Bayesian risks, given by
\begin{equation}
\mathsf{R}_{n_{\rm eff},d}(\lambda_{\rm opt}) = \Theta(\mathsf{R}_{n_{\rm eff}}^{\rm BO})=
    \begin{cases}
        \Theta(\neff^{1-2\gamma})
        &\mathif \Delta = 0 \mathand 1\ll \neff\ll d
        \\
        0
        &\mathif \Delta = 0 \mathand \neff\gg d
    \end{cases} \, ,
\end{equation}
and $\mathsf{R}_{n_{\rm eff},d} = \Theta(1)$ otherwise.
Again, in the diagonal network case, the first and third regimes the rate hold up to logarithmic factors that we specify in eq. (\ref{app:eq:noiseless_log_factors}).
\end{cor}

\begin{figure}[t]
\centering
\begin{tikzpicture}[scale=13.96]

\tikzstyle{every node}=[font=\small]


\definecolor{bulk}{RGB}{178,178,178}
\definecolor{phaseA}{RGB}{255,217,47}
\definecolor{phaseB}{RGB}{100,168,195}
\definecolor{phaseC}{RGB}{166,216,84}
\definecolor{phaseD}{RGB}{141,160,203}
\definecolor{phaseV}{RGB}{252,141,98}

\def\Ox{0.08}
\def\Oy{0.09}
\def\AXx{0.62}
\def\AXy{0.35}

\def\SYone{0.115}
\def\SYtwo{0.295}

\def\legx{0.74}
\def\legy{0.075}

\coordinate (origin) at (\Ox, \Oy);
\coordinate (Y1) at (\Ox, \Oy+0.37 * \AXy+0.01);
\coordinate (Y2) at (\Ox, \Oy+0.92 * \AXy);

\coordinate (HX1) at (\Ox+0.33 * \AXx, \Oy+0.92 * \AXy);
\coordinate (HX2) at (\Ox+0.65 * \AXx, \Oy+0.92 * \AXy);
\coordinate (HX3) at (\Ox+0.95 * \AXx, \Oy+0.92 * \AXy);

\coordinate (MX1) at (\Ox+0.2 * \AXx, \Oy+0.4 * \AXy+0.01);
\coordinate (MX2) at (\Ox+0.55 * \AXx, \Oy+0.45 * \AXy+0.01);
\coordinate (MX3) at (\Ox+0.95 * \AXx, \Oy+0.5 * \AXy+0.01);

\coordinate (LX1) at (\Ox+0.2 * \AXx, \Oy);
\coordinate (LX4) at (\Ox+0.42 * \AXx, \Oy);
\coordinate (LX5) at (\Ox+0.5 * \AXx, \Oy);
\coordinate (LX2) at (\Ox+0.55 * \AXx, \Oy);
\coordinate (LX3) at (\Ox+0.95 * \AXx, \Oy);

\fill[phaseA!50]
  (origin) -- (Y1) -- (Y2) -- (HX1)  -- (MX1) -- (LX1) -- cycle;
\node[anchor=north west] (r) at (Y1) {Ib};
\node[anchor=north west] (r) at (Y2) {Ia};

\fill[phaseB!50]
  (MX1) -- (HX1) -- (HX2)  -- (MX2) -- cycle;
\node[anchor=north west] (r) at (HX1) {II};

\fill[phaseC!50]
  (MX2) -- (HX2) -- (HX3)  -- (MX3) -- (LX3) -- (LX2)--cycle;
\node[anchor=north west] (r) at (HX2) {III};

\fill[phaseD!50]
  (MX1) -- (MX2) -- (LX2)  -- (LX1) -- cycle;
\node[anchor=north west] (r) at (MX1) {IV};

\draw[black, thick] (MX1) -- (HX1);
\draw[black, thick] (MX1) -- (LX1);
\draw[red, thick] (MX2) -- (LX2);
\draw[black, thick] (MX2) -- (HX2);
\draw[black, thick] (LX3) -- (MX3);
\draw[black, thick] (Y1) -- (MX1)--(MX2);
\draw[black, thick] (MX3) -- (HX3) -- (Y2) -- (\Ox,\Oy);


\def\sizx{0.1}
\def\heightspike{0.07}
\def\shiftspike{0.01}
\def\widthspike{0.005}
\def\heightbulk{0.04}
\def\widthbulkverysmall{0.02}
\def\widthbulksmall{0.02}
\def\widthbulkbig{0.045}
\def\heightout{0.02}

\def\xxx{\Ox+0.01}
\def\yyy{\SYtwo}
\draw[->,thick] (\xxx,\yyy) -- (\xxx+\sizx,\yyy);
\draw[draw=black, fill=bulk] (\xxx+\shiftspike,\yyy) rectangle ++(\widthspike,\heightspike);

\def\xxx{\Ox+0.01}
\def\yyy{\SYone}

\draw[->,thick] (\xxx,\yyy) -- (\xxx+\sizx,\yyy);
\draw[draw=black, fill=bulk] (\xxx+\shiftspike,\yyy) rectangle ++(\widthspike,\heightspike);

\draw[draw=black, fill=bulk] plot (\xxx+\shiftspike+\widthspike,\yyy+\heightbulk) to[out=-55,in=100] (\xxx+\widthspike+\widthbulksmall*0.95,\yyy+\heightbulk/3) to[out=-80,in=-85] (\xxx+\widthspike+\widthbulksmall,\yyy) -- (\xxx+\shiftspike+\widthspike,\yyy) -- cycle;

\def\xxx{\Ox+0.22}
\def\yyy{\SYtwo}
\draw[->,thick] (\xxx,\yyy) -- (\xxx+\sizx,\yyy);
\draw[draw=black, fill=bulk] (\xxx+\shiftspike,\yyy) rectangle ++(\widthspike,\heightspike);

\draw (\xxx++0.8*\sizx,\yyy) -- (\xxx+0.8*\sizx,\yyy + \heightout);
\draw (\xxx++0.45*\sizx,\yyy) -- (\xxx+0.45*\sizx,\yyy + \heightout);
\draw (\xxx++0.29*\sizx,\yyy) -- (\xxx+0.29*\sizx,\yyy + \heightout);
\draw (\xxx++0.22*\sizx,\yyy) -- (\xxx+0.22*\sizx,\yyy + \heightout);
\draw (\xxx++0.19*\sizx,\yyy) -- (\xxx+0.19*\sizx,\yyy + \heightout);
\draw (\xxx++0.17*\sizx,\yyy) -- (\xxx+0.17*\sizx,\yyy + \heightout);
\draw (\xxx++0.165*\sizx,\yyy) -- (\xxx+0.165*\sizx,\yyy + \heightout);

\def\xxx{\Ox+0.43}
\def\yyy{0.5*\SYtwo+0.5*\SYone}
\draw[->,thick] (\xxx,\yyy) -- (\xxx+\sizx,\yyy);

\def\xstart{\xxx+\shiftspike+\widthspike}
\def\ystart{\yyy}
\def\xend{\xxx+0.7*\sizx}
\def\exponent{10} 
\draw[domain=0:10*\sizx-10*\xstart, smooth, fill=bulk, variable=\u]
  plot ({\xstart+\u/10},
        {\ystart+0.07*pow(\xstart,\exponent)*(pow(\xstart+\u,-\exponent)-pow(10*\xend,-\exponent)})-- (\xstart+0.7*\sizx,\yyy) -- (\xstart,\yyy)--cycle;

\def\xxx{\Ox+0.19}
\def\yyy{\SYone}
\draw[->,thick] (\xxx,\yyy) -- (\xxx+\sizx,\yyy);
\draw[draw=black, fill=bulk] (\xxx+\shiftspike,\yyy) rectangle ++(\widthspike,\heightspike);

\draw (\xxx++0.8*\sizx,\yyy) -- (\xxx+0.8*\sizx,\yyy + \heightout);
\draw (\xxx++0.45*\sizx,\yyy) -- (\xxx+0.45*\sizx,\yyy + \heightout);
\draw (\xxx++0.29*\sizx,\yyy) -- (\xxx+0.29*\sizx,\yyy + \heightout);
\draw (\xxx++0.22*\sizx,\yyy) -- (\xxx+0.22*\sizx,\yyy + \heightout);
\draw (\xxx++0.19*\sizx,\yyy) -- (\xxx+0.19*\sizx,\yyy + \heightout);
\draw (\xxx++0.17*\sizx,\yyy) -- (\xxx+0.17*\sizx,\yyy + \heightout);
\draw (\xxx++0.165*\sizx,\yyy) -- (\xxx+0.165*\sizx,\yyy + \heightout);

\draw[draw=black, fill=bulk] plot (\xxx+\shiftspike+\widthspike,\yyy+\heightbulk) to[out=-55,in=100] (\xxx+\widthspike+\widthbulksmall*0.95,\yyy+\heightbulk/3) to[out=-80,in=-85] (\xxx+\widthspike+\widthbulksmall,\yyy) -- (\xxx+\shiftspike+\widthspike,\yyy) -- cycle;


\draw[->,thick] (\Ox,\Oy) -- (\Ox + \AXx,\Oy);
\draw[->,thick] (\Ox,\Oy) -- (\Ox,\Oy + \AXy);

\node[anchor=north] (r) at (\Ox + \AXx/2,\Oy-0.05) {Effective number of samples $\neff$ ($n$ for diagonal, $n/d$ for quadratic)};

\node[anchor=south, rotate=90] (r) at (\Ox-0.05,\Oy + \AXy/2) {Regularization $\lambda$};

\node[anchor=north] (r) at (LX1) {$\Theta(1)$};
\node[anchor=north] (r) at (LX2) {$\Theta(d)$};

\node[anchor=east] (r) at (Y1) {$\frac{1}{d}$};

\node[anchor=south] (r) at (HX1) {$\qquad \lambda = \neff/d^{1/2}$};
\node[anchor=south] (r) at (HX2) {$\qquad\quad  \lambda = \neff/d^{1/2+\gamma}$};

\node[anchor=east] (r) at (MX2) [below=2.5mm,left=0mm, rotate=3] {\tiny $\lambda=\neff^{1-\gamma}/\sqrt{d}$};

\def\legs{0.05}
\def\legbox{0.02}

\draw[draw=black, fill=phaseD] (\legx,\legy - \legbox/2   + 2*\legs) rectangle ++(\legbox,\legbox);
\node[anchor=west] (r) at (\legx + \legbox,\legy + 2*\legs) {IV: ${\varepsilon} \asymp \neff^{1-2\gamma}$};
\draw[draw=black, fill=phaseC] (\legx,\legy - \legbox/2  + 0.00125 + 3*\legs) rectangle ++(\legbox,\legbox);
\node[anchor=west] (r) at (\legx + \legbox,\legy + 3*\legs) {III: ${\varepsilon} \asymp \lambda^2d^2/ \neff^2$};

\draw[draw=black, fill=phaseB] (\legx,\legy - \legbox/2   + 4*\legs) rectangle ++(\legbox,\legbox);
\node[anchor=west] (r) at (\legx + \legbox,\legy + 4*\legs) {II: ${\varepsilon} \asymp \left(\lambda d^{1/2} / \neff\right)^{2-1/\gamma}$};

\draw[draw=black, fill=phaseA] (\legx,\legy - \legbox/2  + 0.0025 + 5*\legs) rectangle ++(\legbox,\legbox);
\node[anchor=west] (r) at (\legx + \legbox,\legy + 5*\legs) {Ia,Ib: ${\varepsilon} \asymp 1$};

\end{tikzpicture}
\caption{
Excess risk rates of \cref{res:noiseless} for the noiseless task ($\Delta=0$), the corresponding spectral properties of neural networks.
}
\label{fig:phase_noiseless}
\end{figure}

The phase diagram for the noiseless case is simpler, and exhibits just a subset of the phases. Specifically, phases \textcolor{phaseI}{I}, \textcolor{phaseII}{II} are the same as the noisy cases. For the other two phases, we have
\begin{itemize}
    \item \textcolor{phaseIII}{III (Heavy-tail)}: The spectrum is a perturbed and shifted version of the target's spectrum. The perturbation is of order $\caO(\lambda d^3/n^{3/2})$, and the shift is of about $\frac{\lambda d^2}{4n}$. Thus the risk is mainly caused by the shift, which gives an error of $\Theta(\lambda^2 d^4/n^2)$. When $\lambda=0$, the spectrum is the same as the target's, so the risk is zero.
    \item \textcolor{phaseIV}{IV (Bulk+Outliers)}: The bulk is of size $\Theta(d^{\gamma-3/10}/n^{\gamma-2/5})$, and there are approximately $\frac{n}{d}$ outliers. As $n$ increases, the bulk shrinks, all spikes come out of the bulk one by one, and the risk keeps decreasing. The contribution of the bulk is always negligible, so Result \ref{res:decomposition} implies that the risk is proportional to the total power of the unrecovered spikes: $\mathsf{R}=\frac{1}{d}\sum_{i\geq n/d}(\sqrt{d}i^{-\gamma})^2=\Theta((d/n)^{2\gamma-1})$. Note that this is also information theoretically optimal, because with $n$ data points, we could learn at most $\frac{n}{d}$ uncorrelated directions.
\end{itemize}

\begin{figure}[t]
    \centering
\includegraphics[width=0.9\linewidth]{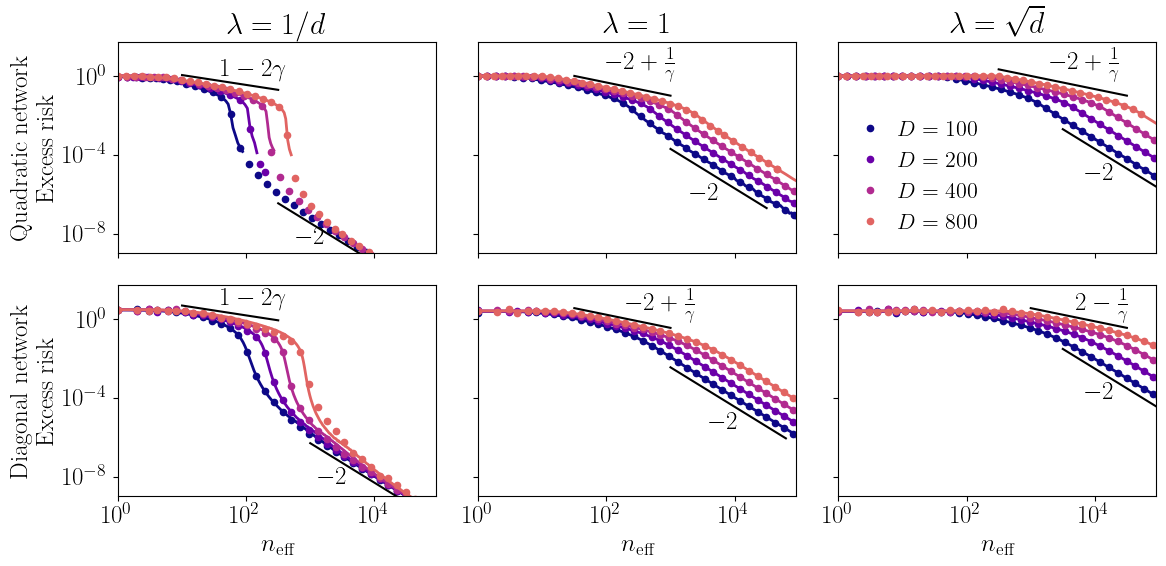}
    \caption{Excess risk in simulations versus non-asymptotic state evolution (solid lines) as a function of $\neff$. The setting is the same as Figure \ref{fig:rates} except that $\Delta=0$. Black lines indicate the decay rates of the excess risk predicted by Result~\ref{res:noiseless}, again showing good agreement.}
    \label{fig:noiseless}
\end{figure}

The Result \ref{res:noiseless} are confirmed by Figure \ref{fig:noiseless}.

\section{Derivation details - Diagonal linear network}\label{app:sec:diagonal_derivation_rates}

Considering the reparametrization defined in Section \ref{sec:intro} and detailed in Appendix \ref{equivalence}, mapping empirical risk minimization with $L_2$ penalty on a two-layer diagonal linear network to LASSO regression, in this section we study the supervised learning problem 
\begin{align}\label{eq:app:lasso_def}
    \hat{\vtheta} = \arg\min_{\vtheta\in\R^d} \left\{\frac{1}{2}\sum_{\mu=1}^n\left(y_\mu-\frac{\langle\vx_\mu,\vtheta\rangle}{\sqrt{d}}\right) + \lambda \lVert\vtheta\rVert_1\right\},
\end{align}
with $\vx_\mu\sim\mathcal{N}(\boldsymbol{0},\boldsymbol{I}_d)$, and 
\begin{align}
    y_\mu = \frac{\langle\vx_\mu,\vtheta^\star\rangle}{\sqrt{d}} + \sqrt{\Delta}\xi_\mu,&& \xi_\mu\sim\mathcal{N}(0,1),&& \mu = 1,\ldots,n\\
    \vtheta^\star \sim\mathcal{N}(\boldsymbol{0},\boldsymbol{\Lambda}),&&\Lambda_{ij}=\delta_{ij} \Lambda_i,&& i=1,\ldots,d
\end{align}
We also define the parameter $Q^\star = d^{-1}\Tr\boldsymbol{\Lambda}\xrightarrow{d\to\infty}\zeta(2\gamma)$. Without loss of generality we choose $\Lambda_1\geq\Lambda_2\geq\ldots\geq\Lambda_d$. The excess risk is defined as 
\begin{align}\label{app:eq:def:lasso_excess_risk}
    R(\hat{\vtheta}) &= \frac{1}{d}\E[(\vx^T\hat{\vtheta}-\vx^T\vtheta^\star)^2].
\end{align}
In sections \ref{app:sec:GAMP_SE} and \ref{app:sec:BO_risk} we derive the state evolution equations \ref{eq:app:ERM_SE_linear}) for the excess risk of the ERM estimator eq. (\ref{eq:app:lasso_def}) and (\ref{app:eq:BO_MMSE}) for the Bayes-optimal estimator in the high-dimensional limit $n,d\to\infty$ with $n/d$ and $\lambda$ fixed. Then, in sections \ref{app:sec:BO_rates} and \ref{app:sec:lasso_rates}, assuming the excess risk equations holds for arbitrary scaling between dimensions and regularization, we derive the Results in Section \ref{sec:phase_diagrams}.

\subsection{Generalized Approximate Message Passing and State Evolution} \label{app:sec:GAMP_SE}
Our theory is built on the analysis of {\it Generalized Approximate Message Passing} (GAMP) algorithms, tailored for Bayes-optimal estimation and (convex) empirical risk minimization. In this section we provide an overview of the derivation of the expressions for $R$ and the LASSO $R$ in our setting from the GAMP framework.

Consider the matrix $\mX\in\R^{n\times d}$, with i.i.d. Gaussian components $X_{ij}\sim\mathcal{N}(0,1)$, the vectors $\vb^t\in\R^d$, $\vomega^t\in\R^n$, and the functions (known as {\it denoisers}) $f_t:\R^d\to\R^d$ and $g_t:\R^n\to\R^n$, with $t\geq 1$. The generic form of GAMP \citep{donoho2009message,ranganGAMP} is given by
\begin{align}\label{app:eq:GAMP_omega}
    \vomega^t &= \mX f_t(\vb^t) - v_tg_{t-1}(\vomega^{t-1}),\\
    \vb^{t+1} &= \mX^Tg_{t}(\vomega^t) + a_tf_t(\vb^t).\label{app:eq:GAMP_b}
\end{align}
The terms $a_t$ and $v_t$ are known in the statistical physics literature as {\it Onsager terms}, and they are defined as
\begin{align}
    a_t = -\frac{1}{d}\sum_{\mu=1}^n \frac{\partial}{\partial \omega_i} g_t(\vomega),\qquad v_t = \frac{1}{d}\sum_{i=1}^d \frac{\partial}{\partial b_i}f_t(\vb).
\end{align}
For separable denoiser functions\footnote{$f:\R^d\to\R^d$ is separable if $\forall i\in\{1,\ldots,d\}\;:[f(\vb\in\R^d)]_i = f_i(b_i)$, for some scalar function $f_i:\R\to\R$}, one can track statistics of the iterated vectors $\vb^t$, $\vomega^t$, leveraging well-known results from \cite{bayati2011dynamics,javanmard2013state}, through the so called {\it state evolution}. 
\subsubsection{GAMP for convex optimization}
Consider the problem of minimizing the empirical risk with loss $\ell(y,z)$ convex in the second argument and convex penalty $r(\theta)$,
\begin{align}\label{app:eq:def_optimization}
    \arg\min_{\vtheta\in\mathcal \R^d}\sum_{\mu=1}^n \ell(y^\mu, \vtheta^T\vx^\mu) + \sum_{i=1}^d r(\theta_i),
\end{align}
It is possible to design a GAMP algorithm whose fixed points are solutions to the problem defined in (\ref{app:eq:def_optimization}). A detailed discussion on this approach can be found in \cite{feng2022}. 
Define the functions
\begin{align}\label{app:eq:lasso_denoiser_g}
    \overline{g}(\omega, y, v) &:= \operatorname{prox}_{v\ell(y,\cdot)}(\omega),\qquad g(\omega,y,v)=\frac{\overline{g}(\omega, y ,v)-\omega}{v}\\
    f(b,a) &:= \operatorname{prox}_{\frac{1}{a}r}\left(\frac{b}{a}\right), \label{app:eq:lasso_denoiser_f}
\end{align}
where the {\it proximal operator} of a convex function $f$ is defined as
\begin{align}
    \operatorname{prox}_f(x)=\arg\min_{z\in\R}\left\{f(z) + \frac{1}{2}(z-x)^2\right\}.
\end{align}
Then, Proposition 4.4 in \cite{feng2022}, guarantees that, given a fixed point $(\vomega, \vb)$ of the GAMP algorithm eq. (\ref{app:eq:GAMP_omega},\ref{app:eq:GAMP_b}) with denoiser functions $g_t(\omega)=g(\omega,y,v_t)$ and $f_t(b)=f(b,a_t)$, the vector $\hat{\vtheta}:= f_t(\vb)$ is the unique minimizer of (\ref{app:eq:def_optimization}). 

As mentioned, in the high-dimensional limit $n,d\to\infty$, with $n/d$ fixed, we can track statistics of the iterated variables through a set of state evolution equations. We stress that the following result hold for the considered linear target function $y = \langle(\vtheta^\star)^T,\vx\rangle/\sqrt{d}$.
\begin{theorem}[\cite{bayati2011dynamics,javanmard2013state}, informal] \label{app:th:state_evolution}
    Define
    \begin{align}
    \begin{cases}
        \hat q^t &= \frac{n}{d}\E_{(z,\omega_t)}[g(\omega_t,z,v_t)^2]\\
        \hat m^t &=\frac{n}{d}\E_{(z,\omega_t)}[\partial_z g(\omega_t,z,v_t)]\\
        \hat v^t &=-\frac{n}{d}\E_{(z,\omega_t)}[\partial_\omega g(\omega_t,z,v_t)]
        \end{cases}
        \qquad
        \begin{cases}
        q^{t+1}&=\frac{1}{d}\E_{(\vxi,\vtheta^\star)}[\lVert f(\sqrt{\hat q^t}\vxi+\hat m^t \vtheta^\star,\hat v^t)\rVert^2]\\
        m^{t+1}&=\frac{1}{d}\E_{(\vxi,\vtheta^\star)}[\langle f(\sqrt{\hat q^t}\vxi+\hat m^t \vtheta^\star,\hat v^t), \vtheta^\star\rangle]\\
        v^{t+1}&=\frac{1}{d}\E_{(\vxi,\vtheta^\star)}[\nabla_b\cdot f(\sqrt{\hat q^t}\vxi+\hat m^t \vtheta^\star, \vtheta^\star,\hat v^t)]
        \end{cases}
    \end{align}
    where $\vxi\sim\mathcal N(\boldsymbol{0},\boldsymbol{I}_d)$ and
    \begin{align}
        \left(\begin{array}{c}
             z \\
             \omega^t 
        \end{array}\right) \sim \mathcal N
        \left(\left(\begin{array}{c}
             0  \\
             0 
        \end{array}\right), 
        \left(\begin{array}{cc}
           Q^\star  & m^t \\
            m^t & q^t
        \end{array}
        \right)
        \right).
    \end{align}
    Then the iterated vectors $\vomega^t$ and $\vb^t$ of the GAMP algorithm (\ref{app:eq:GAMP_omega},\ref{app:eq:GAMP_b}), with denoiser functions (\ref{app:eq:lasso_denoiser_g},\ref{app:eq:lasso_denoiser_f}), respectively converge weakly to the Gaussian vectors $\boldsymbol{\Omega}^t = \sqrt{q^t-m^t}\boldsymbol{w}+m^t \mX \vtheta^\star$ (with $\vw\sim\mathcal{N}(\boldsymbol{0},\boldsymbol{I}_n)$) and $\mB^t = \hat m^t \vtheta^\star+\sqrt{\hat q^t}\vxi$, in the sense that, for any uniformly pseudo-Lipshitz of order $k$, deterministic $\phi:(\R^d\times\R^n)^t\times\R^d\to\R$,
    \begin{align}
        \phi(\vb^0,\vomega^0,\vb^1,\vomega^1,\ldots,\vomega^{t-1},\vb^t)\overset{\rm P}{\simeq} \E\phi(\mB^0, \boldsymbol{\Omega}^0, \mB^1, \boldsymbol{\Omega}^1,\ldots, \boldsymbol{\Omega}^{t-1},\mB^t).
    \end{align}
\end{theorem}
The previous theorem readily implies that, in the high-dimensional limit, $a_t \simeq \hat v^t$, and, given $\hat{\vtheta}^t \simeq f(\vb^t,a_t)$,
\begin{align}
    \frac{1}{d}\langle \hat{\vtheta}^t,\vtheta^\star\rangle \simeq m^t, \qquad \frac{1}{d}\lVert \hat{\vtheta}\rVert^2 \simeq q^t
\end{align}
and the generalization error of the estimator $\hat y(\vx)=f(\vx,\hat{\vtheta}^t) =(\langle\hat{\vtheta^t},\vx)/\sqrt{d}$ is
\begin{align}\label{eq:se_generalization}    R(\hat{\vtheta}^t)\simeq \E_{\vx}\left(\frac{\langle\hat{\vtheta^t},\vx\rangle}{\sqrt{d}} - \frac{\langle\hat{\vtheta^t},\vx\rangle}{\sqrt{d}}\right)^2 = Q^\star - 2m^t + q^t.
\end{align}

\paragraph{LASSO regression}
In the case of LASSO, with $\ell(y,z)=(y-z)^2/2$ and $r(\theta)=\lambda |\theta|$, we have that 
\begin{align}
    g(\omega,y,v)= \frac{y-\omega}{1+v},\qquad f(b,a) = \frac{1}{a}{\rm ST}_{\lambda}(b),
\end{align}
where ${\rm ST}_{\lambda}(b)=\max(b-\lambda,0)-\max(-b-\lambda,0)$ denotes the soft-thresfolding function. The state evolution equations in this setting read
\begin{align}
    \begin{cases}
        \hat q^t &= \frac{n(\Delta + Q^\star - 2m^t + q^t)}{d(1+v^t)^2}\\
        \hat m^t &=\frac{n}{d(1 + v^t)}\\
        \hat v^t &=\frac{n}{d(1 + v^t)}
        \end{cases}
        \qquad
        \begin{cases}
        m^{t+1}&=\frac{1}{d}\sum_{i=1}^d \Lambda_k \operatorname{erfc}\left(\frac{\lambda}{\sqrt{2((\hat m^t)^2 \Lambda_i+\hat q^t)}}\right)\\
        v^{t+1}&=\frac{1}{d\hat m^t}\sum_{i=1}^d\operatorname{erfc}\left(\frac{\lambda}{\sqrt{2((\hat m^t)^2 \Lambda_i+\hat q^t)}}\right)
        \end{cases}
    \end{align}
    and 
    \begin{align}
        q^{t+1} = &\frac{1}{d(\hat m^t)^2}\sum_{i=1}^d\left[ ((\hat m^t)^2\Lambda_i+\hat q^t+\lambda^2)\operatorname{erfc}\left(\frac{\lambda}{\sqrt{2((\hat m^t)^2 \Lambda_i+\hat q^t)}}\right) \right]\\&-\frac{1}{d(\hat m^t)^2}\sum_{i=1}^d\left[\frac{2\lambda}{\sqrt{2\pi}}\sqrt{(\hat m^t)^2 \Lambda_i+\hat q^t}e^{-\lambda^2/(2((\hat m^t)^2 \Lambda_i+\hat q^t))}\right].
    \end{align}
At convergence, substituting the equation for $v$ into the equation for one for $\hat m$, introducing the parameter $\nu = \frac{\lambda}{\hat m}\sqrt{\frac{n}{2d}}$ and leveraging eq. (\ref{eq:se_generalization}), one obtains that the excess risk for LASSO regression in this setting is given by
\begin{align}\nonumber
    R(\hat{\vtheta}) \simeq  \mathsf R_{n,d}(\nu)=& \frac{1}{n}\sum_{i=1}^d \left[\frac{n}{d}\Lambda_i\operatorname{erf}\left(\frac{\nu}{\sqrt{\frac{n}{d}\Lambda_i+ \hat\Delta }}\right) + \left(\hat\Delta+2\nu^2\right)\operatorname{erfc}\left(\frac{\nu}{\sqrt{ \frac{n}{d}\Lambda_i+\hat\Delta }}\right)\right] 
    \\&- \frac{2\nu}{n\sqrt{\pi}}\sum_{i=1}^d \left[\sqrt{\frac{n}{d}\Lambda_i+\hat\Delta}\,e^{-\nu^2/(\frac{n}{d}\Lambda_i+\hat\Delta)}\right], \label{eq:app:ERM_SE_linear_generic}
\end{align}
with $\hat\Delta = \Delta + \mathsf R_{n,d}(\nu)$ and
\begin{align}\label{eq:app:self_consistent_nu_generic}
    \frac{\lambda}{\nu}\sqrt{\frac{n}{2d}} + \frac{1}{d}\sum_{i=1}^d \operatorname{erfc}\left(\frac{\nu}{\sqrt{ \frac{n}{d}\Lambda_i+\hat\Delta}}\right) = \frac{n}{d}.
\end{align}
For the specific choice of covariance $\Lambda_i=di^{-2\gamma}$, this becomes
\begin{align}\nonumber
    R(\hat{\vtheta}) \simeq \mathsf R_{n,d}(\nu)=& \frac{1}{n}\sum_{i=1}^d \left[ni^{-2\gamma} \operatorname{erf}\left(\frac{\nu}{\sqrt{ni^{-2\gamma}+ \hat\Delta }}\right) + \left(\hat\Delta+2\nu^2\right)\operatorname{erfc}\left(\frac{\nu}{\sqrt{ ni^{-2\gamma}+\hat\Delta }}\right)\right] 
    \\&- \frac{2\nu}{n\sqrt{\pi}}\sum_{i=1}^d \left[\sqrt{ni^{-2\gamma}+\hat\Delta}\,e^{-\nu^2/(ni^{-2\gamma}+\hat\Delta)}\right], \label{eq:app:ERM_SE_linear}
\end{align}
and
\begin{align}\label{eq:app:self_consistent_nu}
    \frac{\lambda}{\nu}\sqrt{\frac{n}{2d}} + \frac{1}{d}\sum_{i=1}^d \operatorname{erfc}\left(\frac{\nu}{\sqrt{ ni^{-2\gamma}+\hat\Delta}}\right) = \frac{n}{d}.
\end{align}

\begin{conjecture}\label{conj:diagonal}
    Define $R(\hat{\vtheta},\lambda)$ the excess risk eq.(\ref{app:eq:def:lasso_excess_risk}) of the LASSO estimator $\hat{\vtheta}$ eq. (\ref{eq:app:lasso_def}) with regularization strength $\lambda$. Then, there exists $C>0$ such that, for any $n,d>C$, with probability $1-o_n(1)-o_d(1)$,
    \begin{align}
        |R(\hat{\vtheta},\lambda) - \mathsf R_{n,d}(\nu(\lambda))| = \mathsf R_{n,d}(\nu(\lambda))\cdot o_{n,d}(1),
    \end{align}
    with $\mathsf R_{n,d}(\nu)$ defined in eq. (\ref{eq:app:ERM_SE_linear}) and $\nu(\lambda)$ solution of eq. (\ref{eq:app:self_consistent_nu}).
\end{conjecture}
\subsubsection{Spectral structure of the estimator}\label{app:sec:lasso_spectral}
Theorem \ref{app:th:state_evolution} readily implies Result~\ref{res:structure}. Given the unique fixed point $(\vomega,\vb)$ of GAMP, the minimizer of eq. (\ref{eq:def:lasso}) is given by $\hat\theta = \frac{1}{\hat m}{\rm ST}_{\lambda}(\vb)$, which satisfies, in distribution
\begin{align}\label{app:eq:lasso_result4}
        \hat{\theta}_i \sim {\rm ST}_{\epsilon_{\rm d}}( 
            \theta^\star_i + \delta_{\rm d} z_i) \, ,
    \end{align}
with $\epsilon_{\rm d} := \lambda / \hat m$, $\delta_{\rm d} := \sqrt{\hat q}/\hat m$ and $z_i\sim\mathcal N(0,1)$. Note that $\epsilon_{\rm d} = \nu\sqrt{2d/n}$ and 
\begin{align}
    |\hat\theta_i|\sim \max\left(\left|\,\left|\underbrace{\theta^\star_{i} + z_i\sqrt{\hat \Delta\frac{d}{n}}}_{=: u_i}\right| - \nu\sqrt{\frac{2d}{n}}\right|, \,0\right).
\end{align}
The random variable variable $u_i$ satisfies
\begin{align}
    u_i \sim \mathcal{N}\left(0, \Lambda_i + \hat \Delta\frac{d}{n}\right) \implies& u_i = \begin{cases}
        \theta^\star_{i} + o_p(\sqrt{\Lambda_i}),&\mathif n\Lambda_i\gg d\hat\Delta,\\
        z_i\sqrt{\hat \Delta\frac{d}{n}}(1+o_p(1)) & \mathif n\Lambda_i\ll d\hat\Delta,
    \end{cases}
\end{align}
where we say that $u_i$ corresponds to the signal component $\theta_{\star i}$ in probability when the signal variance $\Lambda_i$ dominates over the effective noise variance $\hat\Delta d/n$; conversely, the effective noise dominates $u_i$.\\

Therefore, the ensemble $\{u_i\}_{i|n\Lambda_i < d\hat\Delta}$ constitutes a "bulk" of i.i.d. Gaussian variables, representing the combined effect of label noise and the limited number of samples. In fact, if the sample size is large enough, namely $n\gg d\hat \Delta\Lambda_d^{-1}$, the effect of the noise becomes undetectable. We refer to the remaining $\{u_i\}_{i|n\Lambda_i > d\hat\Delta}$ as "spikes", representing the components of the true signal $\vtheta^\star$ that we want to learn. Therefore, the scale $i_{\hat\Delta}$ such that $n\Lambda_{i_{\hat \Delta}} \sim d\hat\Delta$ represents the number of "learnable" components. Note that in the diagonal linear network setting, the noise bulk has no sharp boundaries, and the signal components (or spikes) are themselves random variables. We adopt this terminology in parallel with the quadratic network case, but when $n\Lambda_i$ is comparable to $d\hat\Delta$, the signal and noise contribute equally to the variance of $u_i$, so a spike cannot be distinguished from the bulk. However, for clarity of exposition, we will nevertheless classify the indices according to whether $n\Lambda_i$ is larger or smaller than $d\hat\Delta$, acknowledging that this introduces an asymptotically negligible ambiguity near the crossover. \\ The $i^{\rm th}$ components of the LASSO estimator are then obtained by soft-thresholding the variables $u_i$, with the parameter $\epsilon_{\rm d}$ representing a cutoff that induces sparsity in the estimator, setting the smallest values to zero. Note that the cutoff depends on the regularization strength $\lambda$ only through $\nu$.

At this level, we can distinguish the following scenarios:
in terms of number of data,
\begin{table}[H]
    \centering
    \begin{tabular}{c|c}
    {\bf spikes + bulk} & {\bf only spikes}\\
      $n\Lambda_d\ll d\hat \Delta$  & $n\Lambda_d\gg d\hat \Delta$ \\\hline
      not all components can be learned   & all components can be learned
    \end{tabular}
\end{table}
in terms of thresholding strength,
\begin{table}[H]
    \centering
    \begin{tabular}{c|c|c}
     {\bf weak} & {\bf strong} & {\bf extreme}\\
       $\nu^2\ll \max(n\Lambda_d/d,\hat\Delta)$   &  $\max(n\Lambda_d/d,\hat\Delta)\ll\nu^2\ll n\Lambda_1/d$&  $\nu^2\gg n\Lambda_1/d$ \\\hline
       cutoff below spikes:  & cutoff between spikes: & cutoff above all spikes:\\
      all learnable signal is learned & signal is partially learned, noise is filtered &  nothing is learned
    \end{tabular}
\end{table}

\paragraph{Error decomposition} We can interpret the excess risk expression in eq. (\ref{eq:app:ERM_SE_linear_generic}) in light of these considerations on the interplay between signal, noise bulk and regularization, by splitting the sum at the relevant scales identified in this section, namely $i_{\nu}$ such that $n\Lambda_{i_{\nu}} = d\nu^2$ and $i_{\hat\Delta}$ such that $n\Lambda_{i_{\hat\Delta}} = d\hat\Delta$. Note that the extreme regularization case $\nu^2 \gg \sqrt{n}$ is trivial, as the excess risk $\mathsf R \approx \sum_{i=1}^d i\Lambda_i/d$, {\it i.e.} it is dominated by the underfitting of all the signal components.

\begin{enumerate}
    \item Weak thresholding: the excess risk can be decomposed as $\mathsf R \approx \mathsf R_{\rm under} + \mathsf R_{\rm over}+ \mathsf R_{\rm app}$
    \begin{itemize}
        \item Underfitting term (signal components not learned)
        \begin{equation}
           \mathsf R_{\rm under} = \frac{1}{d} \sum_{i_{\hat \Delta} + 1}^d \Lambda_i\operatorname{erf}\left(\frac{\nu}{\sqrt{\hat\Delta }}\right) \approx  \frac{1}{d} \sum_{i_{\hat \Delta} + 1}^d \Lambda_i ,
        \end{equation}
        \item Overfitting term (learned noise)
        \begin{align}
            \mathsf R_{\rm over}& =\frac{1}{n}\sum_{i=i_{\hat\Delta}+1}^d \left[ \left(\hat\Delta+2\nu^2\right)\operatorname{erfc}\left(\frac{\nu}{\sqrt{ \hat\Delta }}\right)- \frac{2\nu}{\sqrt{\pi}}\sqrt{\hat\Delta}\,e^{-\nu^2/\hat\Delta}\right], \\
            &= \frac{d-i_{\hat\Delta}-1}{n}\, \E\left[\left( {\rm ST}_{\varepsilon_{\rm d}}\left(z_i\sqrt{\frac{\hat\Delta d}{n}}\right)\right)^2\right],
        \end{align}   
        \item Approximation term (learned signal)
        \begin{align}
            \mathsf R_{\rm app} =& \frac{1}{n}\sum_{i=1}^{i_{\hat \Delta}} \left[\frac{n}{d}\Lambda_i\operatorname{erf}\left(\frac{\nu}{\sqrt{\frac{n}{d}\Lambda_i+ \hat\Delta }}\right) + \left(\hat\Delta+2\nu^2\right)\operatorname{erfc}\left(\frac{\nu}{\sqrt{ \frac{n}{d}\Lambda_i+\hat\Delta }}\right) \right]\\
   & - \frac{2\nu}{n\sqrt{\pi}}\sum_{i=1}^{i_{\hat \Delta}}\left[\sqrt{\frac{n}{d}\Lambda_i+\hat\Delta}\,e^{-\nu^2/(\frac{n}{d}\Lambda_i+\hat\Delta)}\right]    
        \end{align}

    \end{itemize}
    \item Strong thresholding: the excess risk can be decomposed as
    $\mathsf R \approx \mathsf R_{\rm under} + \mathsf R_{\rm app}$
    \begin{itemize}
        \item Underfitting term (signal components not learned)
        \begin{equation}
           \mathsf R_{\rm under} = \frac{1}{d} \sum_{i_{\nu} + 1}^d \Lambda_i\operatorname{erf}\left(\frac{\nu}{\sqrt{n\Lambda_i }}\right)\approx \frac{1}{d} \sum_{i_{\nu} + 1}^d \Lambda_i,
        \end{equation}
        \item Approximation term (learned signal)
        \begin{align}
            \mathsf R_{\rm app} =& \frac{1}{n}\sum_{i=1}^{i_{\nu}} \left[\frac{n}{d}\Lambda_i\operatorname{erf}\left(\frac{\nu}{\sqrt{\frac{n}{d}\Lambda_i+ \hat\Delta }}\right) + \left(\hat\Delta+2\nu^2\right)\operatorname{erfc}\left(\frac{\nu}{\sqrt{ \frac{n}{d}\Lambda_i+\hat\Delta }}\right) \right]\\
   & - \frac{2\nu}{n\sqrt{\pi}}\sum_{i=1}^{i_{\nu}}\left[\sqrt{\frac{n}{d}\Lambda_i+\hat\Delta}\,e^{-\nu^2/(\frac{n}{d}\Lambda_i+\hat\Delta)}\right]    
        \end{align}

    \end{itemize}
\end{enumerate}

In Section \ref{app:sec:lasso_rates} we observe that these are the relevant scales for computing the leading order terms of the excess risk and its scaling laws. Moreover, we estimate the values of $\nu$ and $\hat\Delta$ as functions of $n,d,\lambda$. These results,  together with the interpretation provided in this section, lead to the identification of the phase diagram regions in fig.~\ref{fig:phase} and the phases descriptions in Section \ref{sec:spectral_properties}.

\subsection{LASSO scaling laws} \label{app:sec:lasso_rates}
From eq.~(\ref{eq:app:ERM_SE_linear}),
\begin{align}
   \mathsf R =  \frac{1}{n}\sum_{i=1}^d[f_1(x_i) + f_2(x_i) - f_3(x_i)],
\end{align}
with $x_i := i n^{-1/(2\gamma)}$ and
\begin{align}
f_1(x) &= x^{-2\gamma}\operatorname{erf}\left(\frac{\nu}{\sqrt{x^{-2\gamma}+\hat\Delta}}\right)\\
f_2(x)&=(\hat\Delta+2\nu^2)\operatorname{erfc}\left(\frac{\nu}{\sqrt{x^{-2\gamma}+\hat\Delta}}\right)\\
f_3(x)&= \frac{2}{\sqrt{\pi}}\nu\sqrt{x^{-2\gamma} + \hat\Delta}\exp\left(-\frac{\nu^2}{x^{-2\gamma}+\hat\Delta}\right)
\end{align}
We observe that the leading order of the functions changes scale around $x\sim \hat\Delta^{-1/(2\gamma)}$.\\
For $x^{-2\gamma}\gg \hat \Delta$
    \begin{align}
    f_1(x) &\approx x^{-2\gamma}\operatorname{erf}\left(\nu x^\gamma\right)\\
    f_2(x)&\approx (\hat\Delta+2\nu^2)\operatorname{erfc}\left(\nu x^\gamma\right)\\
    f_3(x)&\approx \frac{2}{\sqrt{\pi}}\nu x^{-\gamma} \exp\left(-\nu^2 x^{2\gamma}\right)
    \end{align}
For $x^{-2\gamma}\ll \hat \Delta$
    \begin{align}
    f_1(x) &\approx x^{-2\gamma}\operatorname{erf}\left(\nu \hat\Delta^{-1/2}\right)\\
    f_2(x)&\approx (\hat\Delta+2\nu^2)\operatorname{erfc}\left(\nu \hat\Delta^{-1/2}\right)\\
    f_3(x)&\approx \frac{2}{\sqrt{\pi}}\nu \hat\Delta^{1/2} \exp\left(-\nu^2 \hat\Delta^{-1}\right).
    \end{align}
Note that the scale $x_i^{-2\gamma}\sim \hat\Delta $ corresponds precisely to the detectability threshold of signal components observed in Section \ref{app:sec:lasso_spectral} for the LASSO estimator and in Section \ref{app:sec:BO_rates} for the Bayes-optimal estimator.\\
We compute the excess risk $R$ from eq. (\ref{eq:app:ERM_SE_linear}) at leading order as a function of the parameter $\nu$, by splitting the sums at the crossover scales. Note that this procedure corresponds to the decomposition of the excess risk into approximation, overfitting and underfitting errors that we have identified in Section \ref{app:sec:lasso_spectral}. Afterwards, solving the self-consistent eq. (\ref{eq:app:self_consistent_nu}) for $\nu$, we derive the Results in Section \ref{sec:phase_diagrams}. \\
The three main regimes are 
\begin{align}\label{app:eq:lasso_regimes}
    \begin{cases}
        \nu \gg x_1^\gamma \implies \nu^2\gg n,\\
         \sqrt{\max(x_d^{-2\gamma},\hat\Delta)}\ll \nu \ll x_1^\gamma \implies \max(nd^{-2\gamma},\hat\Delta)\ll\nu^2\ll n,\\
         \nu\ll \sqrt{\max(x_d^{-2\gamma},\hat\Delta)}\implies \nu^2\ll \max(nd^{-2\gamma},\hat\Delta),
    \end{cases}
\end{align}
which are the extreme, strong and weak thresholding phases we have identified in Section \ref{app:sec:lasso_spectral}, for the specific choice of covariance $\Lambda_i = di^{-2\gamma}$ .\\ 
In what follows we will often use the expansions
\begin{align}
    \operatorname{erf}(x \ll 1) &= \frac{2}{\sqrt{\pi}}x + o(x),\\
    \operatorname{erfc}(x\gg1) &= \frac{e^{-x^2}}{x\sqrt{\pi}}\left(1 - \frac{1}{2x^2} + \frac{3}{4x^4} + o(x^{-4})\right). 
\end{align}
\paragraph{Extreme thresholding}
For $\nu^2 \gg n$, the dominant term is given by underfitting 
\begin{align}
    \frac{1}{n}\sum_{i=1}^{d} f_1(x_i)\approx  \frac{1}{n}\sum_{i=1}^d (x_i)^{-2\gamma} \approx \zeta(2\gamma),
\end{align}
while the remaining terms are negligible
\begin{align}
    \frac{1}{n}\sum_{i=1}^df_2(x_i)-f_3(x_i)\approx&  \frac{\hat\Delta}{\nu\sqrt{n\pi}}\sum_{i=1}^{\min(d,\lfloor(\hat\Delta/n)^{-1/(2\gamma)}\rfloor)} 
    i^{-\gamma}e^{-i^{2\gamma}\nu^2/n} \\
    \approx&  \frac{\hat\Delta}{\nu\sqrt{n\pi}}\exp(-\nu^2/n).
\end{align}
As expected, in this regime the large soft-thresholding parameter forces the components of the estimator to zero, and $R \approx \zeta(2\gamma)$.

\paragraph{Strong thresholding} 
If instead $\max(nd^{-2\gamma},\hat\Delta)\ll \nu^2 \ll n$, defining $i_\nu := \lfloor(n/\nu^2)^{1/(2\gamma)}\rfloor$ and $i_{\hat\Delta}:=\lfloor(n/\hat\Delta)^{1/(2\gamma)}\rfloor$, we split the sums into four terms:\\
\begin{itemize}
 \item approximation error:
\begin{align*}
\begin{array}{l}
\begin{aligned}
    {\rm i)}\,\frac{1}{n}\sum_{i=1}^{i_\nu } f_1(x_i) - f_3(x_i) \approx&  \frac{2\nu}{\sqrt{n\pi}}\sum_{i=1}^{i_\nu } (1-\exp(-i^{2\gamma}\nu^2/n))i^{-\gamma}\\
    \approx & \frac{2\nu^3}{n\sqrt{n\pi}}\sum_{i=1}^{i_\nu }i^\gamma\\
    \approx & \frac{2}{(1+\gamma)\sqrt{\pi}}\left(\frac{n}{\nu^2}\right)^{-1+1/(2\gamma)}
\end{aligned}\\
\begin{aligned}
    {\rm ii)}\,\frac{1}{n}\sum_{i=1}^{i_\nu } f_2(x_i)\approx& \frac{2\nu^2+\hat\Delta}{n}\sum_{i=1}^{i_\nu } \left[1- \operatorname{erf}(i^\gamma \nu/\sqrt{n})\right]
\end{aligned}\\
\end{array}
\end{align*}
 \item underfitting error:
\begin{align*}
\begin{array}{l}
\begin{aligned}
    {\rm iii)}\,\frac{1}{n}\sum_{i=i_\nu +1}^d f_1(x_i) \approx & \sum_{i=i_\nu +1}^d i^{-2\gamma} \approx \frac{1}{2\gamma-1}\left(\frac{n}{\nu^2}\right)^{-1+1/(2\gamma)}\\
    \approx& 2\left(\frac{n}{\nu^2}\right)^{-1+1/(2\gamma)} + \hat\Delta \nu^{-1/\gamma}n^{-1+1/(2\gamma)} + \Theta\left(\left(\frac{n}{\nu^2}\right)^{-1+1/(2\gamma)}\right)
\end{aligned}\\
\begin{aligned}
 {\rm iv)}\,\frac{1}{n}\sum_{i=i_\nu +1}^d f_2(x_i)-f_3(x_i)\approx &  \frac{\hat\Delta}{\nu\sqrt{n\pi}}\sum_{i=i_\nu +1}^{\min(d,i_{\hat\Delta})}i^{-\gamma}e^{-i^{2\gamma}\nu^2/n}+\boldsymbol{1}_{[\hat\Delta > nd^{-2\gamma}]}\frac{d\hat\Delta^{5/2}}{n\nu^3}e^{-\nu^2/\hat\Delta}\\
 \stackrel{\rm Laplace}{\approx} & \frac{\hat\Delta}{e\sqrt{\gamma(2\gamma-1)}}n^{-1+1/(2\gamma)}\nu^{-1/\gamma}+\boldsymbol{1}_{[\hat\Delta > nd^{-2\gamma}]}\frac{d\hat\Delta^{5/2}}{n\nu^3}\exp\left(-\frac{\nu^2}{\hat\Delta}\right)
\end{aligned}
\end{array}
\end{align*}
\end{itemize}
where in the last step we have approximated the (Riemann) sum by an integral which we solved using the Laplace's method, that is (informally)
\begin{align}
    \int_a^b h(x)e^{Mg(x)}\de x \overset{M\gg 1}{\approx} \sqrt{\frac{2\pi}{M|g''(x_0)|}}h(x_0)e^{Mg(x_0)},\qquad x_0 = \arg\max_{x\in[a,b]}g(x),\;\; g''(x)\leq 0 \forall x\in[a,b].
\end{align}
The dominant term is $\mathsf R \asymp (n/\nu^2)^{-1+1/(2\gamma)}$.

\paragraph{Weak thresholding}
Finally , if $\nu^2 \ll \max(nd^{-2\gamma},\hat\Delta)$, we split the sums into the following four terms:
\begin{itemize}
\item approximation error:
\begin{align}
    \begin{array}{l}
     \begin{aligned}
      {\rm i)}\,  \frac{1}{n}\sum_{i=1}^{\min(i_{\hat\Delta},d)} f_1(x_i) - f_3(x_i) \approx & \frac{2\nu^3}{(1+\gamma)\sqrt{n\pi}}\left(\boldsymbol{1}_{[\hat\Delta > nd^{-2\gamma}]}\left(\frac{n}{\hat\Delta}\right)^{(1+\gamma)/(2\gamma)}+ \boldsymbol{1}_{[\hat\Delta < nd^{-2\gamma}]}d^{1+\gamma}\right)
     \end{aligned}\\
     \begin{aligned}
       {\rm ii)}\,  \frac{1}{n}\sum_{i=1}^{\min(d,i_{\hat\Delta})} f_2(x_i) \approx & (2\nu^2+\hat\Delta)\min\left(n^{-1+1/(2\gamma)}\hat\Delta^{-1/(2\gamma)},\frac{d}{n}\right)
     \end{aligned}\\
 \end{array}
 \end{align}
\item overfitting + underfitting errors:
 \begin{align}
 \begin{array}{l}
     \begin{aligned}
         {\rm iii)}\, \frac{1}{n}\sum_{i=\min(d,i_{\hat\Delta}+1)}^{d} f_1(x_i) \approx& \boldsymbol{1}_{[\hat\Delta > nd^{-2\gamma}]}\frac{2\nu}{\sqrt{\hat\Delta\pi}}\sum_{i=i_{\hat\Delta}+1}^{d} i^{-2\gamma}\\
         \approx& \boldsymbol{1}_{[\hat\Delta > nd^{-2\gamma}]}\frac{2\nu}{\sqrt{\hat\Delta\pi}}\left(\frac{n}{\hat\Delta}\right)^{-1+1/(2\gamma)}
     \end{aligned}\\
     \begin{aligned}
       {\rm iv)}\,   \frac{1}{n}\sum_{i=\min(d,i_{\hat\Delta}+1)}^{d}f_2(x_i) -f_3(x_i)\approx & \boldsymbol{1}_{[\hat\Delta > nd^{-2\gamma}]}\left[(2\nu^2 +\hat\Delta) \left(\frac{d}{n}-n^{-1+1/(2\gamma)}\hat\Delta^{-1/(2\gamma)}\right)\right]\\&- \boldsymbol{1}_{[\hat\Delta > nd^{-2\gamma}]}\left[\frac{2}{\sqrt{\pi}}\nu\sqrt{\hat\Delta}\exp(-\nu^2/\hat\Delta)\frac{d}{n}\right]
     \end{aligned}          
    \end{array}
\end{align}
\end{itemize}
The dominant term is therefore $\mathsf R \asymp (2\nu^2 + \hat \Delta)d/n$. \\
Note that, for $\nu = \Tilde{\Theta}(\hat\Delta)$, even if the cutoff is larger than the noise variance, it is not large enough to completely filter all bulk terms. We will verify that this is the case in Phases IV and V of Fig. \ref{fig:phase}, for $\lambda\ll\sqrt{n/d}$ and $n\ll d$. The leading order terms in the excess error decomposition are equivalent to the ones we have computed in the previous paragraph for the {\bf Strong thresholding} case. However, we must take into account the overfitting contribution from the bulk components $\frac{d\hat\Delta^{5/2}}{n\nu^3}\exp\left(-\frac{\nu^2}{\hat\Delta}\right)$ (term iv), which is otherwise negligible for $\nu^2/\hat\Delta$ larger than any polylogarithmic function of $n,d$.\\

We can now proceed with the solution of the self-consistent equation (\ref{eq:app:self_consistent_nu}), in order to derive the closed-form expressions for the excess risk scaling laws.
\paragraph{Noisy setting} Since $\Delta>0$, then, provided $R = O(1)$, $\hat\Delta = \Theta(1)$. Let $n\gg d$. Eq. (\ref{eq:app:self_consistent_nu}) readily implies that $\nu \asymp \lambda \sqrt{d/n}$, as the second term on the left-hand side is bounded by 1. Therefore, for $n\gg d$,
\begin{align}
   \mathsf R = \begin{cases}
        \Theta(1), &\lambda\gg n/\sqrt{d},\\
        \Theta\left(\frac{n^2}{\lambda^2d}\right)^{-1+1/(2\gamma)},& \max\left(nd^{-\gamma-1/2},\sqrt{n/d}\right)\ll\lambda\ll n/\sqrt{d}\\
        \Theta\left(\frac{d}{n}\max\left(1,\frac{\lambda^2d}{n}\right)\right),& \lambda\ll\max\left(nd^{-\gamma-1/2},\sqrt{n/d}\right)
    \end{cases}
\end{align}
Let instead $n\ll d$. The second term on the left-hand side of eq. (\ref{eq:app:self_consistent_nu}) is
\begin{align}
    \frac{1}{d}\sum_{i=1}^d \operatorname{erfc}\left(\frac{\nu}{\sqrt{x^{-2\gamma}+\hat\Delta}}\right) \approx \begin{cases}
        \sqrt{\frac{n}{\pi}}\nu^{-1}\exp\left(-\nu^2/n\right), & \nu^2\gg n,\\
        2\frac{n^{1/(2\gamma)}}{d}\left(\nu^{2-1/\gamma}\right)(1 + O(1)) + \operatorname{erfc}\left(\frac{\nu}{\hat\Delta}\right),& 1\ll\nu^2\ll n,\\
        1, &\nu\ll 1.
    \end{cases}
\end{align}

For $\lambda\gg \sqrt{n/d}$, this term is subleading and $\nu\asymp \lambda \sqrt{d/n}$. One can observe that eq. (\ref{eq:app:self_consistent_nu}) does not have a positive solution in this regime if $\nu\ll 1$, therefore we exclude this case. Hence, if $\lambda\ll\sqrt{n/d}$,
\begin{align}
    \operatorname{erfc}\left(\frac{\nu}{\hat\Delta}\right) \approx \frac{n}{d}\implies& \sqrt{\frac{\hat\Delta}{\pi}}\nu^{-1} e^{-\nu^2/\hat\Delta}\approx n/d \\
    \implies& \frac{\nu^2}{\hat\Delta} \approx \log \frac{d}{n},
\end{align}
where we used
\begin{align}
    xe^{x^2} = a \implies 2x^2e^{2x^2} = 2a^2 \implies x^2 = \frac{1}{2}{\rm W}_0(2a^2) \stackrel{a\gg 1}{\approx}  \frac{1}{2}\log (2a^2) \approx \log a,
\end{align}
with ${\rm W}_0$ denoting the principal branch of the Lambert W function. Note that, in the under-parametrized regime $n\ll d$, even a small regularization strength $\lambda$ leads to large effective regularization $\nu$ of order $\sqrt{\log d}$.\\

We can conclude that, for $n\ll d$
\begin{align}\label{app:eq:lasso_scaling_laws_noisy}
   \mathsf R = \begin{cases}
        \Theta(1), &\lambda\gg n/\sqrt{d},\\
        \Theta\left(\frac{n^2}{\lambda^2d}\right)^{-1+1/(2\gamma)},& \max\left(nd^{-\gamma-1/2},\sqrt{n/d}\right)\ll\lambda\ll n/\sqrt{d}\\
        \Theta\left(\left(\frac{n}{\log(d/n)}\right)^{-1+1/(2\gamma)} + \frac{\Delta}{\log(d/n)}\right),& \lambda\ll\sqrt{n/d}
    \end{cases}
\end{align}
\paragraph{Noiseless setting} If $\Delta = 0$, then $\hat\Delta = \mathsf R$. Let $n\gg d$. Similarly to the noisy setting, it is straightforward to show that $\nu\asymp\lambda\sqrt{d/n}$ in eq. (\ref{eq:app:self_consistent_nu}), implying $\nu \approx \lambda \sqrt{d/n}$. Therefore, for $n\gg d$
\begin{align}
     \mathsf R =
     \begin{cases}
        \Theta(1), &\lambda^2 \gg \frac{n^2}{d}\\
        \Theta\left(\left(\frac{n^2}{\lambda^2 d}\right)^{-1+1/(2\gamma)}\right), & \frac{n^2}{d^{1+2\gamma}}\ll\lambda^2\ll \frac{n^2}{d}\\
        \Theta\left(\frac{\lambda^2d^2}{n^2}\right), & \lambda^2 \ll  \frac{n^2}{d^{1+2\gamma}}.
    \end{cases}    
\end{align}
Let instead $n\ll d$. In this regime, the Bayes optimal generalization error $\mathsf R \asymp n^{-2\gamma + 1}\gg nd^{-2\gamma}$, which implies $\hat\Delta=\mathsf R\gg nd^{-2\gamma}$. Similarly to the noisy setting, the second term on the left-hand side of eq. (\ref{eq:app:self_consistent_nu}) is
\begin{align}
    \frac{1}{d}\sum_{i=1}^d \operatorname{erfc}\left(\frac{\nu}{\sqrt{x^{-2\gamma}+R}}\right) \approx \begin{cases}
        \sqrt{\frac{n}{\pi}}\nu^{-1}\exp\left(-\nu^2/n\right), & \nu^2\gg n,\\
        2\frac{n^{1/(2\gamma)}}{d}\left(\nu^{2-1/\gamma}\right)(1 + \Theta(1)) + \operatorname{erfc}\left(\frac{\nu}{R}\right),& R\ll\nu^2\ll n,\\
        1, &\nu^2\ll R.
    \end{cases}
\end{align}
If $\nu^2\gg n$, this term is negligible and $\nu\asymp\lambda\sqrt{d/n}$.
For $\mathsf R\ll\nu^2\ll n$, if this term is negligible with respect to $n/d$, and also in this case $\nu\asymp\lambda\sqrt{d/n}$ with $\mathsf R = \Theta(n^2/(\lambda d))^{-1+1/(2\gamma)}$ as in the $n\gg d$ regime. The condition $\nu^2\gg \mathsf R$ becomes 
\begin{align}
\frac{\lambda^2 d }{n}\gg \frac{\lambda^{2-1/\gamma} d^{1-1/(2\gamma)}}{n^{2-1/\gamma}} \implies \lambda \gg \frac{1}{n^{\gamma-1}d^{1/2}}.
\end{align}
For consistency, we verify that the second term on the lhs of (\ref{eq:app:self_consistent_nu}) is negligible with respect to $n/d$:
\begin{align}
    \frac{n^{1/(2\gamma)}}{d}\nu^{2-1/\gamma}\ll \frac{n}{d}\implies \nu \ll n.
\end{align}
Finally, if $\lambda \gg {n^{-\gamma+1}d^{-1/2}}$, repeating the same computations of the noisy setting, we find $\nu^2 \approx \mathsf R \log(d/n)$ and 
\begin{align}
    \mathsf R \approx \left(\frac{n}{\mathsf R \log(d/n)}\right)^{-1+1/(2\gamma)} + \frac{\mathsf R}{\log (d/n)} \implies & \left(1-\frac{1}{\log(d/n)}\right)\mathsf R^{1/(2\gamma)} \approx  \left(\frac{n}{\log(d/n)}\right)^{-1+1/(2\gamma)}\\
    \implies& \mathsf R \approx  n^{-2\gamma+1}(\log(d/n))^{1-1/(2\gamma)}\left(1-\frac{1}{\log(d/n)}\right)^{-2\gamma} \label{app:eq:noiseless_log_factors}\\
    \implies& \mathsf R = \Tilde{\Theta}\left(n^{-2\gamma+1}\right).
\end{align}
In conclusion, for $n\ll d$,
\begin{align}
     \mathsf R =
     \begin{cases}
        \Theta(1), &\lambda^2 \gg \frac{n^2}{d}\\
        \Theta\left(\left(\frac{n^2}{\lambda^2 d}\right)^{-1+1/(2\gamma)}\right), &{n^{-2\gamma+2}d^{-1}}\ll\lambda^2\ll \frac{n^2}{d}\\
        \Tilde{\Theta}\left(n^{-2\gamma+1}\right), & \lambda^2 \ll  {n^{-2\gamma+2}d^{-1}}.
    \end{cases}    
\end{align}

\paragraph{Interpolation peak} Around $n\sim d$ the excess risk exhibits an interpolation peak that diverges as $\lambda\to0^+$. In this section we show that in this regime $R\asymp \lambda^{-2/3}$.\\
The self-consistent equation (\ref{eq:app:self_consistent_nu}) becomes
\begin{align}
    \frac{\lambda}{\nu} \approx \frac{\sqrt 2}{d}\sum_{i=1}^d \operatorname{erf}\left(\frac{\nu}{\sqrt{ni^{-2\gamma} + \hat \Delta}}\right).
\end{align}
As we have done in the previous paragraph, it is easy to verify that a strong effective regularization $\nu^2\gg\hat\Delta$ results in the right-hand side being $\Theta(1)$, and to the inconsistent solution $\nu\sim\lambda$. Hence, taking $\nu^2\ll\hat\Delta$ (weak thresholding regime), 
\begin{align}
    \frac{\lambda}{\nu}\approx \frac{2\sqrt2\nu}{\sqrt\pi\hat\Delta}\implies \nu^2\approx \frac{\sqrt\pi}{2\sqrt 2}\frac{\lambda}{\hat\Delta}
\end{align}
$\nu^2\asymp \lambda\sqrt{\hat\Delta}$ and the excess risk becomes, asymptotically,
\begin{align}
    &\mathsf R \approx 2\nu^2 + R + \Delta - \frac{4}{\sqrt{\pi}}\nu\sqrt{\mathsf R+\Delta}\\
    \implies&\nu\sqrt{\mathsf R+\Delta}\approx \frac{\sqrt{\pi}}{4}\Delta\\
    \implies& \lambda^2(\mathsf R+\Delta)^3 \approx \frac{\pi^2}{256}\Delta^4\\
    \implies& \mathsf R\asymp \frac{\Delta^{4/3}}{\lambda^{2/3}}.
\end{align}

\subsubsection{Additional numerical simulations}\label{app:sec:lasso_additional_numerics}
\begin{figure}[H]
    \centering
    \includegraphics[width=1\linewidth]{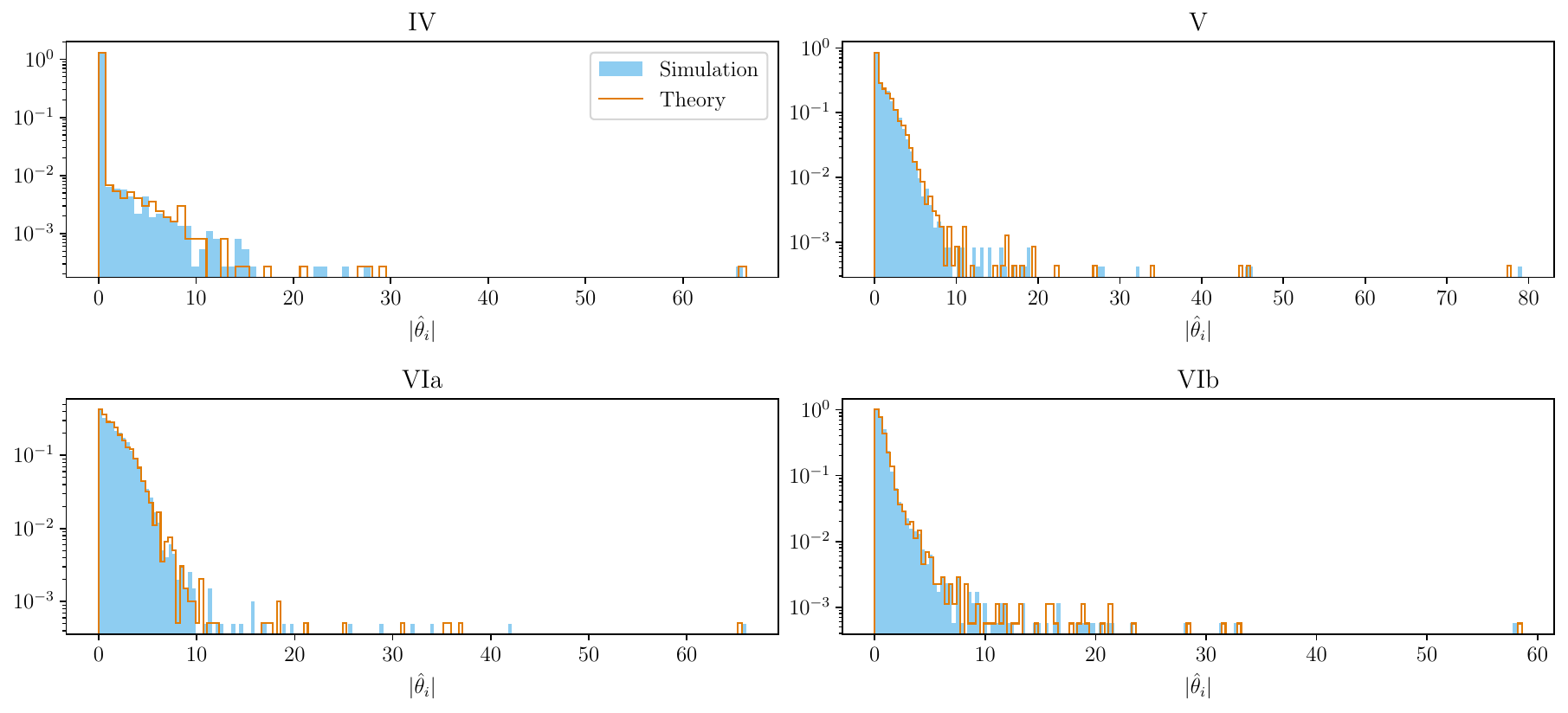}
    \caption{Comparison between spectra from simulations and theory across different training phases. Blue: LASSO estimator's components (in absolute value) histograms after training. Red: theoretical prediction eq. (\ref{app:eq:lasso_result4}). All panels use $d\!=\!5000$ and $\lambda\!=\!d^{-1/2}$. The sample size is $n=200$ for Phase IV, $n=4000$ for Phase V, $n=6000$ for Phase VIa and $n=4\cdot 10^4$ for Phase IVb. We discuss the phenomenology in Section \ref{res:interpretability}.}
    \label{fig:lasso_spectra_lowreg}
\end{figure}

\begin{figure}[H]
    \centering
    \includegraphics[width=1\linewidth]{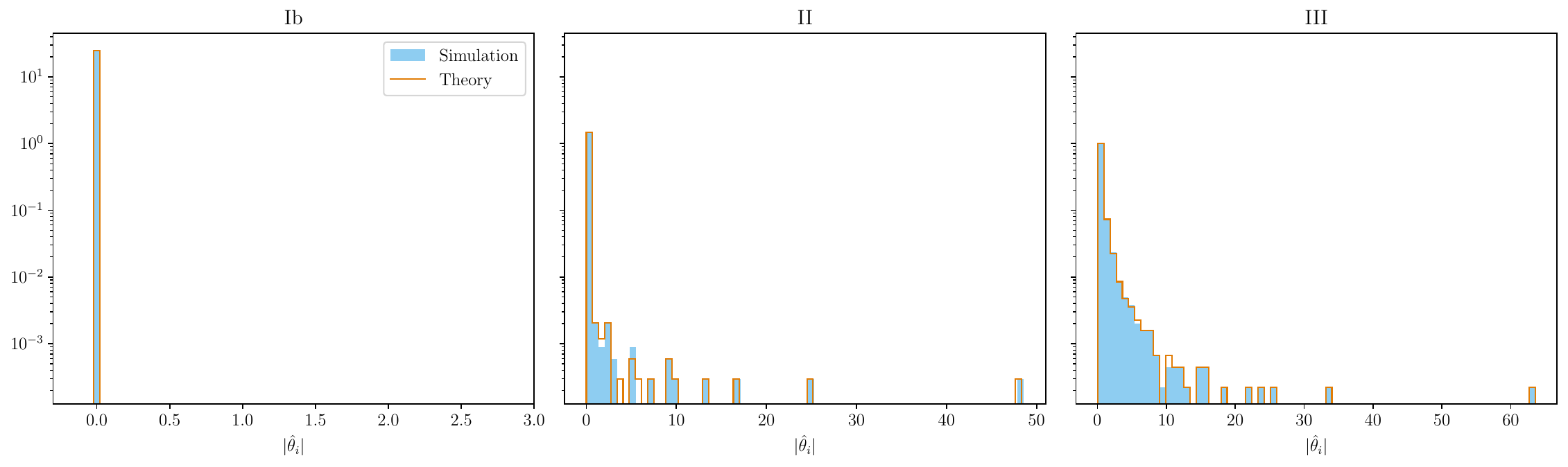}
    \caption{Comparison between spectra from simulations and theory across different training phases. Blue: LASSO estimator's components (in absolute value) histograms after training. Red: theoretical prediction eq. (\ref{app:eq:lasso_result4}). All panels use $d\!=\!5000$;  Phase Ib: $n=200$, $\lambda =2 n/\sqrt{d}$; Phase II: $n=4000$, $\lambda = 10$; for Phase III: $n=4\cdot 10^4$. $\lambda = 3$. We discuss the phenomenology in Section \ref{res:interpretability}.}
    \label{fig:lasso_spectra_largereg}
\end{figure}

\begin{figure}[H]
    \centering
    \includegraphics[width=0.49\linewidth]{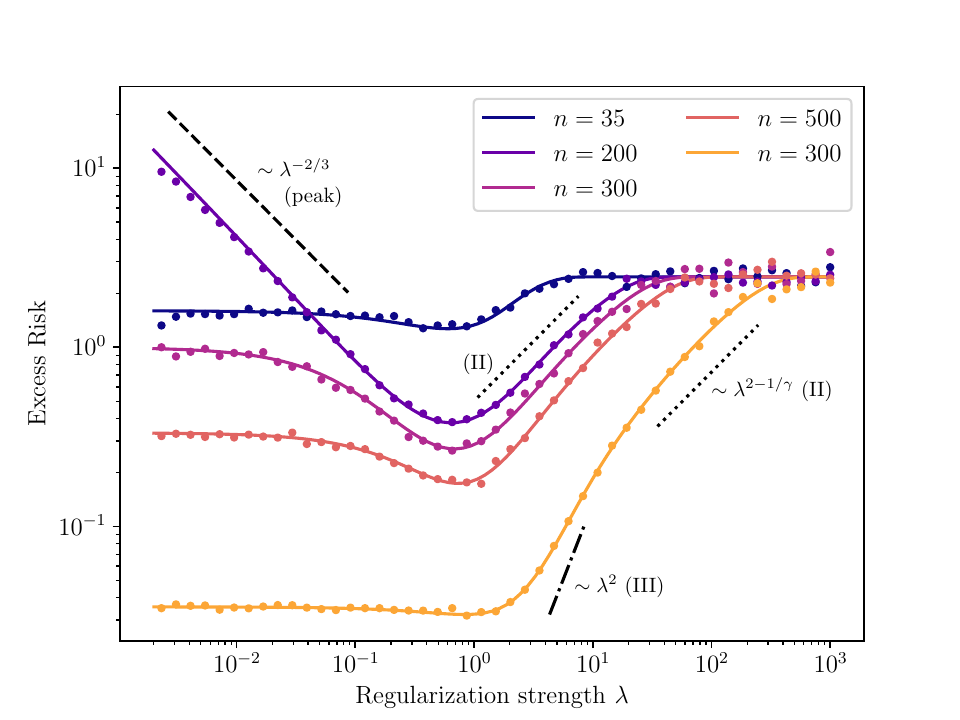}
     \includegraphics[width=0.49\linewidth]{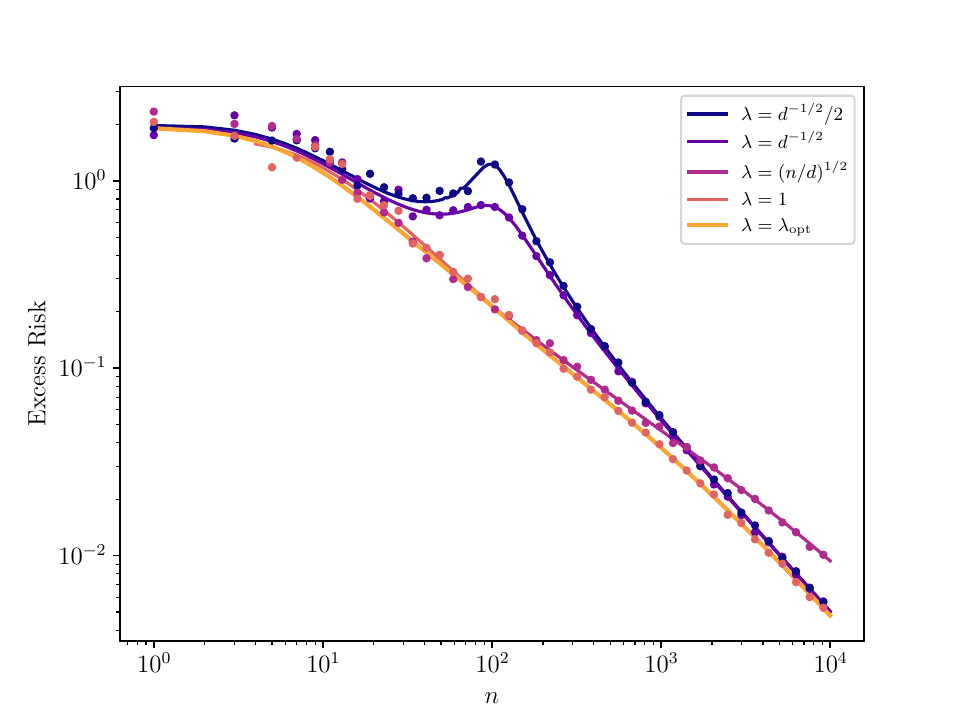}
    \caption{({\bf Left}) Excess risk of the LASSO estimator, as a function of the regularization strength $\lambda$, with $d = 200$, $\Delta = 0.5$, $\gamma = 0.75$. Dots represent numerical experiments, while lines the solution ${\mathsf R}_{n,d}$ of state evolution equations \ref{eq:app:ERM_SE_linear}. The curves correspond to the crossovers between rates observed in Fig.~\ref{fig:phase}. ({\bf Right}) Excess risk of the LASSO estimator, as a function of the sample size $n$, with $d = 100$, $\Delta = 0.5$, $\gamma = 1$. The regularization $\lambda_{\rm opt}$ has been chosen as the minimizer of the theoretical excess risk ${\mathsf R}_{n,d}$ and its value is in accordance with Corollary \ref{cor:optlambda}. Dots represent numerical experiments, while lines the solution of state evolution equations \ref{eq:app:ERM_SE_linear}.}
    \label{fig:lasso_reg}
\end{figure}

In this section, we include additional numerical experiments, visualizing the Results in Sections \ref{sec:phase_diagrams} (Fig.~\ref{fig:lasso_reg}) and \ref{sec:spectral_properties} (Figs.~\ref{fig:lasso_spectra_lowreg},\ref{fig:lasso_spectra_largereg}). The remarkable correspondence between simulations and all results derived from state evolution equations further supports our Conjecture \ref{conj:diagonal}. 

Figure~\ref{fig:lasso_reg} (left) shows the transitions between phases along the vertical lines of Fig.~\ref{fig:phase}.  
For $n = 35$ the excess risk moves from Phase~IV, which is independent of regularization, into Phase~II and soon ($\lambda \approx 35/\sqrt{200}$) enters the plateau region of Phase~Ib.  
For $n = 300$ and $n = 500$, the excess risk starts in the fast–decay region~IVa, reaches its minimum at $\lambda \approx \sqrt{n/d}$ (when the soft-thresholding cutoff reaches the edge of the noise bulk, see Section~\ref{sec:spectral_properties}), then crosses Phase~II and enters the plateau Phase~Ib.  
Finally, for $n = 3000$, the excess risk begins in Phase~IVb, the fastest decay regime, since the noise bulk is negligible for $n \gg d^{2\gamma}$, then grows as it crosses Phase~III and Phase~II before reaching the plateau in Phase~Ib.
Notably, for $n = d = 200$, we observe the interpolation phenomenon when $\lambda < \sqrt{n/d} = 1$, with a peak that grows as the predicted $\lambda^{-2/3}$ in the limit $\lambda \to 0^+$.

\subsection{Bayes-optimal excess risk}\label{app:sec:BO_risk}
The Bayesian predictor $\hat y_{\rm BO}$ is given by the expected value of the target function with respect to the posterior distribution $\mathsf P(\vtheta|\mathcal{D})$. Applying Bayes' theorem, the posterior distribution in this setting reads.
\begin{align}
    \mathsf P (\vtheta|\mathcal D) &= \frac{1}{Z(\mathcal D)} \prod_{i=1}^d \mathcal N(\theta_i;0,\Lambda_i) \prod_{\mu=1}^n \mathcal N\left(y_\mu; \frac{1}{\sqrt{d}}\sum_{j=1}^d X_{\mu j}\theta_j, \Delta\right)\\
    &=\mathcal N(\vtheta; \hat{\vtheta}, \mV),
\end{align}
where, recalling the notation $\mX =(\vx_{1},\ldots,\vx_{n})^T\in\R^{n\times d}$ for the covariate matrix, $\vy = (y_1,\ldots,y_n)^T$ for the label vector and $\boldsymbol{\Lambda} = \operatorname{diag}(\Lambda_1,...,\lambda_d)$ for the weights' covariance, 
\begin{align}
    \hat{\vtheta}:= \frac{1}{\sqrt{d}\Delta} \mV \mX^T \vy,\qquad \mV= \left(\boldsymbol{\Lambda}^{-1} +  \frac{1}{d\Delta} \mX^T\mX\right)^{-1}.
\end{align}
Therefore, the Bayesian predictor is $\hat y^{\rm BO}(\vx)=\hat{\vtheta}^T\vx$ and its excess risk is given by
\begin{align}
    R &= \E[(\vx^T\vtheta^\star - \hat y^{\rm BO}(\vx))^2]\\
    &=\E \lVert \vtheta - \E_{\vtheta|\mathcal D}[\vtheta]\rVert^2\\
    &= \Tr \mV.
\end{align}
Leveraging a classical result from random matrix theory \cite{silverstein1995}, we have that, in the high-dimensional limit $n/d\to\infty$, with fixed ratio $n/d$, the excess risk $R$ concentrates around $ \mathsf R^{\rm BO} $, solution of the fixed-point equations
\begin{align}\label{app:eq:BO_MMSE}
   \mathsf R^{\rm BO} &= \frac{1}{d}\sum_{i=1}^d \frac{1}{\Lambda_i^{-1}+d^{-1}\hat q},\qquad \hat q = \frac{n}{\Delta +  \mathsf R^{\rm BO} }\\
    &=\sum_{i=1}^d \frac{1}{i^{2\gamma}+\hat q},\label{app:eq:BO_SE_quasi_sparse}
\end{align}
where eq. (\ref{app:eq:BO_SE_quasi_sparse}) is the excess risk for the specific choice of covariance $\Lambda_i = di^{-2\gamma}$.\\
The same equations can be derived from the state evolution of Bayes-GAMP, {\it i.e.} the GAMP algorithm tailored to compute marginals of the posterior distribution and the Bayes-optimal predictor. The interested reader can find a more detailed discussion in Appendix D of \cite{ranganGAMP}.

\subsection{Bayes-optimal scaling laws}\label{app:sec:BO_rates}

From eq. (\ref{app:eq:BO_MMSE})
\begin{align}
     \mathsf R^{\rm BO}  = \frac{1}{\hat q}\sum_{i=1}^d \frac{1}{1+(\hat q^{-1/(2\gamma)}i)^{2\gamma}},\qquad \hat q = \frac{n}{\Delta +  \mathsf R^{\rm BO} }
\end{align}
Our goal is to derive the leading order of $ \mathsf R^{\rm BO} $ in the asymptotic regime $n,d\gg 1$. The crossover scale at which the leading behavior of the sum's argument changes is given by $\hat q^{-1/(2\gamma)} i_{\hat q} \approx 1 \implies i_{\hat q} = \lfloor\hat q^{1/(2\gamma)}\rfloor$. If $i_{\hat q} \ll d$, we can split the sum at this relevant scale and retain the leading term for each part\footnote{We stress that, throughout the manuscript, the notation $\approx$ denotes equality up to terms that are asymptotically negligible.}
\begin{align}
     \mathsf R^{\rm BO}  &= \frac{1}{\hat q}\left(\sum_{i=1}^{ \lfloor\hat q^{1/(2\gamma)}\rfloor} \frac{1}{1+(\hat q^{-1/(2\gamma)}i)^{2\gamma}} + \sum_{\lfloor\hat q^{1/(2\gamma)}\rfloor+1}^d \frac{1}{1+(\hat q^{-1/(2\gamma)}i)^{2\gamma}}\right)\\
    &\approx \frac{1}{\hat q}\left(\sum_{i=1}^{ \lfloor\hat q^{1/(2\gamma)}\rfloor} 1 + \sum_{\lfloor\hat q^{1/(2\gamma)}\rfloor+1}^d \hat q i^{-2\gamma}\right)\\
    &\approx \hat q^{-1+1/(2\gamma)} + \frac{1}{2\gamma - 1}\hat q^{-1+1/(2\gamma)}\\
    & =\frac{2\gamma}{2\gamma - 1}\hat q^{-1+1/(2\gamma)},
\end{align}
where we approximate 
\begin{align}
 &\int_{i_{\hat q}+1}^{(d+1)} x^{-2\gamma}\de x  \leq\sum_{i_{\hat q}+1}^d i^{-2\gamma} \leq (i_{\hat q}+1)^{-2\gamma}+ \int_{i_{\hat q}+1}^{d\hat q^{-1/(2\gamma)}} x^{-2\gamma}\de x\\
 \implies & \left\lvert\sum_{i_{\hat q}+1}^d i^{-2\gamma} - \frac{\hat q^{-1+1/(2\gamma)}}{2\gamma-1}\right\rvert = o_{\hat q}\left(q^{-1+1/(2\gamma)}\right).
\end{align}
If instead $i_{\hat q}\gg d$
\begin{align}
     \mathsf R^{\rm BO}  \approx \frac{1}{\hat q}\sum_{i=1}^d 1 = \frac{d}{\hat q}
\end{align}
\paragraph{Noisy setting.} Since $\Delta > 0$, assuming $ \mathsf R^{\rm BO}  = O(1)$, the parameter $\hat q \asymp n$ and the Bayes-optimal generalization error  
\begin{align}
     \mathsf R^{\rm BO}  = \begin{cases}
        \Theta(n^{-1+1/(2\gamma)}),&n\ll d^{2\gamma}\\
        \Theta(d/n), &n\gg d^{2\gamma}.
    \end{cases}
\end{align}
\paragraph{Noiseless setting} When $\Delta = 0$, the parameter $\hat q = n/  R^{\rm BO}$ and the Bayes-optimal generalization error  
\begin{align}
     R^{\rm BO} = \begin{cases}
        \Theta(n^{-2\gamma+1}),&n\ll d\\
        0, &n\gg d.
    \end{cases}
\end{align}

\section{Derivation details - quadratic neural networks}
\label{app:quadratic}

\subsection{BBP simplification}
To derive our results, we make the following assumption, which for the moment we do not control rigorously. We assume that for the sake of simpliying the equations that $\bS^{\star}$ has sub-extensive rank, i.e. it has eigenvalues $\{\sqrt{d}i^{-\gamma}\}_{i=1}^{c(d)}$ for $c(d)\ll d$, and zero otherwise. This technical assumption allows for a great simplification of the density $\mu_\delta$ (the spectrum of $\bS^\star+\delta\bZ$, where $\bZ\sim\GOE(d)$), which can be computed with BBP-like techniques as \citep{huang2018mesoscopic}
\begin{align}
    \mu_\delta(x) 
    = \left(1 - \frac{K}{d} \right)(\mu_{\rm sc}+o(1))(x/\delta) / \delta
    + \frac{1}{d} \sum_{i = 1}^{K} \delta( x - f_\delta(\sqrt{d}i^{-\gamma}) ),
\label{eq:BBP-density}
\end{align}
where $K\ll d$ is the largest integer such that $\sqrt{d}K^{-\gamma}\geq\delta$. $f_\delta(x)=x + \delta^2 / x$ is the BBP-like shifting function. We then send $c(d)\to d$ \textit{a posteriori}, after finding that the error scaling does not depend explicitly on $c(d)$.
For the rest of the section, let us redefine $\alpha:=n/d^2$. \\
Also notice that in the main text, we use a different normalization of the eigenvalues, i.e. $Q^\star=1$, while for the derivation using $Q^\star=\sum_{i=0}^{c(d)}i^{-2\gamma}\approx\zeta(2\gamma)$ is more convenient. Under a remapping $\lambda\to\frac{\lambda}{Q^\star},~\Delta\to\frac{\Delta}{Q^\star}$, this does not modify the scalings we find, but only the constants.

\subsection{Bayesian estimator}
\label{app:noisy_Bayes}
In this section we solve eq. (\ref{eq:SE_Bayes}) for $\Delta>0$.

\paragraph{Phase I: Large-samples}

Phase I is $n\gg d^{2\gamma+1}$. The first equation of eq. (\ref{eq:SE_Bayes}) gives $\hat{q}=\Theta(n/d^2)$, and thus $d^{\frac{1}{2}-\gamma}\gg\frac{2}{\sqrt{\hat q}}$, so all the spikes are outside the bulk. We then have
\begin{align}
\frac{4\pi^2}{3\hat{q}}\int\mu_{1/\sqrt{\hat{q}}}(x)^3\rd x\approx\left(1-\frac{c(d)}{d}\right)\frac{4\pi^2}{3}\int\mu_{\rm sc}(x)^3\rd x=1-\frac{c(d)}{d}
\end{align}
by eq. (\ref{eq:BBP-density}), and thus eq. (\ref{eq:SE_Bayes}) gives
\begin{align}
2\alpha-\frac{c(d)}{d}=\frac{2\alpha\Delta}{\Delta+2(Q^\star -q)},
\end{align}
which gives $\mathsf{R}^{\rm BO}:=Q^\star -q=\frac{\Delta c(d)d}{4n}\to\Theta(d^2/n)$.

\paragraph{Phase II: Under-sampling}

Phase II is $d\ll n\ll d^{2\gamma+1}$. We rewrite the integral in eq.  (\ref{eq:SE_Bayes}) as
\begin{align}
\frac{4\pi^2}{3\hat{q}}\int\mu_{1/\sqrt{\hat{q}}}(x)^3\rd x=\frac{4\pi^2}{3}\int\nu(x)^3\rd x=4\pi^2\int \rd x\nu(x)(H[\nu](x))^2,
\end{align}
where $\nu$ denotes the spectrum density of $Z+\sqrt{\hat{q}}\bS^\star$. The first equality is by change of variables and the second quality is from \cite[Lemma C.1]{maillard2022perturbative}. $H[\nu]$ denotes the Hilbert transform of $\nu$.

We further denote $\delta:=\sqrt{\hat{q}d}$. The first equation of eq. (\ref{eq:SE_Bayes}) suggests that $\hat{q}=\Theta(n/d^2)$, and thus $\delta=\Theta(\sqrt{n/d})$. For $d\ll n\ll d^{2\gamma+1}$ we have $d^{-\gamma+\frac{1}{2}}\ll 2\delta\ll\sqrt{d}$, so according to eq. (\ref{eq:BBP-density}) $\nu$ is composed of a semicircle part (denoted as $\nu_0:=(1-K/d)(\mu_{\rm sc}+\tilde\nu)$, where $\tilde\nu$ is calculated in the end of this section and a discrete part ($\{x_i:=f_1(\delta i^{-\gamma})\}_{i=1}^K$, where $K:=\delta^{1/\gamma}$). Thus we have
\begin{align}
\begin{aligned}
&\int \rd x\nu(x)(H[\nu](x))^2=\int \rd x\nu_0(x)H[\nu_0](x)^2+\frac{2}{d}\sum_{i=1}^K\int \rd x\nu_0(x)H[\nu_0](x)\frac{1}{\pi(x-x_i)}\\
&\qquad+\frac{1}{d^2}\sum_{i,j=1}^K\int \rd x\nu_0(x)\frac{1}{\pi^2(x-x_i)(x-x_j)}+\sum_{j=1}^K\frac{1}{d}\left(H[\nu_0](x_j)+\sum_{\substack{i=1 \\ i \neq j}}^K\frac{1}{\pi d(x_j-x_i)}\right)^2.
\end{aligned}
\end{align}
Denote the right side terms as $I_1$ to $I_4$. For the first term we have
\begin{align}
I_1\approx\frac{1}{3}\int \rd x\mu^3_{\rm sc}(x)\left(1-\Theta\left(\frac{1}{d}\sum_{i=K}^{c(d)}\delta^2i^{-2\gamma}\right)\right)(1-3K/d)\approx\frac{1}{4\pi^2}\left(1-\Theta(\delta^{1/\gamma}d^{-1}\right)),
\end{align}
where we use eq. (\ref{eq:bulk_correction2}). Then we can estimate the leading orders of the other three terms. For the second term, we have
\begin{align}
I_2\approx-\frac{2}{d}\int\rd x\mu_{\rm sc}(x)\frac{x}{2\pi}\sum_{i=1}^K\frac{1}{x_i}=o\left(d^{-1}\sum_{i=1}^K\frac{1}{x_i}\right)=o(\delta^{1/\gamma}d^{-1}),
\end{align}
where we use $H[\mu_{\rm sc}]=\frac{x}{2\pi}$ for $x\in[-2,2]$ and the fact that $x\mu_{\rm sc}$ is an odd function. For the third term we have
\begin{align}
I_3=\Theta\left(d^{-2}\sum_{i,j=1}^K\frac{1}{x_ix_j}\right)=\Theta\left(\left(d^{-1}\sum_{i=1}^K\frac{1}{x_i}\right)^2\right)=\Theta(\delta^{2/\gamma}d^{-2}).
\end{align}
$I_4$ is composed of three terms. The first term is
\begin{align}
\frac{1}{d}\sum_{j=1}^K(H[\nu_0](x_j))^2\approx\frac{1}{d}\sum_{j=1}^Kx_j^{-2}\approx\frac{1}{\delta^2d}K^{2\gamma+1}\int_0^1\frac{x^{2\gamma}}{(1+x^{2\gamma})^2}\rd x=\Theta(\delta^{1/\gamma}d^{-1}),
\end{align}
where we use the fact that $H[\mu_{\rm sc}](x)\approx\frac{1}{x}$ for $x\gg1$. The second term of $I_4$ is
\begin{align}
\begin{aligned}
\sum^K_{\substack{i,j=1\\i\neq j}}\frac{2}{d^2}H[\nu_0](x_j)\frac{1}{\pi(x_j-x_i)}&\approx \sum^K_{\substack{i,j=1\\i\neq j}}\frac{2}{\pi d^2}\frac{1}{x_j(x_j-x_i)}\\
&=\sum^K_{\substack{i,j=1\\i\neq j}}\frac{1}{\pi d^2}\left(\frac{1}{x_j(x_j-x_i)}+\frac{1}{x_i(x_i-x_j)}\right)\\
&=\sum^K_{\substack{i,j=1\\i\neq j}}\frac{1}{\pi d^2x_ix_j}=\frac{1}{\pi d^2}\left(\sum_{i=1}^K\frac{1}{x_i}\right)^2=\Theta(\delta^{2/\gamma}d^{-2}).
\end{aligned}
\end{align}
The last term of $I_4$ can be written as
\begin{align}
\begin{aligned}
\frac{1}{d^3}\sum^K_{\substack{i,j,k=1\\i\neq j,k}}\frac{1}{(x_j-x_i)(x_k-x_i)}\approx\frac{1}{\delta^2d^{3}}\sum^K_{\substack{i,j,k=1\\i\neq j,k}}\frac{1}{(j^{-\gamma}-i^{-\gamma})(k^{-\gamma}-i^{-\gamma})}\approx\frac{1}{\delta^2d^3}\sum_{i=1}^K\sum_{m,n}\frac{i^{2\gamma+2}}{mn\gamma^2}
\end{aligned}
\end{align}
where we assuming $m,n\ll i$ to obtain the leading term. For a fixed $i$, we sum over $m,n=-(i-1),\cdots,-1,1,\cdots,K-i$, which gives
\begin{align}
\sum_{i=1}^K\sum_{p,q}\frac{i^{2\gamma+2}}{pq}=\sum_{i=1}^Ki^{2\gamma+2}(H_{K-i}-H_{i-1}^2)\approx K^{2\gamma+3}\int_0^1x^{2\gamma}\left(\log\frac{1-x}{x}\right)^2dx,
\end{align}
where we denote $H_i=\sum_{p=1}^i\frac{1}{p}=\log i+\Theta(1)$. Thus we have
\begin{align}
\frac{1}{d^3}\sum^K_{\substack{i,j,k=1\\i\neq j,k}}\frac{1}{(x_j-x_i)(x_k-x_i)}=\Theta(\delta^{-2}d^{-3}K^{2\gamma+3})=\Theta(\delta^{3/\gamma}d^{-3}).
\end{align}
For $d\ll n\ll d^{2\gamma+1}$ and $\gamma>\frac{1}{2}$ we have $\delta^{1/\gamma}d^{-1}\ll1$, and thus 
\begin{align}
\frac{4\pi^2}{3}\int\nu(y)^3dy=1+\Theta(d^{-1}\delta^{1/\gamma}).
\label{eq:integral_Bayes}
\end{align}
Taking it back to \eqref{eq:SE_Bayes}, we have
\begin{align}
\frac{2\alpha\Delta}{\Delta+2(Q^\star -q)}-2\alpha=\Theta\left(d^{-1}\left(\frac{4n}{d\Delta}\right)^{\frac{1}{2\gamma}}\right),
\end{align}
which gives
\begin{align}
\mathsf{R}^{\rm BO}:=Q^\star -q=\Theta\left(\left(\frac{d\Delta}{4n}\right)^{1-\frac{1}{2\gamma}}\right).
\end{align}

\paragraph{Phase III: Not enough data} Phase III is $n\ll d$, and thus $\delta\ll1$. By eq. (\ref{eq:BBP-density}) and the expansion in the end of this section, there are no outliers and the first-order correction reads
\begin{align}
\frac{4\pi^2}{3}\int\nu(x)^3\rd x=1-\frac{1}{d}\sum_{i=1}^{c(d)}(\delta i^{-\gamma})^2=1-\zeta(2\gamma)\hat{q}.
\label{eq:integral_Bayes_small_n}
\end{align}
Taking it back to eq. (\ref{eq:SE_Bayes}), we have
\begin{align}
\frac{2\alpha\Delta}{\Delta+2(Q^\star -q)}-2\alpha=-\frac{4\alpha Q^\star }{\Delta+2(Q^\star -q)},
\end{align}
and thus $\mathsf{R}^{\rm BO}:=Q^\star -q=Q^\star $, where we use $Q^\star \approx\zeta(2\gamma)$.
\paragraph{Perturbative expansion of the bulk}
In the end of this section we discuss how to obtain the correction of the bulk in eq. (\ref{eq:BBP-density}). Consider $H:=Z+\sum_{i=1}^k\lambda_iv_iv_i^T$ with $\{\lambda_i\}_{i=1}^k$ smaller than $1$, where $Z\sim \GOE(d)$ and $\{v_i\}_{i=1}^k$ are uniformly sampled from the unit sphere. Its resolvent can be expanded as
\begin{align}
\begin{aligned}
m_H(z):&=\frac{1}{d}\Tr(z-H)^{-1}\\&\approx m_Z(z)+\frac{1}{d}\sum_{i=1}^k\lambda_iv_i^T(z-Z)^{-2}v_i+\frac{1}{d}\sum_{i,j=1}^k\Tr(z-Z)^{-1}v_iv_i^T(z-Z)^{-1}v_jv_J^T(z-Z)^{-1}.
\end{aligned}
\end{align}
For the first-order correction we have $\frac{1}{d}\sum_{i=1}^k\lambda_iv_i^T(z-Z)^{-2}v_i\approx\frac{\sum_{i=1}^k\lambda_i}{d}m_Z'(z)$. For the second-order correction we have
\begin{align}
\begin{aligned}
&\frac{1}{d}\sum_{i,j=1}^k\Tr(z-Z)^{-1}v_iv_i^T(z-Z)^{-1}v_jv_j^T(z-Z)^{-1}\\&\approx\frac{\Tr(z-Z)^{-1}\Tr(z-Z)^{-2}+\Tr(z-Z)^{-3}}{d^2(d+2)}\sum_{i=1}^k\lambda_i^2\\
&\approx-\frac{1}{d}\sum_{i=1}^k\lambda_i^2m_Z(z)m'_Z(z).
\end{aligned}
\end{align}
This gives a correction on the spectrum $\tilde\nu(x)=\frac{\sum_{i=1}^k\lambda_i}{d}\mu'_{\rm sc}(x)-\frac{\sum_{i=1}^k\lambda_i^2}{d}\text{im}(m_Z(x+i0)m_Z'(x+i0))$. Note that the first term is an odd function and the second term is an even function, and the resolvent of GOE is given by
\begin{align}
m_Z(x+i0)=\frac{x}{2}+i\mu_{\rm sc}(x),
\end{align}
so we have 
\begin{align}
\begin{aligned}
\frac{4\pi^2}{3}\int\nu_H(x)^3dx&\approx\frac{4\pi^2}{3}\int\mu_{\rm sc}(x)^3dx+4\pi^2\int\mu_{\rm sc}(x)^2\tilde\nu(x)dx\\
&=1-\left(\frac{1}{d}\sum_{i=1}^k\lambda_i^2\right)2\pi^2\int\mu_{\rm sc}(x)^2(x\mu_{\rm sc}'(x)+\mu_{\rm sc}(x))dx\\
&=1-\left(\frac{1}{d}\sum_{i=1}^k\lambda_i^2\right)\frac{4\pi^2}{3}\int\mu_{\rm sc}(x)^3dx\\
&=1-\frac{1}{d}\sum_{i=1}^k\lambda_i^2.
\end{aligned}
\label{eq:bulk_correction1}
\end{align}
This correction is the leading term only if $\sum_{i=1}^k\lambda_i^3\ll\sum_{i=1}^k\lambda_i^2$. However, if $\lambda_i=i^{-\gamma}$, all the higher-order terms are of the same order, but their sum converges for $|z|>3$. By using a heuristic analytic continuation we obtain
\begin{align}
\frac{4\pi^2}{3}\int\nu_H(x)^3dx=1-\Theta\left(\frac{1}{d}\sum_{i=1}^k\lambda_i^2\right)
\label{eq:bulk_correction2}
\end{align}
instead.

\subsection{ERM}
\label{app:noisy_ERM} 
In this section we solve eq. (\ref{eq:SE_ERM}) for $\Delta>0$. We recall them here:
\begin{equation} \label{eq:SE_ERM_app}
   \begin{dcases}
&4\alpha\delta-\frac{\delta}{\epsilon}=\partial_1 J(\delta,\lambda\epsilon)\\
&Q^\star +\frac{\Delta}{2}+2\alpha\delta^2-\frac{\delta^2}{\epsilon}=(1-\lambda\epsilon\partial_2) J(\delta,\lambda\epsilon),
\end{dcases}
\end{equation}
The key difficulty will be to evaluate $J(a,b)$.
\begin{equation}\label{eq:Jab}
    J(a,b):=\int_b^{+\infty}\mu_a(x)(x-b)^2\rd x
\end{equation}
For convenience, we can think of this as the second moment of the matrix 
$\hat{\bS} = (\bS^\star+\delta\bZ - \lambda\epsilon \bI_d)_+$, where $\bI_d$ is the identity matrix and $(\cdot)_+$  is the operator that sets all the negative eigenvalues to zero. The empirical spectral density $\mu(x)$ of $\hat{\bS}$ is a simple function of $\mu_\delta(x)$ defined in \eqref{eq:BBP-density}
\begin{equation}
\mu(x)
=
\begin{cases}
\mu_\delta(x+\lambda\epsilon), & x>0,\\
0, & x< 0,
\end{cases}
\end{equation}
with possibly a mass at $x=0$. We decompose this integral $J$ in its two parts (see \eqref{eq:BBP-density})
\begin{align}
    \begin{dcases}
        J_1(\delta,\lambda\epsilon)    &=    \int_{0}^{2\delta-\lambda\epsilon}x^2\,    \mu_{\rm sc}\big((x+\lambda\epsilon)/\delta\big)/\delta\,\rd x =\delta^2 \int_{s}^{2} (x-s)^2\,\mu_{\rm sc}(x) \,{\rm d}x \\
        J_2(\delta,\lambda\epsilon)& =\frac{1}{d}\sum_{i=1}^{K}\left[\sqrt{d}i^{-\gamma}+\frac{\delta^2}{\sqrt{d}i^{-\gamma}}-\lambda\epsilon\right]^2
    \end{dcases}
\end{align}
We notice that $J_1(\delta,\lambda\epsilon)$ is homogeneous of degree $2$, hence
\begin{equation}
    \delta \partial_1 J_1(\delta,\lambda\epsilon) + \lambda\epsilon \partial_2J_1(\delta,\lambda\epsilon) = 2J_1(\delta,\lambda\epsilon)
\end{equation}
This is useful, since we can write the risk by combining the state evolution equations \eqref{eq:SE_ERM_app}
\begin{align} \label{eq:ERM_risk}
    \risk:=2\alpha\delta^2-\frac{\Delta}{2}&=Q^\star-(1-\delta\del_1-\lambda\epsilon\del_2)J(\delta,\lambda\epsilon)\\
    &=Q^\star+J_1(\delta,\lambda\epsilon)-(1-\delta\del_1-\lambda\epsilon\del_2)J_2(\delta,\lambda\epsilon)
\end{align}
For the following, it's useful to have clearly in mind how the spectrum of  $\mu(x)$ looks like. Starting with the bulk, one has a cut of a semicircle with center at $-\lambda \epsilon$, which spans between $x=0$ and $x=2\delta-\lambda\epsilon$. In the following, we will frequently work with the normalized shift $s = \lambda \epsilon / \delta$ and the normalized width of the bulk $t = 2 - s$.\\

We will make use of two expansions of $J_1$, the first in the case in which the bulk is a small cut of the semicircle near the edge, so for $s \approx 2^-$ (or equivalently $t \approx 0^+$) and the second one for the case in which the bulk is almost a full half of the semicircle $s\approx 0^+$ (or equivalently $t \approx 2^+$):
\begin{align}\label{eq:Jab_approx}
    \begin{dcases}
        J_1(\delta,\lambda\epsilon)=\delta^2\int_{2-t}^{2}\frac{\sqrt{4-x^2}}{2\pi}[x-(2-t)]^2\rd x=\delta^2\frac{16t^{7/2}}{105\pi}+\delta^2\frac{16t^{9/2}}{315\pi}+\caO(t^{11/2})\quad \text{for}\quad t\to 0^+ \\
        J_1(\delta,\lambda\epsilon)=\delta^2\int_{s}^2\frac{\sqrt{4-x^2}}{2\pi}(x-s)^2dx=\frac{1}{2}\delta^2-\frac{8s}{3\pi}\delta^2+\caO(s^2) \quad \text{for} \quad s\to 0^+\,,
    \end{dcases}
\end{align}
 To show this, starting from the first equation,
we do the change of variables $y=2-x$:
\begin{align}
    \int_{2-t}^{2}&\frac{\sqrt{4-x^2}}{2\pi}[x-(2-t)]^2\rd x=\int_{0}^{t}\frac{\sqrt{4y-y^2}}{2\pi}(y-t)^2\rd y\\
    \overset{0<y<t\ll 1}{\approx}&\frac{1}{\pi}\int_0^t\sqrt{y}(y-t)^2\rd y +\frac{1}{8\pi}\int_0^ty^{3/2}(y-t)^2\rd y\\
    =&\frac{1}{\pi}\int_0^t(y^{5/2}-2ty^{3/2}+t^2y^{1/2})\rd y+\frac{1}{8\pi}\int_0^t(y^{7/2}-2ty^{5/2}+t^2y^{3/2})\rd y\\
    =&\frac{1}{\pi}\left(\frac{2}{7}t^{7/2}-2\frac{2}{5}t^{7/2}+\frac{2}{3}t^{7/2}\right)+\frac{1}{8\pi}\left(\frac{2}{9}t^{9/2}-2\frac{2}{7}t^{9/2}+\frac{2}{5}t^{9/2}\right)=\frac{16t^{7/2}}{105\pi}+\frac{16t^{9/2}}{315\pi}
\end{align}
For the second one, we instead use $0<s\ll 1$ to get 
\begin{align}
    &\int_{s}^{2}\frac{\sqrt{4-x^2}}{2\pi}(x-s)^2\rd x=\int_{0}^{2}\frac{\sqrt{4-x^2}}{2\pi}(x-s)^2\rd x-\int_{0}^{s}\frac{\sqrt{4-x^2}}{2\pi}(x-s)^2\rd x\\
    &\approx \frac{8}{\pi}\int_0^1\sqrt{1-y^2}(y^2-sy)\rd y-\frac{1}{\pi}\int_0^s(x-s)^2\rd x=\frac{8}{\pi}\left(\frac{\pi}{16}-\frac{s}{3}\right)
\end{align}
As for the spikes, they will be at positions $\{f_\delta(\sqrt{d}i^{-\gamma}) - \lambda\epsilon\}_{i=1,...,K}$, as long as they are positive. In formulas, the number of spikes $K$ is the largest integer such that both these inequalities are true
\begin{equation} \label{eq:K_conditions}
    \sqrt{d}K^{-\gamma}>2\delta\,,\qquad
    f_\delta(\sqrt{d} K^{-\gamma}) > \lambda\epsilon
\end{equation}
then the $K$-th spike is outside of the bulk. We will need to consider three cases: $K=0$, $K=c(d)$ and $1\ll K\ll c(d)$. In the intermediate case $1\ll K\ll c(d)$, $K$ is either limited by the BBP transition in the underregularized phases (first condition of \eqref{eq:K_conditions}), or by the shift $\lambda\epsilon$ in the overregularized one (second condition).\\
For any regularization, one can write for $1\ll K\ll c(d)$
\begin{align} \label{eq:J2_intermediate}
      &J_2(\delta,\lambda\epsilon) =\frac{1}{d}\sum_{i=1}^{K}\left[\sqrt{d}i^{-\gamma}+\frac{\delta^2}{\sqrt{d}i^{-\gamma}}-\lambda\epsilon\right]^2\\
    &=\sum_{i=1}^{K}\left[i^{-2\gamma}+\frac{\delta^4}{d^2i^{-2\gamma}}+\frac{1}{d}\lambda^2\epsilon^2-2\lambda\epsilon\frac{i^{-\gamma}}{\sqrt
    {d}}-2\lambda\epsilon\frac{\delta^2}{d^{3/2}i^{-\gamma}}+2\frac{\delta^2}{d}\right]\\
    &\approx Q^\star-\frac{K^{1-2\gamma}}{1-2\gamma}+\frac{K}{d}(\lambda^2\epsilon^2+2\delta^2)+\frac{\delta^4}{d^2}\frac{K^{1+2\gamma}}{1+2\gamma}-\frac{2\lambda\epsilon\delta^2}{d^{3/2}}\frac{K^{1+\gamma}}{1+\gamma}-\frac{2\lambda\epsilon}{\sqrt{d}}\frac{K^{1-\gamma}}{1-\gamma}-\frac{2\lambda\epsilon}{\sqrt{d}}\zeta(\gamma)\mathbf{1}_{\gamma>1}
\end{align}
where we used that for any $\eta\neq 1$
\begin{align}
\sum_{i=1}^Ki^{-\eta}\approx\frac{K^{1-\eta}}{1-\eta}+\zeta(\eta)\mathbf{1}_{\eta>1}
\label{eq:sum_i}
\end{align}
Some terms in this sum will be negligible if we consider $\lambda\epsilon\ll 2\delta$ or $\lambda\epsilon\gg 2\delta$, but not if the two quantities are of the same order.
\subsubsection{Overregularized phases}
Since the semicircle contribution stems from the label noise and missing samples, we expect it to be shifted away for large regularizations. This can be translated into the condition $\lambda\epsilon>2\delta$. The different phases will then be determined by the value of $K$, the number of (non-zero) spikes that are recovered in the spectrum of the student.\\

We can look at two cases: either $\lambda\epsilon\gg 2\delta$ (far from the boundary) or $0<2-\frac{\lambda\epsilon}{\delta} \ll 1$ (close to the regularization boundary). In both cases, $J=J_2$, while the derivative with respect to $\delta$ in \eqref{eq:SE_ERM_app} can be neglected only deep in the phases.\\
Far from the regularization boundary, we can summarize the conditions on $K,~\delta~\text{and}~\lambda\epsilon$ as: $\sqrt{d}\gg \sqrt{d}K^{-\gamma}\geqslant\lambda\epsilon\gg 2\delta$. We notice that the BBP-shift is negligeable. That is $\forall 1\leqslant i\leqslant K (\leqslant c(d)~)$:
\begin{align}
    \frac{\delta^2}{\sqrt{d}i^{-\gamma}}\leqslant \frac{\delta^2}{\sqrt{d}K^{-\gamma}}\ll \sqrt{d}K^{-\gamma}\leqslant \sqrt{d}i^{-\gamma}\quad \implies\quad  f_\delta (\sqrt{d}i^{-\gamma})=\sqrt{d}i^{-\gamma}+\frac{\delta^2}{\sqrt{d}i^{-\gamma}}\approx \sqrt{d}i^{-\gamma}
\end{align}
Then we can write the truncated moment $J(\delta,\lambda\epsilon)$ in this limit
\begin{align} \label{eq:Jab_overreg}
    J_2(\delta,\lambda\epsilon)& =\frac{1}{d}\sum_{i=1}^{K}\left[\sqrt{d}i^{-\gamma}+\frac{\delta^2}{\sqrt{d}i^{-\gamma}}-\lambda\epsilon\right]^2
    \approx \frac{1}{d}\sum_{i=1}^{K}\left[\sqrt{d}i^{-\gamma}-\lambda\epsilon\right]^2\\
    &=\sum_{i=1}^{K}\left[i^{-2\gamma}-\frac{2\lambda\epsilon i^{-\gamma}}{\sqrt{d}}+\frac{\lambda^2\epsilon^2}{d}\right]=\sum_{i=1}^{K}i^{-2\gamma}+\frac{K}{d}\lambda^2\epsilon^2-\lambda\epsilon\sum_{i=1}^{K}\frac{2i^{-\gamma}}{\sqrt{d}}\\
    &\implies \del_1J_2(\delta,\lambda\epsilon)\approx 0
\end{align}
This implies
\begin{align} \label{eq:SE_quad_overreg}
\begin{dcases}
4\alpha\delta-\frac{\delta}{\epsilon}\approx 0\\
Q^\star +\frac{\Delta}{2}+2\alpha\delta^2-\frac{\delta^2}{\epsilon}= (1-\lambda\epsilon\del_2)J_2(\delta,\lambda\epsilon)
\end{dcases}
\quad \overset{\delta>0}{\implies} \quad
\begin{dcases}
    \epsilon\approx \frac{1}{4\alpha}\\
    \delta\approx\sqrt{\frac{1}{\alpha}}\sqrt{2\risk+\Delta}\\
    \risk=Q^\star-(1-\lambda\epsilon\del_2)J_2(\delta,\lambda\epsilon).
\end{dcases}
\end{align}
The overregularization condition can be written in terms of $\lambda,~d$ and $n$ as 
\begin{equation}
    \lambda\epsilon\gg 2\delta \quad \implies \quad \frac{\lambda}{4\alpha}\gg 2\sqrt{\frac{1}{\alpha}}\sqrt{2\risk+\Delta}\quad
\end{equation}

If instead we have $0<2\delta-\lambda\epsilon \ll 1$, close to the regularization boundary, we need to look at all six terms of $J_2$, and we can plug in $2\delta\approx \lambda\epsilon$ after taking derivatives, so we need to treat this case by case.\\

We can now treat the cases $K=0$, $1\ll K\ll c(d)$ and $K=c(d)$ separately to determine the risk scalings. We expect that $K$ grows with the number of samples.
\paragraph{Phase Ib: Trivial phase}
If $K=0$ then $J(\delta,\lambda\epsilon)=0$ and \eqref{eq:SE_quad_overreg} trivially gives
\begin{equation}
    \begin{dcases}
4\alpha\delta-\frac{\delta}{\epsilon}= 0\\
Q^\star +\frac{\Delta}{2}+2\alpha\delta^2-\frac{\delta^2}{\epsilon}= 0
\end{dcases}
\quad \implies \quad
\begin{dcases}
    \epsilon= \frac{1}{4\alpha}\\
    \delta=\sqrt{\frac{1}{\alpha}}\sqrt{2Q^\star+\Delta}\\
    \risk=Q^\star
\end{dcases}
\end{equation}
and the conditions $K=0$ and $2\delta<\lambda\epsilon$ give
\begin{align}
    &\lambda\epsilon> \sqrt{d}\quad \implies\quad \frac{\lambda d^2}{4n}> \sqrt{d}\quad \implies\quad n< \lambda d^{\frac{3}{2}}\\
    \mathand &2\delta<\lambda\epsilon \quad\implies\quad \frac{\lambda}{4\alpha}> 2\sqrt{\frac{1}{\alpha}}\sqrt{2Q^\star+\Delta}\quad \implies\quad \lambda> 8\sqrt{\alpha}\sqrt{2Q^\star+\Delta}
\end{align}
So phase Ib is 
\begin{equation}
\boxed{\risk=Q^\star\mathif\lambda>8\sqrt{Q^\star+\Delta}\sqrt{\frac{n}{d^2}}\mathand n< \lambda d^{\frac{3}{2}}}.
\end{equation}

\paragraph{Phase II: Under-sampling} For intermediate values of $K$, we can get the following scaling for $K$, by setting the smallest recovered spike equal to the shift. Far from the regularization boundary, when $\lambda\epsilon\gg 2\delta$, we have
\begin{equation} \label{eq:K_cond}
    0=\sqrt{d}K^{-\gamma}+\frac{\delta^2}{\sqrt{d}K^{-\gamma}}-\lambda\epsilon\approx \sqrt{d}K^{-\gamma}-\lambda\epsilon\quad \implies\quad K\approx\left(\frac{\sqrt{d}}{\lambda\epsilon}\right)^{1/\gamma}\approx \left(\frac{4n}{\lambda d^{3/2}}\right)^{1/\gamma}
\end{equation}
where we also used $\epsilon\approx (4\alpha)^{-1}$ from \eqref{eq:SE_quad_overreg}.
We now write the sum  $J(\delta,\lambda\epsilon)$ in this scaling,  continuing from \eqref{eq:Jab_overreg} (and considering $\gamma\neq 1$)
\begin{align}
    J(\delta,\lambda\epsilon)  &\approx Q^\star-\frac{K^{1-2\gamma}}{2\gamma-1}-2K^{-\gamma}\sum_{i=1}^Ki^{-\gamma}+K^{1-2\gamma}\\
    &\approx Q^\star+\left(1-\frac{1}{2\gamma-1}-\frac{2}{1-\gamma}\right)K^{1-2\gamma}-2K^{-\gamma}\zeta(\gamma)\mathbf{1}_{\gamma>1}\\
    &=Q^\star-\frac{2\gamma^2}{(1-\gamma)(2\gamma-1)}\left(\frac{\sqrt{d}}{\lambda\epsilon}\right)^{-2+1/\gamma} -2\frac{\lambda\epsilon}{\sqrt{d}}\zeta(\gamma)\mathbf{1}_{\gamma>1}
\end{align}
where we used again \eqref{eq:sum_i}.

For the derivative, it follows that 
\begin{align}
    \lambda\epsilon\del_2J(\delta,\lambda\epsilon)&=-\frac{2\gamma^2}{(2\gamma-1)(1-\gamma)}\frac{1-2\gamma}{\gamma}\left(\frac{\sqrt{d}}{\lambda\epsilon}\right)^{-2+1/\gamma} -2\frac{\lambda\epsilon}{\sqrt{d}}\zeta(\gamma)\mathbf{1}_{\gamma>1}\\
    &=\frac{2\gamma}{1-\gamma}\left(\frac{\sqrt{d}}{\lambda\epsilon}\right)^{-2+1/\gamma} -2\frac{\lambda\epsilon}{\sqrt{d}}\zeta(\gamma)\mathbf{1}_{\gamma>1}
\end{align}
\noindent
Finally, we get for the risk
\begin{align}
\risk\approx\frac{2\gamma}{2\gamma-1}\left(\frac{4n}{\lambda d^{3/2}}\right)^{-2+1/\gamma} 
  \end{align}
The conditions are
\begin{align}
    \begin{cases}
        1\ll K\ll d\\
        \lambda\epsilon\gg2\delta
    \end{cases}
    \implies
    \begin{cases}
        \lambda d^{3/2}\ll n\ll \lambda d^{3/2}c(d)^\gamma\\
        \lambda\gg\sqrt{\frac{n}{d^2}}
    \end{cases}.
\end{align}
Therefore, we can write 
\begin{equation}
    \boxed{R=\frac{2\gamma}{2\gamma-1}\left(\frac{n}{\lambda d^{3/2}}\right)^{-2+1/\gamma}\mathif \lambda\gg\sqrt{\frac{n}{d^2}}\mathand \lambda d^{3/2}\ll n\ll \lambda d^{3/2+\gamma}}.
\end{equation}

\noindent
If instead $0<\frac{\lambda\epsilon}{\delta}-2 \ll 1$, we get the following scaling for $K=K(\delta,\lambda\epsilon)$
\begin{equation} \label{eq:K_cond_bdry}
    0=\sqrt{d}K^{-\gamma}+\frac{\delta^2}{\sqrt{d}K^{-\gamma}}-\lambda\epsilon \quad \implies\quad K\approx\left(\frac{2\sqrt{d}}{\lambda\epsilon}\right)^{1/\gamma}.
\end{equation}

We compute the second moment from \eqref{eq:J2_intermediate}:
\begin{align}
     J_2(\delta,\lambda\epsilon)&\approx Q^\star+\frac{K^{1-2\gamma}}{1-2\gamma}+\frac{K}{d}(\lambda^2\epsilon^2+2\delta^2)+\frac{\delta^4}{d^2}\frac{K^{1+2\gamma}}{1+2\gamma}-\frac{2\lambda\epsilon\delta^2}{d^{3/2}}\frac{K^{1+\gamma}}{1+\gamma}-\frac{2\lambda\epsilon}{\sqrt{d}}\frac{K^{1-\gamma}}{1-\gamma}-\frac{2\lambda\epsilon}{\sqrt{d}}\zeta(\gamma)\mathbf{1}_{\gamma>1}\\
     &\approx Q^\star+\left(6+\frac{1}{1-2\gamma}-\frac{4}{1-\gamma}-\frac{4}{1+\gamma}+\frac{1}{1+2\gamma}\right)K^{1-2\gamma}-2\frac{\lambda\epsilon}{\sqrt{d}}\zeta(\gamma)\mathbf{1}_{\gamma>1}\\
    &=Q^\star+\frac{24\gamma^{4}}{\left( 1-\gamma^2\right) \left(1-4 \gamma^2\right)}K^{1-2\gamma} -2\frac{\lambda\epsilon}{\sqrt{d}}\zeta(\gamma)\mathbf{1}_{\gamma>1} 
\end{align}
    The derivatives thus are (evaluating at $2\delta\approx \lambda\epsilon\approx 2\sqrt{d}K^{-\gamma}$ after deriving)
\begin{align}
    &\lambda\epsilon\del_2J_2(\delta,\lambda\epsilon)\approx \lambda\epsilon\del_{\lambda\epsilon} K(\delta,\lambda\epsilon)\left[ K^{-2\gamma}+\frac{1}{d}(\lambda^2\epsilon^2+2\delta^2)+\frac{\delta^4}{d^2}K^{2\gamma}-2\frac{\lambda\epsilon}{d}\frac{\delta^2}{\sqrt{d}K^{-\gamma}}-2\frac{\lambda\epsilon}{d}\sqrt{d}K^{-\gamma}\right]\\
    &\qquad\qquad+2\lambda^2\epsilon^2\frac{K}{d}-\frac{2\lambda\epsilon\delta^2}{d^{3/2}}\frac{K^{1+\gamma}}{1+\gamma}-\frac{2\lambda\epsilon}{\sqrt{d}}\frac{K^{1-\gamma}}{1-\gamma}-\frac{2\lambda\epsilon}{\sqrt{d}}\zeta(\gamma)\mathbf{1}_{\gamma>1}\\
    &=\lambda\epsilon\del_{\lambda\epsilon} K(\delta,\lambda\epsilon)\underbrace{\left[K^{-2\gamma}+\frac{1}{d}(2\delta^2-\lambda^2\epsilon^2)+\frac{\delta^4}{d^2}K^{2\gamma}\right]}_{=0~\mathrm{from~\eqref{eq:K_cond}^2}}+8K^{1-2\gamma}-\frac{2\lambda\epsilon\delta^2}{d^{3/2}}\frac{K^{1+\gamma}}{1+\gamma}-\frac{2\lambda\epsilon}{\sqrt{d}}\frac{K^{1-\gamma}}{1-\gamma}-\frac{2\lambda\epsilon}{\sqrt{d}}\zeta(\gamma)\mathbf{1}_{\gamma>1}\\
    &=8K^{1-2\gamma}-\frac{2\lambda\epsilon\delta^2}{d^{3/2}}\frac{K^{1+\gamma}}{1+\gamma}-\frac{2\lambda\epsilon}{\sqrt{d}}\frac{K^{1-\gamma}}{1-\gamma}-\frac{2\lambda\epsilon}{\sqrt{d}}\zeta(\gamma)\mathbf{1}_{\gamma>1}\\
    & \implies(1-\lambda\epsilon\del_2)J_2(\delta,\lambda\epsilon)\approx Q^\star+\left(\frac{1}{1-2\gamma}+\frac{1}{1+2\gamma}-2\right)K^{1-2\gamma} =Q^\star-\frac{8\gamma^2}{4\gamma^2-1}K^{1-2\gamma} 
\end{align}
Similarly, the derivative term $\del_\delta K$ cancels naturally and
\begin{align}
    \delta\del_1J_2(\delta,\lambda\epsilon)\approx 4\frac{K}{d}\delta^2+4\frac{\delta^4}{d^2}\frac{K^{1+2\gamma}}{1+2\gamma}-\frac{4\lambda\epsilon\delta^2}{d^{3/2}}\frac{K^{1+\gamma}}{1+\gamma}=\left(4+\frac{4}{1+2\gamma}-\frac{8}{1+\gamma}\right)K^{1-2\gamma}=\underbrace{\frac{8\gamma^2}{(1+2\gamma)(1+\gamma)}}_{A(\gamma)}K^{1-2\gamma}
\end{align}
In this case, the risk is
\begin{align} \label{eq:risk_II}
        \risk=Q^\star-(1-\delta\del_1-\lambda\epsilon\del_2)J_2(\delta,\lambda\epsilon)\approx\frac{24\gamma^3}{\left(4 \gamma^2-1\right)(1+\gamma)}K^{1-2\gamma}
\end{align}
while the first equation of \eqref{eq:SE_ERM_app} becomes 
\begin{align}
        4\alpha d-\frac{d}{\epsilon}\approx A(\gamma)K
    \quad\implies\quad
        \epsilon\approx\frac{1}{4\alpha-A(\gamma)\frac{K}{d}}\overset{K\ll d}{\approx}\frac{1}{4\alpha}
\end{align}
This gives $K^\gamma\approx\frac{8n}{\lambda d^{3/2}}$. Since $\delta=\frac{1}{\sqrt{\alpha}}\sqrt{2\risk+\Delta}\approx\sqrt{\frac{\Delta d^2}{n}}$, the conditions on the regularization and the number of samples are
\begin{align}
    \begin{cases}
        1\ll K\ll d\\
        0<\frac{\lambda\epsilon}{\delta}-2\ll1
    \end{cases}
    \implies
    \begin{cases}
        0<\frac{\lambda}{4\sqrt{\Delta}}\sqrt{\frac{d^2}{n}}-1\ll1\\
        d\ll n\ll d^{2\gamma+1}
    \end{cases}.
\end{align}
Therefore, we obtain the same error scaling close to the boundary
\begin{equation}
\boxed{\risk=\frac{24\gamma^3}{(4\gamma^2-1)(1+\gamma)}\left(\frac{8n}{\lambda d^{3/2}}\right)^{\frac{1-2\gamma}{\gamma}}\mathif d\ll n\ll d^{2\gamma+1}\mathand 0<\frac{\lambda}{4\sqrt{\Delta}}\sqrt{\frac{d^2}{n}}-1\ll1}.
\end{equation}

\paragraph{Phase III: Large sample phase} 
We write the truncated moment $J(\delta,\lambda\epsilon)=J_2(\delta,\lambda\epsilon)$ in the case where all spikes are recovered, i.e. $K=c(d)$, continuing from \eqref{eq:Jab_overreg} for $\lambda\epsilon\gg 2\delta$
\begin{align}
    J_2(\delta,\lambda\epsilon)& \approx\sum_{i=1}^{c(d)}\left[i^{-2\gamma}-\frac{2\lambda\epsilon i^{-\gamma}}{\sqrt{d}}+\frac{\lambda^2\epsilon^2}{d}\right]=Q^\star+\frac{c(d)}{d}\lambda^2\epsilon^2-\lambda\epsilon\sum_{i=1}^{c(d)}\frac{2i^{-\gamma}}{\sqrt{d}}\\
    \implies \quad & 
    \lambda\epsilon\del_2J(\delta,\lambda\epsilon)\approx 2\frac{c(d)}{d}\lambda^2\epsilon^2-\lambda\epsilon\sum_{i=1}^{c(d)}\frac{2i^{-\gamma}}{\sqrt{d}}
\end{align}
Plugging these into the expression for the risk we get with $\epsilon\approx (4\alpha)^{-1}$ (from \eqref{eq:SE_quad_overreg})
\begin{align}
    \risk\approx \frac{c(d)}{d}\frac{\lambda^2d^4}{16n^2}
\end{align}
We take the limit $c(d)\to d$. The condition $K=c(d)$ requires $\lambda\epsilon\ll d^{-\gamma+\frac{1}{2}}$. Using $\epsilon\approx (4\alpha)^{-1}$ and $\delta\approx\sqrt{\frac{\Delta}{4\alpha}}$ from \eqref{eq:SE_quad_overreg}, the conditions on the regularization and the number of samples are
\begin{align}
    \begin{cases}
        \lambda\epsilon\ll d^{-\gamma+\frac{1}{2}}\\
        \lambda\epsilon\gg2\delta
    \end{cases}
    \implies
    \begin{cases}
        \lambda\gg\sqrt{\frac{n}{d^2}}\\
         n\gg \lambda d^{3/2+\gamma}
    \end{cases}.
\end{align}
Therefore, we get phase III:
\begin{equation}
\boxed{R= \frac{\lambda^2d^4}{16n^2}\mathif \lambda\gg\sqrt{\frac{n}{d^2}}\mathand n\gg \lambda d^{3/2+\gamma}}.
\end{equation}

Consider now the boundary, where $0<\frac{\lambda\epsilon}{\delta}-2\ll 1$. We compute the second moment from \eqref{eq:J2_intermediate}:
\begin{align}
     &J_2(\delta,\lambda\epsilon)\approx Q^\star+\frac{c(d)}{d}(\lambda^2\epsilon^2+2\delta^2)+\frac{\delta^4}{d^2}\frac{c(d)^{1+2\gamma}}{1+2\gamma}-\frac{2\lambda\epsilon\delta^2}{d^{3/2}}\frac{c(d)^{1+\gamma}}{1+\gamma}-\frac{2\lambda\epsilon}{\sqrt{d}}\zeta(\gamma)\mathbf{1}_{\gamma>1}\\
     &\implies\begin{dcases}
         \delta\del_1J_2(\delta,\lambda\epsilon)\approx4\frac{c(d)}{d}\delta^2+4\frac{\delta^4}{d^2}\frac{c(d)^{1+2\gamma}}{1+2\gamma}-\frac{4\lambda\epsilon\delta^2}{d^{3/2}}\frac{c(d)^{1+\gamma}}{1+\gamma}\\
         \lambda\epsilon \del_2J_2(\delta,\lambda\epsilon)\approx 2\frac{c(d)}{d}\lambda^2\epsilon^2-\frac{2\lambda\epsilon\delta^2}{d^{3/2}}\frac{c(d)^{1+\gamma}}{1+\gamma}-\frac{2\lambda\epsilon}{\sqrt{d}}\zeta(\gamma)\mathbf{1}_{\gamma>1}
     \end{dcases}
 \end{align}
 If we combine these expression into the risk, evaluating at $2\delta\approx \lambda\epsilon$, we get
 \begin{align}
     \risk= Q^\star-(1-\delta\del_1-\lambda\epsilon\del_2)J_2(\delta,\lambda\epsilon)\approx \frac{6c(d)}{4d}\lambda^2\epsilon^2-3\frac{\lambda^4\epsilon^4}{16d^2}\frac{c(d)^{1+2\gamma}}{1+2\gamma}+\frac{\lambda^3\epsilon^3}{d^{3/2}}\frac{c(d)^{1+\gamma}}{1+\gamma}
 \end{align}
 We again get a scaling for $\epsilon$ from the first equation of \eqref{eq:SE_ERM_app}, using the fact that all spikes are recovered, i.e. $2\delta\ll \sqrt{d}c(d)^{-\gamma}\ll \sqrt{d}$, and $\alpha\gg 1$: 
\begin{align}
        4\alpha -\frac{1}{\epsilon}\approx \underbrace{4\frac{c(d)}{d}}_{\ll \alpha}+\underbrace{4\frac{\delta^2}{d^2}\frac{c(d)^{1+2\gamma}}{1+2\gamma}}_{\ll 1}-\underbrace{\frac{8\delta}{d^{3/2}}\frac{c(d)^{1+\gamma}}{1+\gamma}}_{\ll 1}\approx 0
    \quad\implies\quad
        \epsilon\approx\frac{1}{4\alpha}
\end{align} The risk becomes at leading order
 \begin{align}
     \risk\approx \frac{6\lambda^2}{16\alpha^2}\frac{c(d)}{d}+\frac{\lambda^3}{64\alpha^3d^{3/2}}\frac{c(d)^{1+\gamma}}{1+\gamma}-3\frac{\lambda^4}{16^2\alpha^4d^2}\frac{c(d)^{1+2\gamma}}{1+2\gamma}\approx\frac{3\lambda^2}{8\alpha^2}\frac{c(d)}{d}
 \end{align}
We take the limit $c(d)\to d$. Using $\epsilon\approx (4\alpha)^{-1}$ and $\delta\approx\sqrt{\frac{\Delta}{4\alpha}}$ from \eqref{eq:SE_quad_overreg}, the conditions on the regularization and the number of samples are
\begin{align}
    \begin{cases}
        \lambda\epsilon\ll d^{-\gamma+\frac{1}{2}}\\
        0<\frac{\lambda\epsilon}{\delta}-2\ll1
    \end{cases}
    \implies
    \begin{cases}
        0<\frac{\lambda}{4\sqrt{\Delta}}\sqrt{\frac{d^2}{n}}-1\ll1\\
         n\gg \lambda d^{3/2+\gamma}
    \end{cases}.
\end{align}
We get the same risk scaling close to the boundary:
\begin{equation}
    \boxed{R=\frac{3\lambda^2d^4}{8n^2}\mathif 0<\frac{\lambda}{4\sqrt{\Delta}}\sqrt{\frac{d^2}{n}}-1\ll1\mathand n\gg \lambda d^{3/2+\gamma}}
\end{equation}

\subsubsection{Underregularized phases}
We now turn to the case where $2\delta>\lambda\epsilon$. The resulting condition on the regularization will need to be checked in every underregularized phase for consistency with the lower boundary of the overregularized phases.\\

We are now ready to treat the different underregularized phases. Again, we suppose that for growing number of samples, $K$ grows.
\paragraph{Phase Ia: Trivial phase} This phase is defined by $K=0$, meaning $J_2 \equiv 0$ and $2\delta>\lambda\epsilon>\sqrt{d}$. Starting with $2\delta\approx\lambda\epsilon\iff 0<t\ll 1$ , we use the approximation \eqref{eq:Jab_approx} and $s=2-t\approx 2$ to get 
\begin{align}
    J_1(\delta,\lambda\epsilon)\approx \delta^2\frac{16t^{7/2}}{105\pi}\quad \implies \quad       \partial_1 J(\delta,\lambda\epsilon)
    \approx
    2\delta\left(t+\frac{7}{4}s\right)
    \frac{16t^{5/2}}{105\pi}
    \approx
    \delta\frac{16t^{5/2}}{15\pi}.
\end{align}
The state evolution equations then give
\begin{align} \label{eq:SE_ERM_smallt}
      \begin{dcases}
    \risk\approx Q^\star+\delta^2\frac{16t^{7/2}}{105\pi}\\
    \delta^2\approx\frac{Q^\star+\sfrac{\Delta}{2}}{2\alpha-\frac{16t^{7/2}}{105\pi}}\\
    \epsilon\approx\left[4\alpha-  \frac{16t^{5/2}}{15\pi}\right]^{-1}
    \end{dcases}
    \quad\implies\quad
        \begin{dcases}
    \risk\approx Q^\star+\frac{Q^\star+\sfrac{\Delta}{2}}{2\alpha}\frac{16t^{7/2}}{105\pi}\\
    \delta^2\approx\frac{Q^\star+\sfrac{\Delta}{2}}{2\alpha}\\
    \epsilon\approx\left[4\alpha-  \frac{16t^{5/2}}{15\pi}\right]^{-1}
    \end{dcases}
\end{align}
Where we used that $\epsilon>0~\implies~4\alpha>\frac{16t^{5/2}}{15\pi}~\implies~\alpha\gg t^{7/2}$. The condition $2\delta>\sqrt{d}$ then implies $n<4d(2Q^\star+\Delta)$. 
At this point there are two possible choices, either $\alpha \gg t^{5/2}$ or $\alpha =\Theta( t^{5/2})$. 

In the first case we get $\epsilon^{-1} \approx 4\alpha$ under the condition that $t^{5/2}\ll\alpha$. We then assume that $0<\frac{2\delta}{\lambda\epsilon}-1\ll1$ and use $t$ as a parameter to indicate how close we are to the boundary. Using the definition of $t$, we have $\lambda = (2-t)\delta/\epsilon \approx \lambda_c - t \lambda_c/2$, where we define
\begin{equation}
    \lambda_c:=4\sqrt{\alpha \left(2Q^\star + \Delta\right)}.
\end{equation}
Then to leading order close to the boundary, 
\begin{equation} 
t \approx 2\left(1-\frac{\lambda}{\lambda_c}\right)
\end{equation}
with $0<t\ll \alpha^{2/5}$. The condition $t>0$ corresponds to approaching the boundary from inside the phase, namely $\lambda<\lambda_c$. Then the error is
\begin{equation}
\begin{aligned}
\risk&\approx Q^\star+
    \delta^2\frac{16t^{7/2}}{105\pi}\approx Q^\star+\frac{2^{11/2}(2Q^\star+\Delta) d^2}{105\pi n}\left(1-\frac{\lambda}{\lambda_c}\right)^{7/2}.
    \end{aligned}
\end{equation}
The conditions for the boundary are
\begin{align}
    \begin{cases}
        2\delta>\sqrt{d}\\
        0<\frac{2\delta}{\lambda\epsilon}-1\ll1\\
        t^{5/2}\ll\alpha
    \end{cases}
    \implies
    \begin{cases}
        n<(2Q^\star+\Delta)d\\
        0<1-\frac{\lambda}{\lambda_c}\ll\left(\frac{n}{d^2}\right)^{2/5}.
    \end{cases}.
\end{align}
This gives the boundary error rate
\begin{equation}
\boxed{R = Q^\star+\frac{2^{11/2}(2Q^\star+\Delta) d^2}{105\pi n}\left(1-\frac{\lambda}{\lambda_c}\right)^{7/2}\mathif 0<1-\frac{\lambda}{\lambda_c}\ll\left(\frac{n}{d^2}\right)^{2/5}\mathand n<(2Q^\star+\Delta)d}
\end{equation}

Now, for the second case we immediately get $t\approx\left(\frac{15\pi}{4\alpha}\right)^{2/5}$, which will also imply that $\epsilon^{-1}\ll \alpha $. Looking at the left hand side of the first equation of \eqref{eq:SE_ERM}, one has the simplification
\begin{equation}
    4\alpha\delta-\frac{\delta}{\epsilon} \approx 4\alpha\delta
\end{equation}
 In this case, the regularization will be of an order fixed by $\lambda \approx 2\delta/\epsilon \ll \sqrt{\alpha}$. This choice thus corresponds to being deep in the phase, away from the boundary $\lambda =\lambda_c$. The excess risk is given by
 \begin{equation}
\risk\approx Q^\star+\delta^2\frac{16t^{7/2}}{105\pi}\approx Q^\star+\frac{\Delta+2Q^\star}{7}\left(\frac{15\pi}{4}\right)^{2/5}\left(\frac{n}{d^2}\right)^{2/5}.
 \end{equation}
The conditions for the boundary are
\begin{align}
    \begin{cases}
        2\delta>\sqrt{d}\\
        \epsilon^{-1}\ll\alpha
    \end{cases}
    \implies
    \begin{cases}
        n<(2Q^\star+\Delta)d\\
        \lambda\ll\sqrt{\frac{n}{d^2}}.
    \end{cases}.
\end{align}
Summarizing, in Phase Ia we have
\begin{equation}
    \boxed{R = Q^\star+\frac{\Delta+2Q^\star}{7}\left(\frac{15\pi}{4}\right)^{2/5}\left(\frac{n}{d^2}\right)^{2/5}\mathif \lambda\ll\sqrt{\frac{n}{d^2}}\mathand n<(2Q^\star+\Delta)d}
\end{equation}

\paragraph{Phases IV and V: Benign and harmful overfitting}
When $1\ll K\ll c(d)$, the contribution from $J_2$ to the risk is nonzero. We can write the condition as $\sqrt{d}c(d)^{-\gamma}\ll 2\delta\ll\sqrt{d}$.\\
Since $K$ is limited by the BBP transition in the underregularized case (first condition of \eqref{eq:K_conditions}), we have $\sqrt{d}K^{-\gamma}\approx \delta\iff K^\gamma\approx\frac{\sqrt{d}}{\delta}$
so that we can write
\begin{align}
    &J_2(\delta,\lambda\epsilon) =\frac{1}{d}\sum_{i=1}^{K}\left[\sqrt{d}i^{-\gamma}+\frac{\delta^2}{\sqrt{d}i^{-\gamma}}-\lambda\epsilon\right]^2\\
    &=\sum_{i=1}^{K}\left[i^{-2\gamma}+\frac{\delta^4}{d^2i^{-2\gamma}}+\frac{1}{d}\lambda^2\epsilon^2-2\lambda\epsilon\frac{i^{-\gamma}}{\sqrt
    {d}}-2\lambda\epsilon\frac{\delta^2}{d^{3/2}i^{-\gamma}}+2\frac{\delta^2}{d}\right]\\
    &\approx Q^\star+\frac{K^{1-2\gamma}}{1-2\gamma}+\frac{K}{d}(\lambda^2\epsilon^2+2\delta^2)+\frac{\delta^4}{d^2}\sum_{i=1}^{K}i^{2\gamma}-\frac{2\lambda\epsilon\delta^2}{d^{3/2}}\sum_{i=1}^{K}i^{\gamma}-\frac{2\lambda\epsilon}{\sqrt{d}}\sum_{i=1}^{K}i^{-\gamma}\\
    &\approx Q^\star+\frac{K^{1-2\gamma}}{1-2\gamma}+K^{1-2\gamma}\left(2+\frac{\lambda^2\epsilon^2}{\delta^2}\right)+\frac{K^{1-2\gamma}}{1+2\gamma}-2\frac{\lambda\epsilon}{\delta}\frac{K^{1-2\gamma}}{1+\gamma}-2\frac{\lambda\epsilon}{\delta}\frac{K^{1-2\gamma}}{1-\gamma}-\frac{2\lambda\epsilon}{\sqrt{d}}\zeta(\gamma)\mathbf{1}_{\gamma>1}\\
    &\approx Q^\star+\left(\frac{\delta}{\sqrt{d}}\right)^{2-1/\gamma}\left[\frac{-1}{2\gamma-1}+2+\frac{\lambda^2\epsilon^2}{\delta^2}+\frac{1}{2\gamma+1}-\frac{\lambda\epsilon}{\delta}\frac{2}{1+\gamma}-\frac{\lambda\epsilon}{\delta}\frac{2}{1-\gamma}\right]-\frac{2\lambda\epsilon}{\sqrt{d}}\zeta(\gamma)\mathbf{1}_{\gamma>1}\\
    &\approx Q^\star+\left(\frac{\delta}{\sqrt{d}}\right)^{2-1/\gamma}\left[2-\frac{2}{4\gamma^2-1}+\frac{\lambda^2\epsilon^2}{\delta^2}-\frac{\lambda\epsilon}{\delta}\frac{4}{1-\gamma^2}\right]-\frac{2\lambda\epsilon}{\sqrt{d}}\zeta(\gamma)\mathbf{1}_{\gamma>1}
\end{align}
The derivatives are 
\begin{align}
        \delta\del_1J_2(\delta,\lambda\epsilon)&\approx (2-1/\gamma)\left(\frac{\delta}{\sqrt{d}}\right)^{2-1/\gamma}\left[2-\frac{2}{4\gamma^2-1}+\frac{\lambda^2\epsilon^2}{\delta^2}-\frac{\lambda\epsilon}{\delta}\frac{4}{1-\gamma^2}\right]\\
        &\qquad\qquad +\delta\left(\frac{\delta}{\sqrt{d}}\right)^{2-1/\gamma}\left[\frac{-2\lambda^2\epsilon^2}{\delta^3}+\frac{\lambda\epsilon}{\delta^2}\frac{4}{1-\gamma^2}\right]\\
        &\approx \left(\frac{\delta}{\sqrt{d}}\right)^{2-1/\gamma}\left[4-\frac{2}{\gamma(2\gamma+1)}-\frac{2}{\gamma}+\frac{\lambda\epsilon}{\delta}\frac{4}{\gamma(1+\gamma)}-\frac{1}{\gamma}\frac{\lambda^2\epsilon^2}{\delta^2}\right]\\
        \lambda\epsilon\del_2J_2(\delta,\lambda\epsilon)&\approx \left(\frac{\delta}{\sqrt{d}}\right)^{2-1/\gamma}\left[2\frac{\lambda^2\epsilon^2}{\delta^2}-\frac{\lambda\epsilon}{\delta}\frac{4}{1-\gamma^2}\right]-\frac{2\lambda\epsilon}{\sqrt{d}}\zeta(\gamma)\mathbf{1}_{\gamma>1}
\end{align}
Now, the contributions from $J_1$ is exactly the same as in the previous phase, written explicitly as
\begin{align}
    \begin{dcases}
    J_1(\delta,\lambda\epsilon)\approx \delta^2\frac{16t^{7/2}}{105\pi}+\delta^2\frac{16t^{9/2}}{315\pi}+\mathcal{O}(t^{11/2})\\
        \delta\del_1J_1(\delta,\lambda\epsilon)\approx   2\delta^2\frac{16t^{7/2}}{105\pi}+\delta^2\frac{7}{2}\frac{\lambda\epsilon}{\delta}
    \frac{16t^{5/2}}{105\pi}+2\delta^2\frac{8t^{7/2}}{35\pi}+\mathcal{O}(t^{9/2})\\\qquad\qquad\quad\overset{s\approx 2}{\approx}\delta^2\frac{80t^{7/2}}{105\pi}+\delta^2\frac{16t^{5/2}}{15\pi}+\mathcal{O}(t^{9/2})\\
        \lambda\epsilon\del_2J_1(\delta,\lambda\epsilon)\approx -\lambda\epsilon\delta\frac{8t^{5/2}}{15\pi}-\lambda\epsilon\delta\frac{8t^{7/2}}{35\pi}+\mathcal{O}(t^{9/2})\approx -\delta^2\frac{16t^{5/2}}{15\pi}-\delta^2\frac{16t^{7/2}}{35\pi}+\mathcal{O}(t^{9/2})
    \end{dcases}
\end{align}
where we used $s=\frac{\lambda\epsilon}{\delta}=2-t\approx 2$.
such that we can rewrite the state evolution equations \eqref{eq:SE_ERM} and the risk \eqref{eq:ERM_risk} as 
\begin{align} \label{app:eq:5special}
&
    \begin{dcases}
        4\alpha\delta^2-\frac{\delta^2}{\epsilon}\approx\left(\frac{\delta}{\sqrt{d}}\right)^{2-1/\gamma}\left[4-\frac{2}{\gamma(2\gamma+1)}+\frac{8}{\gamma(1+\gamma)}-\frac{6}{\gamma}\right]+\delta^2\frac{80t^{7/2}}{105\pi}+\delta^2\frac{16t^{5/2}}{15\pi}+\mathcal{O}(t^{9/2})\\
        Q^\star+\frac{\Delta}{2}+2\alpha\delta^2-\frac{\delta^2}{\epsilon}\approx Q^\star-\left[2+\frac{2}{4\gamma^2-1}\right]\left(\frac{\delta}{\sqrt{d}}\right)^{2-1/\gamma}+\delta^2\frac{64t^{7/2}}{105\pi}+    \delta^2\frac{16t^{5/2}}{15\pi}+\mathcal{O}(t^{9/2})
    \end{dcases}\\
        &\implies-\risk:=-2\alpha\delta^2+\frac{\Delta}{2}\approx -\delta^2\frac{16t^{7/2}}{105\pi}-\left(\frac{\delta}{\sqrt{d}}\right)^{2-1/\gamma}\left[6-\frac{2(\gamma-1)}{\gamma(4\gamma^2-1)}-\frac{6}{\gamma}+\frac{8}{\gamma(1+\gamma)}\right]
\end{align}
using again that $\lambda\epsilon\approx 2\delta$.
Notice that the risk has to be positive, so we have $4\alpha \delta^2 \geq \Delta$.

The interplay between the two terms gives us two different regimes: an overfitting one with large error when the noise piece dominates $4\alpha \delta^2 \ll \Delta$ and another where the error is almost zero $4\alpha \delta^2\lessapprox \Delta$. We will only be concerned with the second case, where we immediately have $\delta \approx \sqrt{\Delta /4\alpha}$, and we discuss how the first one is outside of the scope of this work at the end of the section.

It is convenient to start by looking at points very far away from the boundaries of the phases in $n$. Phase I was concerning the region was in the regime $n \ll d$, we will instead look at $n \gg d$. For this case we have
\begin{equation}
    \left(\frac{\delta}{\sqrt{d}}\right)^{2-1/\gamma} \asymp\left(\frac{d}{n}\right)^{1-1/2\gamma}
\end{equation}
The scaling of $t$ is found by looking at the first equation in \eqref{app:eq:5special}. The term $4\alpha\delta$ is of order one, so we can discard all the lower orders. The first term on the right hand side is subleading for $n\gg d$, while for the second on the left we have $\delta^2/\epsilon \approx \lambda \delta/ 2 \approx \lambda \sqrt{\Delta / 16\alpha}$. As long as $\lambda\ll \sqrt{\alpha}$, this piece is negligible, and this is the condition for being away from the boundary in $\lambda$. Under this extra assumption one obtains as in phase Ia that $t\asymp \alpha^{2/5}$, which is going to zero if $n\ll d^2$. More precisely we have
\begin{equation}
    4\alpha\approx\frac{16t^{5/2}}{15\pi}
\end{equation}
Summing up, we have for the risk
\begin{equation}
    \risk
    \approx
    \frac{\Delta}{7}
    \left(\frac{15\pi}{4}\right)^{2/5}
    \left(\frac{n}{d^2}\right)^{2/5}
    +
    \left(\frac{\Delta d}{4n}\right)^{1-1/2\gamma}
    \frac{24\gamma^3}{(4\gamma^2-1)(\gamma+1)}
\end{equation}
in the regime $d \ll n \ll d^2$ and $\lambda \ll \sqrt{\alpha}$. Depending on the value of $n$, either the first one or the second one dominates. In particular the first one dominates for $n^{\rm cross}$ that satisfies the relation 
\begin{equation}
    \left(\frac{n^{\rm cross}}{d^2}\right)^{2/5} \gg \left(\frac{d}{n^{\rm cross}}\right)^{1-1/2\gamma} \implies n^{\rm cross} \gg d^{\frac{18\gamma - 5}{14\gamma -5}}
\end{equation}
where the exponent is in the range $(9/7, 2)$. We thus have phase IV
\begin{equation}
    \boxed{\risk = \frac{24\gamma^3}{(4\gamma^2-1)(\gamma+1)}\left(\frac{\Delta d}{4n}\right)^{1-1/2\gamma}
    \mathif \lambda\ll\sqrt{\frac{n}{d^2}}\mathand d\ll n\ll d^{\frac{18\gamma - 5}{14\gamma -5}}}
\end{equation}
and phase V
\begin{equation}
    \boxed{\risk = \frac{\Delta}{7}
    \left(\frac{15\pi}{4}\right)^{2/5}
    \left(\frac{n}{d^2}\right)^{2/5}\mathif \lambda\ll\sqrt{\frac{n}{d^2}}\mathand d^{\frac{18\gamma - 5}{14\gamma -5}} \ll n\ll d^2}
\end{equation}
Now it's time to check the boundaries. The easiest one to check is when $\lambda\asymp \sqrt{\alpha}$. 
The starting point is assuming that $0<  \frac{2\delta}{\lambda\epsilon}-1\ll 1$ and using $t$ as a parameter to indicate how close we are to the boundary. Looking back at the first equation in \eqref{app:eq:5special}, at the zeroth order in $t$ one has $\epsilon^{-1}\approx 4\alpha$ under the condition that $t^{5/2}\ll\alpha$, from which one derives $\lambda = (2-t)\delta/\epsilon = (2 - t) \sqrt{4\Delta\alpha}$.
Calling 
$\lambda_c = 4\sqrt{\Delta\alpha}$, then to leading order close to the boundary, 
\begin{equation} 
t \approx 2\left(1-\frac{\lambda}{\lambda_c}\right)
\end{equation}
with $0<t\ll \alpha^{2/5}$. The condition $t>0$ corresponds to approaching the boundary from inside the phase, namely $\lambda<\lambda_c$. Then the error is
\begin{equation}
\begin{aligned}
\risk&\approx
    \delta^2\frac{16t^{7/2}}{105\pi}+
    \left(\frac{\Delta d}{4n}\right)^{1-1/2\gamma}
    \frac{24\gamma^3}{(4\gamma^2-1)(\gamma+1)}\\
    &\approx\frac{2^{11/2}\Delta d^2}{105\pi n}\left(1-\frac{\lambda}{\lambda_c}\right)^{7/2}+
    \left(\frac{\Delta d}{4n}\right)^{1-1/2\gamma}
    \frac{24\gamma^3}{(4\gamma^2-1)(\gamma+1)},
\end{aligned}
\end{equation}
which gives the boundary error rate
\begin{equation}
    \boxed{\begin{aligned}
    R = \frac{2^{11/2}\Delta d^2}{105\pi n}\left(1-\frac{\lambda}{\lambda_c}\right)^{7/2}+
    \left(\frac{\Delta d}{4n}\right)^{1-1/2\gamma}
    \frac{24\gamma^3}{(4\gamma^2-1)(\gamma+1)}\\\qquad\quad\mathif 0<1-\frac{\lambda}{\lambda_c}\ll\left(\frac{n}{d^2}\right)^{2/5}\mathand d\ll n\ll d^2
    \end{aligned}}.
\end{equation}
The last thing to do is check the boundaries in $n$.
Starting with $n = d^{1 + \varepsilon}$ for very small $\varepsilon$, we can use the result for the risk from the previous phase and obtain
\begin{equation}
    \risk
    \asymp
    d^{-\varepsilon(1-1/2\gamma)}
\end{equation}
which means that the error is indeed going to zero even very close to the boundary. At the same time, this is consistent with the assumption that $K \gg 1$: as soon as $n$ scales larger than linear in $d$, there are infinitely many spikes out of the bulk. One could have a fine grained analysis that studies the impact of learning finitely many spikes, but that's beyond the scope of this work. The same can be said for the boundary $n = d^{2 - \varepsilon}$ where the scaling of the risk is 
\begin{equation}
    \risk
    \asymp
    d^{-2\varepsilon/5}
\end{equation}
For $n\propto d$, the error will decrease in discrete steps, marked by the emergence of the spikes. For $n\propto d^2$, there will be an interpolation peak, as discussed in its own section later.

\paragraph{Phase VI: Large sample phase}
In this phase, similarly to the overregularized case, we assume $K=c(d)$, which means $\lambda\epsilon<2\delta\ll\sqrt{d}c(d)^{-\gamma}$. 

We start again by calculating the contribution from the spikes, similar to the previous phases with $K=c(d)$
\begin{align}
        &J_2(\delta,\lambda\epsilon) =\frac{1}{d}\sum_{i=1}^{c(d)}\left[\sqrt{d}i^{-\gamma}+\frac{\delta^2}{\sqrt{d}i^{-\gamma}}-\lambda\epsilon\right]^2\\
    &=\sum_{i=1}^{c(d)}\left[i^{-2\gamma}+\frac{\delta^4}{d^2i^{-2\gamma}}+\frac{1}{d}\lambda^2\epsilon^2-2\lambda\epsilon\frac{i^{-\gamma}}{\sqrt
    {d}}-2\lambda\epsilon\frac{\delta^2}{d^{3/2}i^{-\gamma}}+2\frac{\delta^2}{d}\right]\\
    &\approx Q^\star+\frac{c(d)}{d}(\lambda^2\epsilon^2+2\delta^2)-\frac{2\lambda\epsilon}{\sqrt{d}}\sum_{i=1}^{c(d)}i^{-\gamma}+\frac{\delta^4}{d^2}\frac{c(d)^{1+2\gamma}}{1+2\gamma}-\frac{2\lambda\epsilon\delta^2}{d^{3/2}}\frac{c(d)^{1+\gamma}}{1+\gamma}\\
    &\approx Q^\star+\frac{c(d)}{d}(\lambda^2\epsilon^2+2\delta^2)
    -\frac{2\lambda\epsilon}{\sqrt{d}}\sum_{i=1}^{c(d)}i^{-\gamma}
\end{align}
The last two terms can be discarded, since we can show that it is much smaller than 1 and thus $Q^\star$, using $\lambda\epsilon<2\delta\ll\sqrt{d}c(d)^{-\gamma}$, while the sum that remains will cancel in later since it is linear in $\lambda\epsilon$. Now we consider the two possible relations between $\delta$ and $\lambda\epsilon$: $2\delta\gg \lambda\epsilon$ and $1\gg 2\delta- \lambda\epsilon>0$.\\
If  $2\delta\gg \lambda\epsilon$, we use $0<s\ll 1$ and the first term of the relevant approximation \eqref{eq:Jab_approx} to write
\begin{align}
    J(\delta,\lambda\epsilon)\approx Q^\star+\delta^2\left(\frac{1}{2}+2\frac{c(d)}{d}\right)-\frac{2\lambda\epsilon}{\sqrt{d}}\sum_{i=1}^{c(d)}i^{-\gamma}
\end{align}
Then we write the state equations \eqref{eq:SE_ERM_app}
\begin{align}
    &
    \begin{dcases}
        4\alpha\delta^2-\frac{\delta^2}{\epsilon}\approx \delta^2\left(1+4\frac{c(d)}{d}\right)\\
    Q^\star+\frac{\Delta}{2}+2\alpha\delta^2-\frac{\delta^2}{\epsilon}\approx Q^\star+\delta^2\left(\frac{1}{2}+2\frac{c(d)}{d}\right)
    \end{dcases}\\
    &\implies \risk:=2\alpha\delta^2-\frac{\Delta}{2}\approx \delta^2\left(\frac{1}{2}+2\frac{c(d)}{d}\right)
\end{align}
This system admits the solution 
\begin{align}
    \delta^2=\frac{\Delta}{4\alpha-1-4\frac{c(d)}{d}}\qquad \epsilon=\frac{1}{4\alpha-1-4\frac{c(d)}{d}}
\end{align}
Since we expect $\risk\ll 1$ for enough samples, we get $\delta^2\ll 1$ and thus $\alpha\gg 1$. We can now plug this in the solution into the expression of the risk to get
\begin{align}
    \risk\approx \frac{\Delta}{8\alpha}\left(1+4\frac{c(d)}{d}\right)
\end{align}
The condition $2\delta\gg \lambda\epsilon$ translates to $\lambda\ll 4\sqrt{\Delta}\sqrt{\alpha}$, while $2\delta\ll \sqrt{d}c(d)^{-\gamma}$ gives 
\begin{align}
    4\delta^2=\frac{4\Delta}{4\alpha}\ll dc(d)^{-2\gamma}\quad \implies\quad n\gg \Delta dc(d)^{2\gamma}
\end{align}
Thus we have the phase VIb (taking $c(d)\to d$)
\begin{equation}
    \boxed{\risk= \frac{5\Delta d^2}{8n} \mathif \lambda\ll \sqrt{\frac{n}{d^2}}\mathand n\gg d^{2\gamma+1}}
\end{equation}
We now need to look for a phase $d^2\ll n\ll d^{2\gamma+1}$, that is not covered yet. It will be given by the case where not all spikes are outside the bulk: $\sqrt{d}c(d)^{-\gamma}\ll 2\delta\ll \sqrt{d}$ and $c(d)\gg K\gg 1$. Similarly to phase IV/V, we have the scaling $K\approx \left(\frac{\delta}{\sqrt{d}}\right)^{-1/\gamma}$. We get for the spike contribution
\begin{align}
        &J_2(\delta,\lambda\epsilon) =\frac{1}{d}\sum_{i=1}^{K}\left[\sqrt{d}i^{-\gamma}+\frac{\delta^2}{\sqrt{d}i^{-\gamma}}-\lambda\epsilon\right]^2\\
    &=\sum_{i=1}^{K}\left[i^{-2\gamma}+\frac{\delta^4}{d^2i^{-2\gamma}}+\frac{1}{d}\lambda^2\epsilon^2-2\lambda\epsilon\frac{i^{-\gamma}}{\sqrt
    {d}}-2\lambda\epsilon\frac{\delta^2}{d^{3/2}i^{-\gamma}}+2\frac{\delta^2}{d}\right]\\
    &= Q^\star-\sum_{i=K+1}^{c(d)}i^{-2\gamma}+\frac{K}{d}(\lambda^2\epsilon^2+2\delta^2)+\frac{\delta^4}{d^2}\sum_{i=1}^{K}i^{2\gamma}-\frac{2\lambda\epsilon\delta^2}{d^{3/2}}\sum_{i=1}^{K}i^{\gamma}-\frac{2\lambda\epsilon}{\sqrt{d}}\sum_{i=1}^{K}i^{-\gamma}\\
    &\approx Q^\star-\frac{K^{1-2\gamma}}{2\gamma-1}+\frac{K}{d}(\lambda^2\epsilon^2+2\delta^2)-\frac{2\lambda\epsilon}{\sqrt{d}}\sum_{i=1}^{K}i^{-\gamma}\approx Q^\star -\left(\frac{\delta}{\sqrt{d}}\right)^{2-1/\gamma}\left[\frac{1}{2\gamma-1}-\frac{\lambda^2\epsilon^2}{\delta^2}-2\right]-\frac{2\lambda\epsilon}{\sqrt{d}}\sum_{i=1}^{K}i^{-\gamma}
\end{align}
We first assume $\lambda\epsilon\ll 2\delta$. Then $J_1(\delta,\lambda\epsilon)\approx \frac{1}{2}\delta^2$ from \eqref{eq:Jab_approx}, and the state evolution becomes
\begin{align}
&
    \begin{dcases}
        4\alpha\delta^2-\frac{\delta^2}{\epsilon}\approx -\left(\frac{\delta}{\sqrt{d}}\right)^{2-1/\gamma}\left[\frac{1}{2\gamma-1}-2\right]\left(2-\frac{1}{\gamma}\right)+\delta^2\approx \delta^2\\
        Q^\star+\frac{\Delta}{2}+2\alpha\delta^2-\frac{\delta^2}{\epsilon}\approx Q^\star -\left(\frac{\delta}{\sqrt{d}}\right)^{2-1/\gamma}\left[\frac{1}{2\gamma-1}-2\right]+\frac{1}{2}\delta^2\approx \frac{1}{2}\delta^2
    \end{dcases}\\
    &\implies \risk\approx  \frac{1}{2}\delta^2
\end{align}
We can drop the first term in each equation since $2\delta\gg \sqrt{d}c(d)^{-\gamma}\gg \sqrt{d}d^{-\gamma}$ and 
\begin{equation}
    \left(\frac{\delta}{\sqrt{d}}\right)^{2-1/\gamma}\ll \delta^2 \quad \iff\quad \delta\gg d^{\frac{1}{2}-\gamma}
\end{equation}
We thus have the solution 
\begin{equation}
    \delta^2=\frac{\Delta}{4\alpha-1}\mathand \epsilon\approx\frac{1}{4\alpha-1}
\end{equation}
Since we know the behaviour at $n\ll d^2$, we can use $n\gg d^2$ to neglect the constant in the denominator to get $\risk=\frac{\Delta d^2}{8n}$ for the risk. The conditions $\lambda\epsilon\ll 2\delta$ and $2\delta\ll d^{\frac{1}{2}-\gamma}$ give respectively $\lambda\ll 4\sqrt{\Delta}\sqrt{\alpha}$ and $n\ll \Delta d^{1+2\gamma}$, and we thus have phase VIa:
\begin{equation}
    \boxed{\risk= \frac{\Delta d^2}{8n} \mathif \lambda\ll \sqrt{\frac{n}{d^2}}\mathand d^{2}\ll n\ll d^{1+2\gamma}}.
\end{equation}
This is the same scaling as phase VIb. 

The boundary is the same as phases IV and V:
\begin{equation}
    \boxed{\begin{aligned}
    R = \frac{2^{11/2}\Delta d^2}{105\pi n}\left(1-\frac{\lambda}{\lambda_c}\right)^{7/2}+
    \left(\frac{\Delta d}{4n}\right)^{1-1/2\gamma}
    \frac{24\gamma^3}{(4\gamma^2-1)(\gamma+1)}\\\qquad\quad\mathif 0<1-\frac{\lambda}{\lambda_c}\ll\left(\frac{n}{d^2}\right)^{2/5}\mathand d^2\ll n\ll d^{2\gamma+1}
    \end{aligned}}.
\end{equation}

\paragraph{Interpolation peak $n \asymp d^2$} We expect an interpolation peak to appear for $n\asymp d^2$. It can only happen in an under-regularized regime, which for this number of samples means $\lambda \ll1$. Inspired by \cite{erba2025nuclear}, we can conjecture that it will be for $\alpha = 1/4$. This will in fact be the correct position, as we will show.

For having an interpolation peak, we need the bulk to exist, so $\lambda \epsilon \ll 2\delta$. Depending on the number of spikes we can either have a first case with some but not all spikes out $d^{-\gamma + 1/2} \ll \delta \ll \sqrt{d}$, or a second one where none are out $\delta \gg \sqrt{d}$.

The first case has $J(\delta,\lambda\epsilon)\approx J_1(\delta)+J_2(\delta,\lambda\epsilon)$,
where by eq. (\ref{eq:Jab_approx})
\begin{align}
&J_1(\delta)\approx Q^\star +C(\gamma)\left(\frac{\delta}{\sqrt{d}}\right)^{2-\frac{1}{\gamma}}\\
&J_2(\delta,\lambda\epsilon)\approx\frac{\delta^2}{2}-\frac{8}{3\pi}\lambda\epsilon\delta.
\end{align}
Then eq. (\ref{eq:SE_ERM}) reduces to
\begin{align}
\begin{cases}
&4\alpha\delta^2-\frac{\delta^2}{\epsilon}=\delta^2-\frac{8}{3\pi}\lambda\epsilon\delta+C(\gamma)(2-1/\gamma)\left(\frac{\delta}{\sqrt{d}}\right)^{2-\frac{1}{\gamma}}\\
&Q^\star +\frac{\Delta}{2}+2\alpha\delta^2-\frac{\delta^2}{\epsilon}=Q^\star +\frac{1}{2}\delta^2+C(\gamma)\left(\frac{\delta}{\sqrt{d}}\right)^{2-\frac{1}{\gamma}}.
\end{cases}
\end{align}
Since $\delta \ll \sqrt{d}$, the one can drop the $(\delta/\sqrt{d})^{2-1/\gamma}$ piece from both equations.
By using $\alpha=1/4$, we obtain
\begin{align}
\epsilon=\frac{\Delta}{2}\lambda^{-2/3}\left(\frac{3\pi}{8}\right)^2,\ \delta=\left(\frac{3\pi\Delta^2}{32}\right)^{1/3}\lambda^{-1/3}
\end{align}
as the leading order solution.
Then we have
\begin{align}
\mathsf{R}:=2\alpha\delta^2-\frac{\Delta}{2}\approx2\left(\frac{3\pi\Delta^2}{32}\right)^{2/3}\lambda^{-2/3}.
\end{align}
The conditions for this case can be summarised in $\max(\lambda\epsilon,d^{-\gamma+1/2})\ll\delta\ll\sqrt{d}$, which reduces to
\begin{align}
d^{-3/2}\ll\lambda\ll1
\end{align}
The second case has conditions that can be compressed in $\max(\lambda\epsilon,\sqrt{d})\ll\delta$, which will cover the remaining case
\begin{align}
\lambda\ll d^{-3/2}.
\end{align}
In the second case one has $J(\delta,\lambda\epsilon)\approx\frac{\delta^2}{2}-\frac{8}{3\pi}\lambda\epsilon\delta$. Under this simplification, $\delta,\,\epsilon$ become
\begin{align}
\epsilon=\left(Q^\star +\frac{\Delta}{2}\right)\lambda^{-2/3}\left(\frac{3\pi}{8}\right)^2,\qquad\ \delta=\left(\frac{3\pi}{8}\right)^{1/3}\left(Q^\star +\frac{\Delta}{2}\right)^{2/3}\lambda^{-1/3}
\end{align}
and thus
\begin{align}
\mathsf{R}:=2\alpha\delta^2-\frac{\Delta}{2}\approx2\left(\frac{3\pi}{8}\right)^{2/3}\left(Q^\star +\frac{\Delta}{2}\right)^{4/3}\lambda^{-2/3}.
\end{align}
In conclusion, we have
\begin{equation}
\boxed{\risk= \Theta\left(\lambda^{-2/3}\right) \mathif \lambda\ll 1\mathand n=\frac{1}{4}d^2}.
\end{equation}

\subsection{Universal error decomposition of feature learning}
\label{app:decomposition}
In this section we derive Result~\ref{res:decomposition}. 

(i) When a part of the spikes are outside the bulk and a part of the spikes are inside, and the regularization does not cut off the bulk (e.g., in phases IV and V), we can rewrite the SE as
\begin{align}
\begin{cases}
&4\alpha\delta^2-\frac{\delta^2}{\epsilon}=\delta\partial_\delta (J_1(\delta,\lambda\epsilon)+J_2(\delta,\lambda\epsilon))\\
&Q^\star +\frac{\Delta}{2}+2\alpha\delta^2-\frac{\delta^2}{\epsilon}=(1-\lambda\epsilon\partial_{\lambda\epsilon})(J_1(\delta,\lambda\epsilon)+J_2(\delta,\lambda\epsilon)),
\end{cases}
\end{align}
where the auxiliary functions are defined as
\begin{align}
J_1(\delta,\lambda\epsilon):=\frac{1}{d}\sum_{i=1}^{K(\delta)}(s_i+\frac{\delta^2}{s_i}-\lambda\epsilon)^2, 
\end{align}
and
\begin{align}
J_2(\delta,\lambda\epsilon):=\delta^2\int_{\lambda\epsilon/\delta}^2\mu_{\rm sc}(x)(x-\lambda\epsilon/\delta)^2.
\end{align}
Recall we are considering a general model with $s_i$ denoting the $i-$th eigenvalue in a descending order and $K(\delta)\ll d$ satisfying $s_{K(\delta)}=\delta$. The excess risk is given by
\begin{align}
\mathsf{R}:=2\alpha\delta^2-\frac{\Delta}{2}=Q^\star +(\delta\partial_\delta+\lambda\epsilon\partial_{\lambda\epsilon}-1)(J_1(\delta,\lambda\epsilon)+J_2(\delta,\lambda\epsilon)).
\end{align}
Then we have
\begin{align}
\begin{aligned}
(\delta\partial_\delta+\lambda\epsilon\partial_{\lambda\epsilon}-1)J_1(\delta,\lambda\epsilon)&=\frac{1}{d}\delta K'(\delta)(2\delta-\lambda\epsilon)^2+\frac{2}{d}\sum_{i=1}^{K(\delta)}(s_i+\frac{\delta^2}{s_i}-\lambda\epsilon)(\frac{2\delta^2}{s_i}-\lambda\epsilon)-\frac{1}{d}\sum_{i=1}^{K(\delta)}(s_i+\frac{\delta^2}{s_i}-\lambda\epsilon)^2\\
&=\frac{1}{d}\delta K'(\delta)(2\delta-\lambda\epsilon)^2+\frac{1}{d}\sum_{i=1}^{K(\delta)}\left[(\frac{\delta^2}{s_i}-\lambda\epsilon)^2+\frac{\delta^2}{s_i}(s_i+\frac{\delta^2}{s_i}-\lambda\epsilon)\right]-\frac{1}{d}\sum_{i=1}^{K(\delta)}s_i^2.
\end{aligned}
\end{align}
and
\begin{align}
(\delta\partial_\delta+\lambda\epsilon\partial_{\lambda\epsilon}-1)J_2(\delta,\lambda\epsilon)=J_2(\delta,\lambda\epsilon).
\end{align}
Now we obtain eq. (\ref{eq:error_decomposition}) by using $Q^\star :=\frac{1}{d}\sum_{i=1}^ds_i^2$.

(ii) When the regularization cuts off the bulk and a part of the spikes (e.g., in phase III), we can rewrite the SE as
\begin{align}
\begin{cases}
&4\alpha\delta^2-\frac{\delta^2}{\epsilon}=\delta\partial_\delta J_1(\delta,\lambda\epsilon)\\
&Q^\star +\frac{\Delta}{2}+2\alpha\delta^2-\frac{\delta^2}{\epsilon}=(1-\lambda\epsilon\partial_{\lambda\epsilon})J_1(\delta,\lambda\epsilon),
\end{cases}
\end{align}
where the auxiliary function is still defined as
\begin{align}
J_1(\delta,\lambda\epsilon):=\frac{1}{d}\sum_{i=1}^{K(\delta,\lambda\epsilon)}(s_i+\frac{\delta^2}{s_i}-\lambda\epsilon)^2, 
\end{align}
but the cutoff becomes $s_{K(\delta,\lambda\epsilon)}+\frac{\delta^2}{s_{K(\delta,\lambda\epsilon)}}-\lambda\epsilon=0$. The excess risk is given by
\begin{align}
\begin{aligned}
\mathsf{R}:&=2\alpha\delta^2-\frac{\Delta}{2}=Q^\star +(\delta\partial_\delta+\lambda\epsilon\partial_{\lambda\epsilon}-1)J_1(\delta,\lambda\epsilon)\\
&=Q^\star +\frac{2}{d}\sum_{i=1}^{K(\delta,\lambda\epsilon)}(s_i+\frac{\delta^2}{s_i}-\lambda\epsilon)(\frac{2\delta^2}{s_i}-\lambda\epsilon)-\frac{1}{d}\sum_{i=1}^{K(\delta,\lambda\epsilon)}(s_i+\frac{\delta^2}{s_i}-\lambda\epsilon)^2\\
&=\frac{1}{d}\sum_{i=1}^{K(\delta,\lambda\epsilon)}\left[(\frac{\delta^2}{s_i}-\lambda\epsilon)^2+\frac{\delta^2}{s_i}(s_i+\frac{\delta^2}{s_i}-\lambda\epsilon)\right]+\frac{1}{d}\sum_{i=K(\delta,\lambda\epsilon)+1}^{d}s_i^2,
\end{aligned}
\end{align}
which gives eq. (\ref{eq:error_decomposition2}).

\subsection{Noiseless setting}
\subsubsection{Bayesian estimator}
In this section we solve eq. (\ref{eq:SE_Bayes}) for $\Delta=0$. For $\Delta=0$, the first equation of eq. (\ref{eq:SE_Bayes}) gives $\hat{q}=\frac{2\alpha}{Q^\star -q}=\frac{\alpha}{\mathsf{R}^{\rm BO}}$. 

\paragraph{Phase I: Perfect recovery} We can verify that $\alpha=\frac{c(d)}{2d}$ and $\mathsf{R}^{\rm BO}=0$ is a solution of eq. (\ref{eq:SE_Bayes}) for $\Delta=0$ because
\begin{align}
\lim_{\hat{q}\to\infty}\frac{4\pi^2}{3\hat{q}}\int\mu_{1/\sqrt{\hat{q}}}(x)^3\rd x=1-\frac{c(d)}{d}.
\end{align}
Then $\alpha=\frac{c(d)}{2d}$ is the perfect recovery threshold, which suggests that 
\begin{align}
\mathsf{R}^{\rm BO}=0\mathif n>\frac{c(d)}{2}d.
\end{align}

\paragraph{Phase II: Under-sampling}
if $d\ll n\ll d^2$, eq. (\ref{eq:integral_Bayes}) combined with eq. (\ref{eq:SE_Bayes}) gives
\begin{align}
-2\alpha=\Theta\left(d^{-1}\left(\frac{\alpha d}{\mathsf{R}^{\rm BO}}\right)^{1/2\gamma}\right),
\end{align}
and thus
\begin{align}
\mathsf{R}^{\rm BO}=\Theta\left(\left(\frac{n}{d}\right)^{2\gamma-1}\right).
\end{align}
Note that for this case we still have $\delta^{1/\gamma}d^{-1}\ll 1$, so the derivation of eq. (\ref{eq:integral_Bayes}) is valid.

\paragraph{Phase III: Not enough data}  If $n\ll d$, eq. (\ref{eq:integral_Bayes_small_n}) combined with eq. (\ref{eq:SE_Bayes}) gives
\begin{align}
2\alpha=Q^\star \frac{2\alpha}{Q^\star -q},
\end{align}
and thus $\mathsf{R}^{\rm BO}:=Q^\star -q=Q^\star $.

\subsubsection{ERM}
In this section we solve the equations in Appendix \ref{app:noisy_ERM} for $\Delta=0$.
\paragraph{Phase I: Trivial Phase}
The results in Phase I of Appendix \ref{app:noisy_ERM} holds also for $\Delta=0$, which gives Phase I:
\begin{align}
\mathsf{R}=Q^\star , \mathif n<\frac{1}{8}Q^\star d\mathor \lambda>\frac{4n}{d^{3/2}}.
\end{align}

\paragraph{Phase II: Over-regularization phase}
The analysis of Phase II in Appendix \ref{app:noisy_ERM} also holds for $\Delta=0$, which gives
\begin{align}
\begin{cases}
&4\alpha\delta-\frac{\delta}{\epsilon}=0\\
&Q^\star+2\alpha\delta^2-\frac{\delta^2}{\epsilon}=Q^\star -\frac{2\gamma}{2\gamma-1}\left(\frac{\lambda\epsilon}{\sqrt{d}}\right)^{\frac{2\gamma-1}{\gamma}}.
\end{cases}
\end{align}
The leading order solution is $\epsilon=\frac{1}{4\alpha}$, $\delta^2=\frac{\gamma}{(2\gamma-1)\alpha}\left(\lambda\frac{d^{3/2}}{4n}\right)^{\frac{2\gamma-1}{\gamma}}$ and
\begin{align}
\mathsf{R}:=2\alpha\delta^2=\frac{2\gamma}{2\gamma-1}\left(\lambda\frac{d^{3/2}}{4n}\right)^{\frac{2\gamma-1}{\gamma}}.
\end{align}
The condition $\max(\delta,d^{-\gamma+1/2})\ll\lambda\epsilon\ll\sqrt{d}$ gives
\begin{align}
\max\left(\frac{d^{\gamma-3/2}}{n^{\gamma-1}},\frac{n}{d^{\gamma+\frac{3}{2}}}\right)\ll\lambda\ll\frac{d^{3/2}}{n}.
\end{align}

\paragraph{Phase III: Intermediate over-regularization phase}
Phase III is given by the analysis of Phase III in Appendix \ref{app:noisy_ERM}. We get the same solution
\begin{align}
\epsilon=\frac{1}{4\alpha},\ \delta^2=\frac{\lambda^2d^6}{32n^3}
\end{align}
for $\Delta=0$. Thus we have
\begin{align}
\mathsf{R}:=2\alpha\delta^2=\frac{\lambda^2d^4}{16n^2}.
\end{align}
The condition $\delta\ll\lambda\epsilon\ll d^{-\gamma+1/2}$ gives
\begin{align}
\lambda\ll\frac{n}{d^{\gamma+3/2}}\mathand n\gg d^2.
\end{align}

\paragraph{Phase IV: Under-sampling phase}
Phase IV is given by the analysis of Phases IV and V in Appendix \ref{app:noisy_ERM}. The risk reduces to
\begin{align}
\risk:=2\alpha\delta^2=\left(\frac{24\gamma^3}{(4\gamma^2-1)(\gamma+1)}\right)\left(\frac{\delta}{\sqrt{d}}\right)^{2-\frac{1}{\gamma}}+\delta^2\frac{16t^{7/2}}{105\pi}
\label{eq:phaseIV-noiseless}
\end{align}
for $\Delta=0$ and $\frac{16}{15\pi}t^{5/2}\approx4\alpha$, where we assume $\lambda\ll\alpha\delta$. Then the second term on the right side is negligible since $t^{7/2}\ll\alpha$. \eqref{eq:phaseIV-noiseless} gives
\begin{align}
\delta=2^{-\gamma}\left(\frac{24\gamma^3}{(4\gamma^2-1)(\gamma+1)}\right)^\gamma\frac{d^{\gamma+1/2}}{n^\gamma},
\end{align}
and then we further have
\begin{align}
\mathsf{R}:=2\alpha\delta^2=2^{-2\gamma+1}\left(\frac{24\gamma^3}{(4\gamma^2-1)(\gamma+1)}\right)^{2\gamma}\left(\frac{d}{n}\right)^{2\gamma-1}.
\end{align}
The condition $d\ll n\ll d^2$ and $\lambda\ll\alpha\delta$, $0<2-\frac{\lambda\epsilon}{\delta}\ll1$, $d^{-\gamma+\frac{1}{2}}\ll\delta\ll\sqrt{d}$ reduces to
\begin{align}
\lambda\ll\frac{d^{\gamma-3/2}}{n^{\gamma-1}}\mathand d\ll n\ll d^2,
\end{align}
where we use $\delta^2t^{5/2}\sim4\alpha\delta^2-\frac{\lambda\delta}{2}>0$. Note that we further have $t^{5/2}\sim4\alpha-\frac{\lambda}{2\delta}\ll1$ satisfied.

\section{Comparison with $L_2$ regularization}
In this section we compare the scaling laws we have obtained for ERM to the ones of ridge regression for a linear model, proving in particular their sub-optimality. The ridge estimator is defined as
\begin{equation}\label{app:def:ridge_estimator}
    \hat \vtheta_{\rm ridge} = \underset{\vtheta \in\R^d}{\arg\min} \frac{1}{n}\sum_{\mu=1}^n \left(y_\mu - {\langle\vtheta,\vx_\mu\rangle}\right)^2 + \lambda\lVert\vtheta\rVert_2^2.
\end{equation}
This can be mapped to both the diagonal network and quadratic network case, depending on the choice of $\vx$. For simplicity, we assume $\vx\sim\mathcal{N}(0,I_d)$.
\cite{cheng2024dimension} readily implies the following.
\begin{theorem}[Excess risk rates for ridge regression]. Assume that $y=\langle\vtheta^\star,\vx\rangle+\sqrt{\Delta}\zeta$ with $\zeta\sim\mathcal{N}(0,1)$ and $\mathbb{E}[||\btheta^\star||_2^2]=\Theta(1)$. For $n,d\gg 1$, the excess risk associated to the estimator defined in (\ref{app:def:ridge_estimator}) satisfies
    \begin{equation}
    \mathsf R_{n,d} = \Theta\,\begin{cases}
        1, &\mathif n\ll d\mathor\lambda \gg 1,\\
        \lambda ^ 2,&\mathif n\gg d \mathand \sqrt{d/n}\ll\lambda \ll 1\\
        \Delta d / n,&\mathif n\gg d \mathand \lambda \ll \sqrt{d/n}\\
        \Delta \lambda^{-1/2},&\mathif n = d\mathand \lambda \ll 1.
    \end{cases}
\end{equation}
\end{theorem}
\begin{proof}
The excess risk concentrates with high probability, for $n,d\gg 1$, around the following deterministic expression \citep{cheng2024dimension}
\begin{equation}\label{app:eq:ridge_risk}
    \mathsf R_{n,d} = \frac{n\nu^2}{n(1+\nu)^2-d}\mathbb{E}[||\vtheta^\star||_2^2] + \frac{d\Delta}{n(1+\nu)^2-d} 
\end{equation}
with $\nu$ the unique non-negative solution of
\begin{equation}
    \frac{n}{d}\left(1-\frac{\lambda}{\nu}\right) = \frac{1}{1+\nu}.
\end{equation}
Therefore 
\begin{equation}
    \nu = \Theta\,\begin{cases}
        d/n + \lambda,&\mathif n \ll d,\\
        \lambda, & \mathif n\gg d,\\
        \sqrt{\lambda},&\mathif n = d\mathand\lambda\ll 1.
    \end{cases}
\end{equation}
Substituting into \eqref{app:eq:ridge_risk}, the result follows.
\end{proof}
Therefore, we cannot obtain a non-trivial risk for $n\ll d$ with $L_2$ regularization.

\section{Numerical details}
\label{app:numerical}

The state equations \eqref{eq:SE_ERM} can be iterated in a more numerically convenient form that depends on an extended set of parameters
$(q,m,\Sigma,\hat{q},\hat{m},\hat{\Sigma})$ as presented in \cite{erba2025nuclear}, Appendix A.4.4.
Then, we can compute $\delta$ and $\epsilon$ as $\delta = \sqrt{\hat{q}}/{\hat{m}}$ and $\epsilon = 2 / \hat{m}$.
The limiting distribution of $S^\star$ is not easy to compute exactly as $d\to\infty$, and for finite $d$ the contribution of the spikes to the state evolution is the sum $J_1(\delta,\lambda\epsilon)$, whose upper limit depends on the width of the bulk $\delta$. The state equations include derivatives of this sum, where delta spikes are neglected due to the numerical difficulty of representing them. Having obtained results that do not match the experiments with these approximations, we resorted to computing the integral $J$ by a Monte-Carlo procedure. The overlaps $m,q,\Sigma$ are computed using finite size samples of matrices and their eigen-decomposition at each step $t$ of the state evolution
\begin{align}
    M=\sqrt{\hat{q}^t}Z+\hat{m}^tS^\star=O\diag (\nu_1,\ldots,\nu_d)O^T
\end{align}
where $Z\sim \GOE(d)$ and $S^\star=\tfrac{\sqrt{d}}{\sum_ii^{-2\gamma}}\diag(1,2^{-\gamma},\cdots,d^{-\gamma})$ can be taken as a diagonal matrix, since this amounts to a rotation of $M$, which does not affect the distribution of $Z$ by rotational invariance. One can then apply the spectral denoiser described in \cite{erba2025nuclear} and compute the overlaps using the reconstructed matrix $\Tilde{M}=O\diag(\Tilde{\nu}_1,\ldots,\Tilde{\nu}_d)O^T$, where $\Tilde{\nu}_i=\frac{1}{\hat{\Sigma}}\relu(\nu_i-2\reg)$ are the denoised eigenvalues. Finally, the order parameters can be computed as
\begin{align}
    \begin{cases}
        m^{t+1}=\frac{1}{d}\E_M\Tr[(S^\star)^T\Tilde{M}]\\
        q^{t+1}=\frac{1}{d}\E_M\Tr[\Tilde{M}^T\Tilde{M}]\\
        \Sigma^{t+1}=\frac{2}{d}\E_M\left[\sum_{i=1}^{d}\frac{\Theta(\nu_i-2\reg)}{\Hat{\Sigma}}+\sum_{i<j}\frac{\Tilde{\nu}_i-\Tilde{\nu}_j}{\nu_i-\nu_j}\right]
    \end{cases}
\end{align}
The expectation is taken over $n_{samples}\sim10$ samples for $d\sim10^2$, independently for each order parameter, for a total of $3n_{samples}$ sampled matrices per iteration of the state evolution. 

Since the ERM problem is convex we resort to using LBFGS with Wolfe line search. We used the PyTorch implementation of the optimiser, taking care of evaluating the network efficiently at each pass. For the specifics of the implementation we refer to the code repository \url{https://github.com/SPOC-group/QuadraticNetPowerlaw}. Convergence is typically achieved with a precision of at least $10^{-8}$ in a few hundred iterations. The main challenge is in storing the dataset in memory. For each run we used up to $1800$ gigabytes of RAM on nodes with 2 Intel Xeon 8360Y CPUs. Our total computing cost (including intial explorations) is around $250000$ CPU hours.

\end{document}